\documentclass{article}

 \usepackage[main,preprint]{neurips_2026}

\PassOptionsToPackage{numbers}{natbib}

\usepackage[utf8]{inputenc} 
\usepackage[T1]{fontenc}    
\usepackage{hyperref}       
\usepackage{url}            
\usepackage{booktabs}       
\usepackage{amsfonts}       
\usepackage{nicefrac}       
\usepackage{microtype}      
\usepackage{xcolor}         
\usepackage{siunitx}
\usepackage{graphicx}
\usepackage{dsfont}
\usepackage{subcaption} 
\usepackage{wrapfig}
\usepackage{amssymb}
\usepackage{colortbl}
\usepackage{arydshln}
\usepackage{multicol,multirow}
\usepackage{float}
\usepackage{placeins} 

\usepackage[acronym]{glossaries}
\glsdisablehyper

\makeglossaries
\newacronym{mlwp}{MLWP}{Machine learning Weather Prediction}
\newacronym{nwp}{NWP}{Numerical Weather Prediction}
\newacronym{ml}{ML}{Machine Learning}
\newacronym{lam}{LAM}{Limited-Area Modeling}
\newacronym{crps}{CRPS}{Continuous Ranked Probability Score}
\newacronym{cdf}{CDF}{Cumulative Distribution Function}
\newacronym{rmse}{RMSE}{Root Mean Squared Error}
\newacronym{ssr}{SSR}{Spread-Skill Ratio}
\newacronym{lsd}{LSD}{Log Spectral Distance}
\newacronym{gnn}{GNN}{Graph Neural Network}
\newacronym{cnn}{CNN}{Convolutional Neural Network}
\newacronym{mse}{MSE}{Mean Squared Error}

\title{CRPS-LAM: Probabilistic Regional Weather Forecasting with Continuous Ranked Probability Score}

\author{%
  Erik Wikingsson \\
  Linköping University \\
  \texttt{erik.wikingsson@gmail.com} \\
  \And
  Joel Oskarsson \\
  ETH AI Center \\
  \texttt{joel.oskarsson@ai.ethz.ch} \\
  \AND
  Tomas Landelius \\
  SMHI \& Linköping University \\
  \texttt{tomas.landelius@smhi.se} \\
  \And
  Fredrik Lindsten \\
  Linköping University \\
  \texttt{fredrik.lindsten@liu.se} \\
}

\usepackage[capitalise,noabbrev]{cleveref}

\begin{document}

\maketitle

\begin{abstract}
Limited-Area Models (LAMs) enable weather forecasting over regional domains at higher resolutions than what is computationally feasible for global models. At such high resolutions, machine learning approaches for weather prediction increasingly rely on ensemble methods to produce probabilistic forecasts.
However, existing machine learning LAMs are not scalable due to relying on computationally costly diffusion models or inefficient graph neural networks. We tackle this by introducing a new hybrid CNN/GNN architecture, tailored to the LAM weather forecasting problem. Using this architecture, we construct the DET-LAM deterministic model, producing LAM forecasts both more efficiently and accurately than its graph-based competitor. We then tackle the ensemble forecasting problem, by using this architecture as a backbone for the generative model CRPS-LAM. CRPS-LAM is trained using a Continuous Ranked Probability Score (CRPS) objective, enabling efficient training and sampling in a single forward pass. This yields a speedup of $\approx \times 39$ compared to diffusion-based baselines. We evaluate our approach on regional domains in northern Europe, demonstrating that CRPS-LAM produces skillful and well-calibrated forecasts across a range of atmospheric variables.
\end{abstract}

\section{Introduction}
Machine learning methods for simulating atmospheric dynamics are revolutionizing modern weather forecasting \citep{pangu,graphcast,lang2024aifsecmwfsdatadriven,gencast,lang2024aifscrpsensembleforecastingusing,fourcastnet3}.
Offering fast and accurate forecasts, these \gls{mlwp} methods have great potential for predicting and understanding both everyday weather conditions and extreme events.
While initial \gls{mlwp} methods used deterministic models to produce single forecasts \citep{pangu,lang2024aifsecmwfsdatadriven}, the field is increasingly moving towards more useful ensemble forecasting models \citep{gencast,lang2024aifscrpsensembleforecastingusing,fourcastnet3,FGNalet2025skillfuljointprobabilisticweather}.
These methods use conditional generative models to sample forecasts from the distribution of possible future atmospheric states.
Such sampled ensemble members can then be used to estimate probabilities of specific atmospheric conditions, or studied by meteorologists in order to understand possible future scenarios.
These developments towards \gls{mlwp} ensemble forecasting are happening both for the global domain, and for the regional setting considered in this work.
Regional \gls{mlwp} allows for building models operating at higher resolutions than what is possible in a global setting, due to excessive memory and compute costs. Regional models also enable training on region-specific datasets that provide finer-grained or locally calibrated information \citep{DANRA, CERRA}.

Initial attempts at \gls{mlwp} ensemble forecasting for regional domains have relied on simplified and often unrealistic problem settings, and either fail to capture high-frequency features \citep{oskarsson2024probabilistic} or use computationally expensive diffusion models \citep{stormCast,larsson2025diffusionlam,hrrrcast}. Furthermore, compared to previous work, we adopt a more realistic formulation for boundary conditioning following \citet{adamov2025buildingmachinelearninglimited} and extend it to the probabilistic setting.

Building on a methodology that has been proven successful in global forecasting \citep{FGNalet2025skillfuljointprobabilisticweather}, we show that regional ensemble forecasting models can be trained effectively by matching marginal distributions using a \gls{crps} loss. We propose a probabilistic regional forecasting model where stochasticity is introduced through a single latent noise vector, enabling the generation of skillful and realistic ensemble members in a single forward pass. 

\begin{wrapfigure}[14]{r}{0.35\textwidth}%
    \centering%
    \vspace{-1.8em}
    \includegraphics[width=0.35\textwidth]{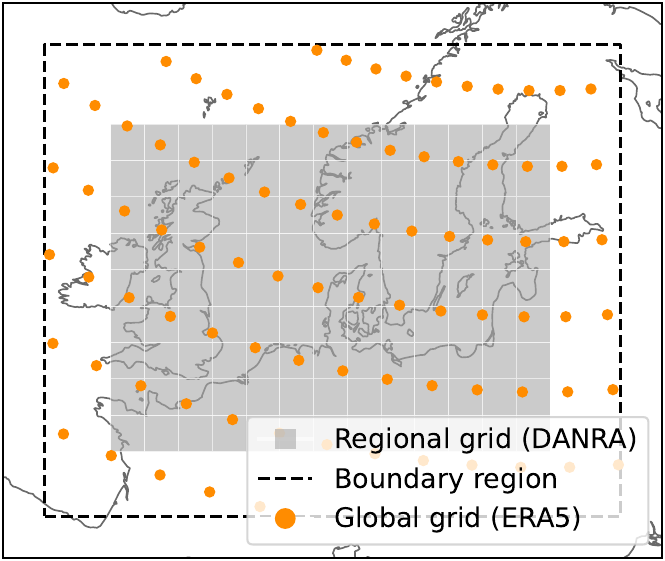}%
    \caption{Regional and global grid points might not align in our setting. 
    Low resolution example for clarity.}
    \label{fig:grid_locations}
    \vspace{-2.0em}
\end{wrapfigure}

\paragraph{\gls{lam} Problem Definition.}
In this paper, we study probabilistic \gls{mlwp} in the context of \gls{lam}. We adopt the problem formulation of \citet{adamov2025buildingmachinelearninglimited} and extend it to the probabilistic forecasting setting.

In \gls{lam} forecasting, the goal is to produce high-resolution forecasts over a bounded spatial domain $\Omega^I$, which we refer to as the \emph{interior}. 
While the model is restricted to a limited domain, there is also influence from large-scale atmospheric dynamics from outside the region, particularly at longer lead times. 
To account for this, we condition on information from a boundary region $\Omega^B$, for example obtained from a global forecast at coarser resolution. Importantly, this conditioning is treated as a \emph{soft} rather than a hard constraint, allowing the regional model to deviate from the global model where appropriate and thereby avoid inheriting its systematic biases.
The inputs to the model are:

\begin{itemize}
    \item Initial interior states $X_{t-1}^I, X_t^I$ within $\Omega^I$. These are assumed to be given, for example, from a data assimilation system in an operational setting.
    \item Interior forcing inputs $F_t^I$ within $\Omega^I$, which may include time-dependent variables (e.g., time of day) as well as static spatial features (e.g., orography and land--sea masks).
    \item Boundary information $B_t$, consisting of atmospheric states $X_{t-1}^B, X_t^B, X_{t+1}^B$ and forcing features $F_t^B$, defined on $\Omega^B$. Note that $\Omega^B$ may spatially overlap with $\Omega^I$  but boundary information is typically available at a distinct and lower-resolution grid (see  \cref{fig:grid_locations}).
\end{itemize}

Our objective is to generate forecasts over the interior domain by modeling the conditional distribution
$p\big(X_{t+1}^{I} \mid X_t^I, X_{t-1}^I, F_t^I, B_t\big)$,
which can be accomplished by a mean prediction from a deterministic model or an ensemble forecast by a probabilistic model. An overview of the limited-area forecasting setup and our model is illustrated in \cref{fig:architecure_overview}.

\paragraph{Our main contributions are:}
\begin{enumerate}
     \item We propose a hybrid \gls{gnn} and \gls{cnn} architecture to be used as the backbone for \gls{lam} weather models. This design combines the best of both worlds: handling spatially irregular boundary inputs using the \gls{gnn} component while leveraging the computational and memory benefits of \glspl{cnn}.
    \item We introduce \textbf{DET-LAM}, a deterministic forecasting model which shows clear improvement in speed and accuracy compared to graph-based baselines.
    \item We introduce \textbf{CRPS-LAM}, a probabilistic forecasting model producing accurate ensemble forecasts through single forward pass sampling.
    \item We evaluate our models for high-resolution forecasting over northern European domains, showing the advantage over purely graph-based existing methods.
\end{enumerate}

\section{Related Work}
\paragraph{Probabilistic Global Weather Forecasting.}
Probabilistic global \gls{mlwp} has seen significant progress through using generative models based on latent variable formulations \citep{oskarsson2024probabilistic, SwinVRNN}, diffusion models \citep{gencast, andrae2024continuousensembleweatherforecasting, shi2024codicastconditionaldiffusionmodel}, and flow-matching approaches \citep{couairon2024archesweatherarchesweathergendeterministic}. While flow-based methods have demonstrated strong performance, they typically require iterative sampling at inference time, resulting in substantial computational overhead (often $20\text{--}40\times$ slower).
\citet{stock2025swiftautoregressiveconsistencymodel} address this limitation by distilling a global weather diffusion model into a consistency model, capable of generating samples in a single forward pass.
Others immediately train for single step forecasting, injecting random noise as a model input and training using a \gls{crps} loss function to match the correct distribution \citep{prob_fc_scoring_rules, lang2024aifscrpsensembleforecastingusing, lang_multi-scale, fourcastnet3, FGNalet2025skillfuljointprobabilisticweather, oskarsson2024probabilistic}.
These methods differ primarily in how the \gls{crps} is estimated and how the stochasticity is incorporated into the model.

\paragraph{\acrfull{lam}.}

In the \gls{lam} setting, regional \gls{mlwp} is achieved by combining coarse resolution boundary information with dynamics simulated forward on a high-resolution grid.
Deterministic machine learning \glspl{lam} include Graph-FM \citep{oskarsson2023graph-lam,adamov2025buildingmachinelearninglimited}, using a \gls{gnn} architecture, and YingLong \citep{xu2024yinglongskillfulhighresolution}, using a neural operator approach.
Probabilistic \gls{lam} forecasting has so far mainly been explored using diffusion models \citep{stormCast, larsson2025diffusionlam, hrrrcast}. 
As for global forecasting, diffusion approaches can achieve strong predictive performance but at a high computational cost.
\citet{oskarsson2024probabilistic} propose a \gls{lam} version of the Graph-EFM latent variable model, capable of single step sampling.
Graph-EFM incorporates \gls{crps}-based regularization, but uses a more involved training objective with additional hyperparameters, compared to our approach.
As shown in our experiments, Graph-EFM also tends to produce overly smooth forecast fields. 
While existing probabilistic methods all incorporate boundary information, they all do so at the same spatio-temporal resolution as the regional domain. 
In real-world scenarios, boundary data typically has a much lower resolution.
Only \citet{adamov2025buildingmachinelearninglimited} handle boundary inputs at a different resolution, but only for deterministic models. We improve on their approach with our new hybrid architecture and extend it to the probabilistic setting.
\begin{wrapfigure}[13]{r}{0.35\textwidth}%
    \centering%
    \vspace{-.9em}
    \includegraphics[width=0.35\textwidth]{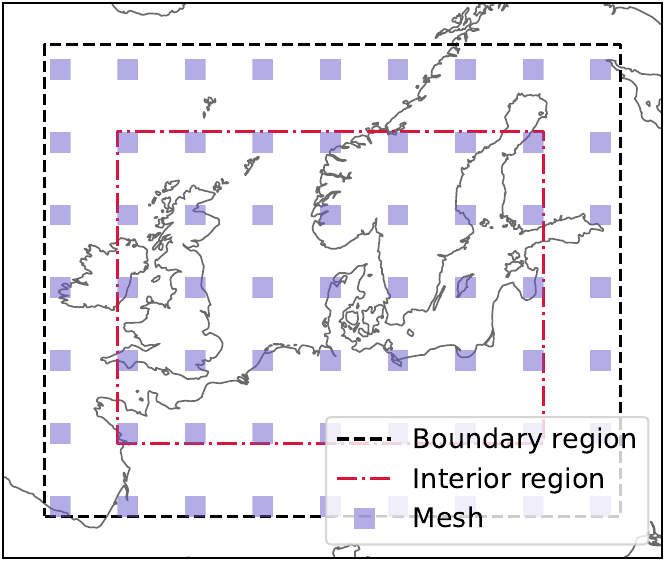}%
    \caption{Example mesh covering both interior and boundary regions.}
    \label{fig:mesh_position}
\end{wrapfigure}



\paragraph{High-Resolution Regional Weather Forecasting.}
Apart from \glspl{lam}, alternative approaches to high-resolution regional forecasting include stretch-grid models and downscaling.
Stretch-grid models \citep{stretch_grid_norway} combine global and regional forecasting by using a higher resolution over a specific region of the globe. 
\Citet{nordhagen2025highresolutionprobabilisticdatadrivenweather} extend the stretch-grid approach to probabilistic forecasting with a similar \gls{crps}-based training as CRPS-LAM, however the method still requires full global weather simulations, making it less scalable than a \gls{lam} approach.
Downscaling instead aims to produce high-resolution weather fields by learning a mapping from a low-resolution global forecast to high-resolution regional predictions \citep{CorrDiff,srivastava2024precipitation,larsson2026climatedownscalingstochasticinterpolants}.
While this enables generating regional predictions, downscaling methods do not explicitly simulate the underlying fine-scale physical processes at high resolution, and can thus not incorporate information from regional initial conditions.
As our methods combine both regional initial conditions and coarse boundary information, they can seamlessly transition from simulating dynamics to pure downscaling as the information in the initial condition becomes less relevant.

\section{Method}

\paragraph{Method Overview.}
We approach the \gls{lam} forecasting problem by constructing a hybrid \gls{cnn}/\gls{gnn} backbone architecture, and training it as a generative model using the \gls{crps} scoring rule as the loss function.
This backbone leads to improved memory and compute efficiency compared to previous approaches.
We use the backbone model $f_\theta$ as 
\begin{equation}
\hat{X}^{I}_{t+1} = f_\theta(X_t^I, X_{t-1}^I, F_t^I, B_t, z)
\end{equation}
where $z$ is latent Gaussian noise used to sample an ensemble member one step into the future in a single forward pass.
The \gls{crps} loss function provides a straightforward way to train such a model for calibrated forecasts.
We next describe details of the different parts in this modeling framework.

\begin{figure}[t]
\vspace{-1.0em}

\begin{center}
\includegraphics[width=\textwidth]{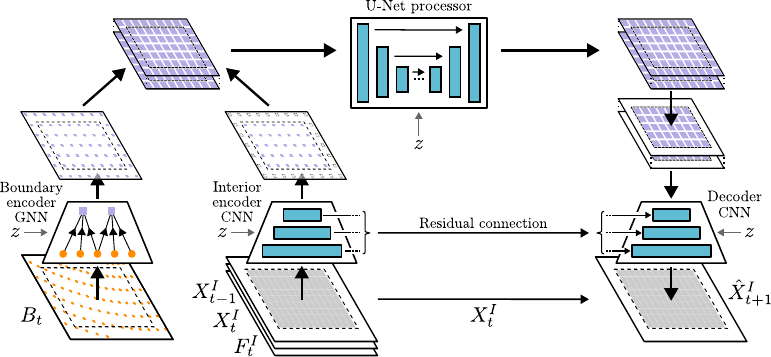}
\end{center}
\caption{
Overview of the CRPS-LAM backbone architecture.
The same latent noise vector $z$ is used in each component, but linearly transformed by independent parameters before each conditional normalization layer.
}
\label{fig:architecure_overview}
\vspace{-1.0em}

\end{figure}

\paragraph{Architecture.}
Our architecture follows an encode--process--decode framework \citep{battaglia2018relational}. First, all inputs are mapped onto a shared mesh representation; processing is then performed on this mesh, and finally the forecast is decoded over the interior region of interest. 
We carefully choose this mesh such that it lines up with a subset of points of the interior grid, but also extends out across the boundary region (see \cref{fig:mesh_position}).
An overview of the full architecture is shown in \cref{fig:architecure_overview}.
In the \gls{lam} setting, interior and boundary data may be defined on different grids, motivating the use of separate encoders. The interior data are typically given on a regular grid, and we therefore follow \citet{siddiqui2024exploringdesignspacedeeplearningbased} and employ a U-Net architecture to efficiently extract spatial features. In contrast, the boundary data are defined on a lower-resolution latitude--longitude grid, for which we use a \gls{gnn} encoder to flexibly map the inputs onto the regular grid mesh.
 
In early experiments with a purely GNN-based architecture 
(following \citet{adamov2025buildingmachinelearninglimited}), we found that GNN modules incur a substantial memory footprint.
This motivated a hybrid design in which CNN-based modules are used whenever the data lie on a regular grid, while GNNs are retained only for irregular boundary grids. 
To further reduce memory usage, we use only 32 dimensions for edge representation vectors, compared to the much higher hidden dimension used in other model components.
This results in a significant reduction in memory cost, without negatively affecting performance. The mesh itself is constructed on a regular grid, which allows us to again leverage a U-Net architecture for the processing stage. After processing, the interior nodes of the mesh are decoded using a CNN-based decoder to produce forecasts over the interior domain.

\paragraph{Deterministic Model (DET-LAM).}
Our efficient backbone can directly be used as a deterministic \gls{lam} model by simply setting the noise variables $z$ to a constant 1. We call this model DET-LAM.
Following earlier works on deterministic \glspl{lam} \citep{adamov2025buildingmachinelearninglimited,oskarsson2023graph-lam,wijnands2025comparison}, we train this model by minimizing a \gls{mse} loss function averaged over variables, grid points and lead times.
DET-LAM is first trained for single-step prediction, and then finetuned using rolled out forecasts.

\paragraph{Probabilistic Model (CRPS-LAM).}
Scoring rules provide a principled framework for evaluating probabilistic forecasts and has been used to train ensemble forecasting models \citep{lang2024aifscrpsensembleforecastingusing,fourcastnet3,FGNalet2025skillfuljointprobabilisticweather}. A more detailed introduction to \gls{crps} as a training objective is provided in \cref{apx:crps_intro}. Following \citet{FGNalet2025skillfuljointprobabilisticweather}, we introduce stochasticity via a low-dimensional latent variable. Specifically, a noise vector $z \sim \mathcal{N}(0, I)$ of dimension 32 is transformed through a linear layer and injected into the network using conditional normalization layers \citep{chen2021adaspeechadaptivetextspeech, song2020score, karras2022elucidating}. Given interior states $X_t^I, X_{t-1}^I$, forcing features $F_t^I$ and boundary conditions $B_t$, the model defines a conditional distribution
$p\big(X_{t+1}^{I} \mid X_t^I, X_{t-1}^I, F_t^I, B_t\big)$.
Multi-step forecasts are obtained through autoregressive rollouts, where predictions are recursively fed back as inputs. We train the model by minimizing the almost-fair \gls{crps} estimator $\text{CRPS}_{\alpha\text{-fair}}$ \citep{lang2024aifscrpsensembleforecastingusing} using two sampled ensemble members $\hat{X}_{t+1}^{I,(1)}$ and $\hat{X}_{t+1}^{I,(2)}$. The overall loss is defined as
\begin{equation}\label{eq:training_loss}
   \mathcal{L} = \frac{1}{T|G|} \sum_{t=0}^{T-1} \sum_{g \in G} \sum_{d} \omega_d \sigma_d \text{CRPS}_{\alpha\text{-fair}}(\hat{X}_{t+1,g,d}^{I,(1)},\hat{X}_{t+1,g,d}^{I,(2)}, X^I_{t+1, g,d})
\end{equation}
where the \gls{crps} is computed for each lead time $t$, grid point $g$ and variable $d$ by comparing predictions from the two ensemble members and the ground truth $X^I_{t+1,g,d}$.
This is then weighted by the inverse variance $\sigma_d$ of the one-step-differences for the variable, as well as a manual variable-specific weight~$\omega_d$.
Finally the \gls{crps} is summed over variables and averaged across all grid points $G \subset \Omega^I$and lead times $1, \dots, T$.
We choose the manual weights $\omega_d$ similar to existing works for the datasets used in experiments \citep{oskarsson2023graph-lam,adamov2025buildingmachinelearninglimited}. 
Training follows a similar rollout schedule as the deterministic model.

While the point-wise loss primarily enforces marginal consistency at each grid point, joint dependencies are captured through the shared latent variable $z$, which influences all outputs simultaneously. Additionally, the use of convolutional architectures promotes spatial coherence through weight sharing. We have also experimented with adding a spectral loss component, similar to \citet{nordhagen2025highresolutionprobabilisticdatadrivenweather} and \citet{fourcastnet3}. However, this requires choosing additional weighting factors, and we found that rather than improving model performance it could introduce artifacts in the forecasts.

\section{Experiments}\label{sec:experiments}
We start by evaluating the models in a simplified setting using the MEPS dataset in \cref{sec:MEPS_experiments}, where we compare our \gls{crps}-based approach to Diffusion-LAM \citep{larsson2025diffusionlam}, a \gls{cnn}-based diffusion model, and Graph-EFM \citep{oskarsson2024probabilistic}, a graph-based latent variable approach. We then experiment with the more realistic high-resolution DANRA dataset in \cref{sec:danra_experiments}.
Forecasts are evaluated using point-wise metrics, including \gls{rmse}, \gls{ssr}, and \gls{crps}. In addition, we analyze the spectral characteristics of the predictions.

\subsection{A Simplified Setting: the MEPS Dataset}\label{sec:MEPS_experiments}
\begin{figure}[tbph]
\begin{center}
\vspace{-0.5em} 
\includegraphics[width=\textwidth]{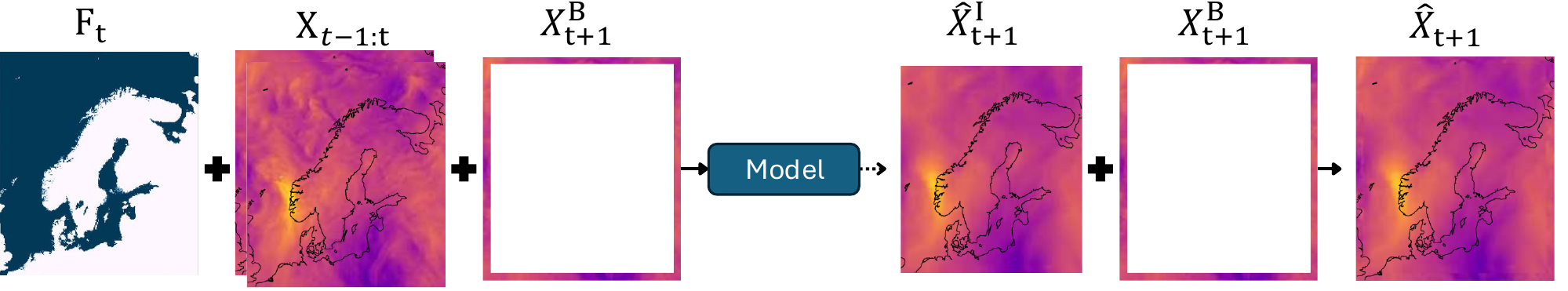}
\end{center}
\caption{An overview of the forecasting process showing the inputs and outputs of the models in the MEPS problem setting (adapted from \citet{larsson2025diffusionlam}).}
\label{fig:model_process}
\vspace{-1em} 
\end{figure}

We start by considering a previously studied simplified version of the problem defined above, in which the boundary and interior share the same spatial grid and do not overlap \citep{oskarsson2023graph-lam,oskarsson2024probabilistic,larsson2025diffusionlam}. In particular, we decompose the domain $\Omega = \Omega^B \cup \Omega^I$ into a non-overlapping interior domain $\Omega^I$ and a boundary domain $\Omega^B$ such that $\Omega^B \cap \Omega^I = \emptyset$, but assume that all variables are defined on a common regular grid. The boundary states $X_t^B$ are extracted directly from the ground truth on the same grid, rather than coming from an external coarse-resolution global model. Consequently, the conditioning on $B_t$ does not involve handling a different spatial resolution. Since both $\Omega^I$ and $\Omega^B$ lie on the same regular grid, we can use this grid as our mesh. Hence, the grid-to-mesh encoders used in the general setting can be reduced to separate multilayer perceptrons applied pointwise on the interior and boundary inputs. In particular, no graph-based operator is required to map between irregular and regular representations. An overview of the resulting limited-area setup is shown in \cref{fig:model_process}, and additional details are provided in \cref{apx:MEPS}.

\begin{figure}[tbp]
    \centering
        \includegraphics[width=0.8\textwidth]{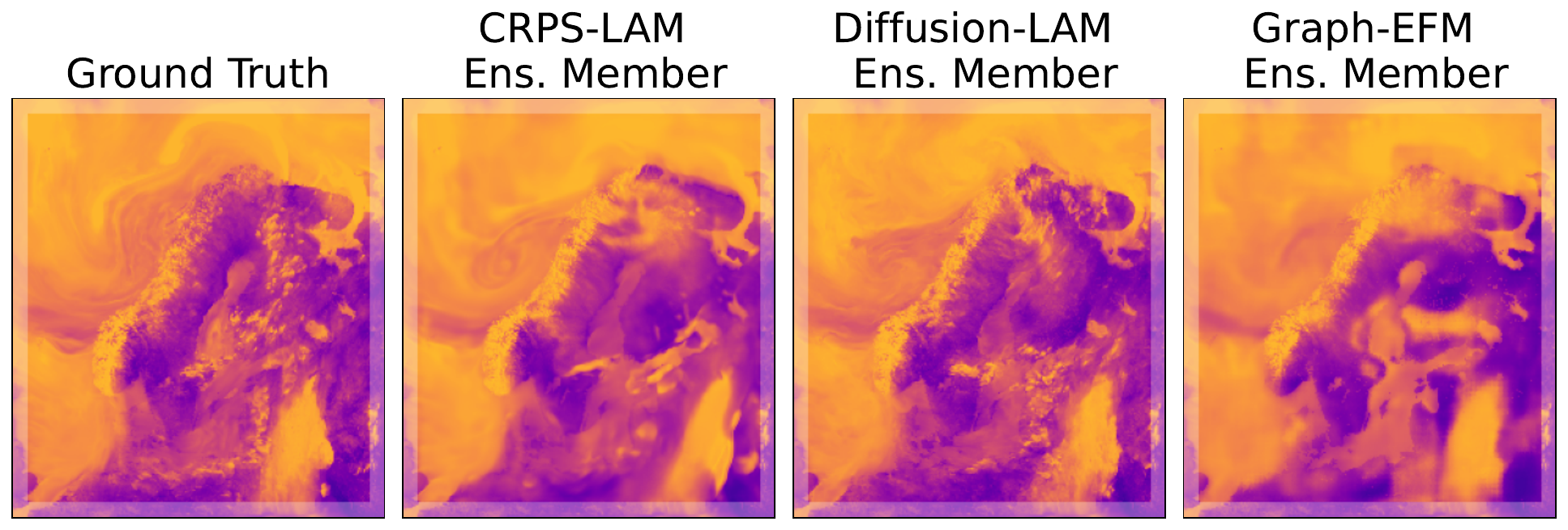}
    \caption{Forecasts at \SI{57}{h} lead time for relative humidity at \SI{2}{\meter} (\texttt{r\_2}). The faded area constitutes the boundary region.}
    \vspace{-1em} 
    \label{fig:meps_forecast}
\end{figure}

\begin{figure}[tbh]
    \centering
    \includegraphics[width=\textwidth, trim={0.3cm 0.3cm 0.3cm 0.3cm}, clip]{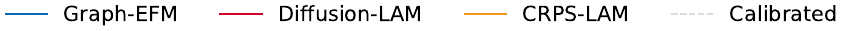}
    \begin{subfigure}[b]{0.3\textwidth}
        \centering
        \includegraphics[width=\textwidth]{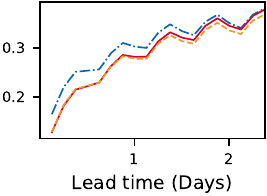}
        \caption{\gls{rmse}}
        \label{fig:subfigure1}
    \end{subfigure}
    \hfill
    \begin{subfigure}[b]{0.3\textwidth}
        \centering
        \includegraphics[width=\textwidth]{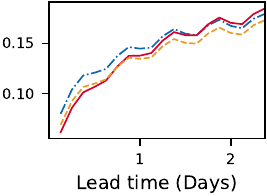}
        \caption{\gls{crps}}
        \label{fig:subfigure1}
    \end{subfigure}
    \hfill
    \begin{subfigure}[b]{0.3\textwidth}
        \centering
        \includegraphics[width=\textwidth]{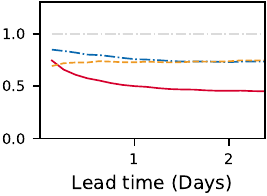}
        \caption{\gls{ssr}}
        \label{fig:subfigure2}
    \end{subfigure}
    \caption{The mean of the normalized \gls{rmse}, \gls{crps}, and \gls{ssr} for all variables on the MEPS dataset.}
    \label{fig:meps_quantitative_results}
\end{figure}

\paragraph{Experiments.}
The models are evaluated on the MEPS \gls{lam} dataset over the Nordic region.
All forecast trajectories are initialized from the ground-truth states and generated autoregressively with \SI{3}{\hour} steps up to a lead time of \SI{57}{\hour}, with 25 ensemble members produced for each model. 
The results show that CRPS-LAM achieves performance competitive with both Diffusion-LAM and Graph-EFM in terms of the sample quality exemplified in \cref{fig:meps_forecast}, as well as pixelwise performance metrics in \cref{fig:meps_quantitative_results}. A well-calibrated ensemble should have $\text{SSR} \approx 1$, yet all models exhibit some degree of underdispersion. CRPS-LAM achieves ensemble calibration comparable to the best-performing model, Graph-EFM, whereas Diffusion-LAM shows stronger underdispersion, particularly at longer lead times. By investigating the spectra of the fields produced by the models in \cref{fig:spectra_all_0,fig:spectra_all_1,fig:spectra_all_2,fig:spectra_all_3} in the appendix, we also note that CRPS-LAM retains more fine-scale details than Graph-EFM, but loses some of the highest frequency components still captured by Diffusion-LAM. This can also be seen in \cref{fig:meps_forecast} and in the more detailed evaluation in \cref{apx:MEPS}.
In this setting, we find that producing ensemble forecasts with the diffusion approach is $\approx 39$
times slower than CRPS-LAM, due to the sequential sampling.
Given the encouraging results of CRPS-LAM on the MEPS data, this validates the approach as an efficient and accurate ensemble forecasting method.
We next scale up the model to a more realistic \gls{lam} setting.

\subsection[Realistic High-Resolution LAM Forecasting: DANRA]{Realistic High-Resolution \gls{lam} Forecasting: DANRA}\label{sec:danra_experiments}
The simplified assumption in the MEPS experiments and in previous works \citep{oskarsson2024probabilistic,larsson2025diffusionlam}, that boundary information $B_t$ is defined on the same grid as the interior states $X_t^I$ and is extracted from ground truth, are not realistic for an operational \gls{lam} setting. In practice, the boundary domain $\Omega^B$ is provided by an external global forecasting system and typically differs from the interior domain $\Omega^I$ in both spatial resolution, grid structure, and possibly time step. As a result, the boundary information $B_t$ is defined on a grid that is not aligned with that of $X_t^I$, and we must use an encoder that can map this representation to a common representation for the processor.

In this section, we consider this more realistic formulation of the problem and demonstrate how our framework accommodates heterogeneous inputs across $\Omega^I$ and $\Omega^B$. In particular, we consider $B_t$ obtained either from a reanalysis dataset or from an operational global forecast, both defined on a different grid from the interior. Additionally, the interior is sampled at \SI{3}{\hour} intervals, whereas the boundary data is available at coarser \SI{6}{\hour} intervals.

\paragraph{DANRA Setting.}
 In this experiment we train the model on the DANRA \citep{DANRA} high-resolution regional reanalysis dataset, aiming to learn a forecasting model for the domain shown in \cref{fig:grid_locations,fig:mesh_position}.
The boundary region extends \SI{400}{\kilo\meter} beyond the interior, and for CRPS-LAM we include boundary grid points overlapping with the interior, meaning that $\Omega^I \subset \Omega^B$.
For a more detailed description of the data we refer the reader to \cref{apx:danra_details} and \citet{adamov2025buildingmachinelearninglimited}.

\paragraph{Interior Data.}
For the interior we use DANRA data covering $1472\times$\SI{1972}{\kilo\meter} at \SI{3}{\hour} time steps and a spatial resolution of \SI{2.5}{\kilo\meter} ($589 \times 789$ pixels). We model 7 surface variable and 6 vertical variables on pressure levels 100, 200, 400, 600, 700, 850, 925, and \SI{1000}{\hecto\pascal}. The state $X^I_t$ has a dimension of 55 at each grid point, where surface variables and pressure levels are concatenated,
resulting in a total dimensionality of 25 million for $X_t^I$.
In addition to the atmospheric states we include the forcing shown in \cref{tab:danra_variables} in the appendix.

\paragraph{Boundary Data.}
We use a boundary region extending \SI{400}{\kilo\meter} beyond the interior, which is deemed sufficient for this setting in an investigation by \citet{adamov2025buildingmachinelearninglimited}. 
To enable laying out the mesh as a regular grid we specify the boundary area as a rectangle in the local coordinate system, meaning that some corner points will be further than \SI{400}{\kilo\meter} away from the interior. For training we use ERA5 \SI{0.25}{\degree} as provided through Weatherbench 2 \citep{rasp2023weatherbench} with \SI{6}{\hour} time steps as boundary conditions.
We evaluate models both with ERA5 boundary forcing and using IFS forecasts, \citep{ecmwf2024ifs} with the same variables and resolutions. 
Note that the boundary data has a different spatial and temporal resolution compared to the interior data.
The spatial mismatch is naturally handled by our hybrid architecture.
For the temporal mismatch we rely on separate boundary encoders and forcing features encoding the exact timestamp \citep{adamov2025buildingmachinelearninglimited}.

\paragraph{Baselines.}
Apart from CRPS-LAM and DET-LAM we consider the deterministic Graph-FM model \citep{adamov2025buildingmachinelearninglimited}, which was developed for this high-resolution DANRA setting.
Graph-FM is a fully graph-based model, and we use the original implementation of \citet{adamov2025buildingmachinelearninglimited} with an \SI{800}{\kilo\meter} boundary region without spatial overlap with the interior.
As an ensemble forecasting baseline, we also train a version of Graph-EFM \citep{oskarsson2024probabilistic} on the DANRA setting. We were unable to scale Diffusion-LAM to a level where it produced competitive results in the DANRA setting, even with the largest possible backbone within the constraints of our hardware. Consequently, we exclude diffusion-based methods from our main set of baselines. This further highlights a practical advantage of CRPS-LAM, which achieves strong performance without the substantial computational demands of diffusion-based approaches. We compare our models also with an operational \gls{nwp} baseline \citep{bengtsson2017}. Unfortunately, the \gls{nwp} forecasts include only a limited set of variables, at lead times up to \SI{18}{\hour} \citep{adamov2025buildingmachinelearninglimited}.

\begin{figure}[tbp]
    \centering
    \includegraphics[width=\textwidth]{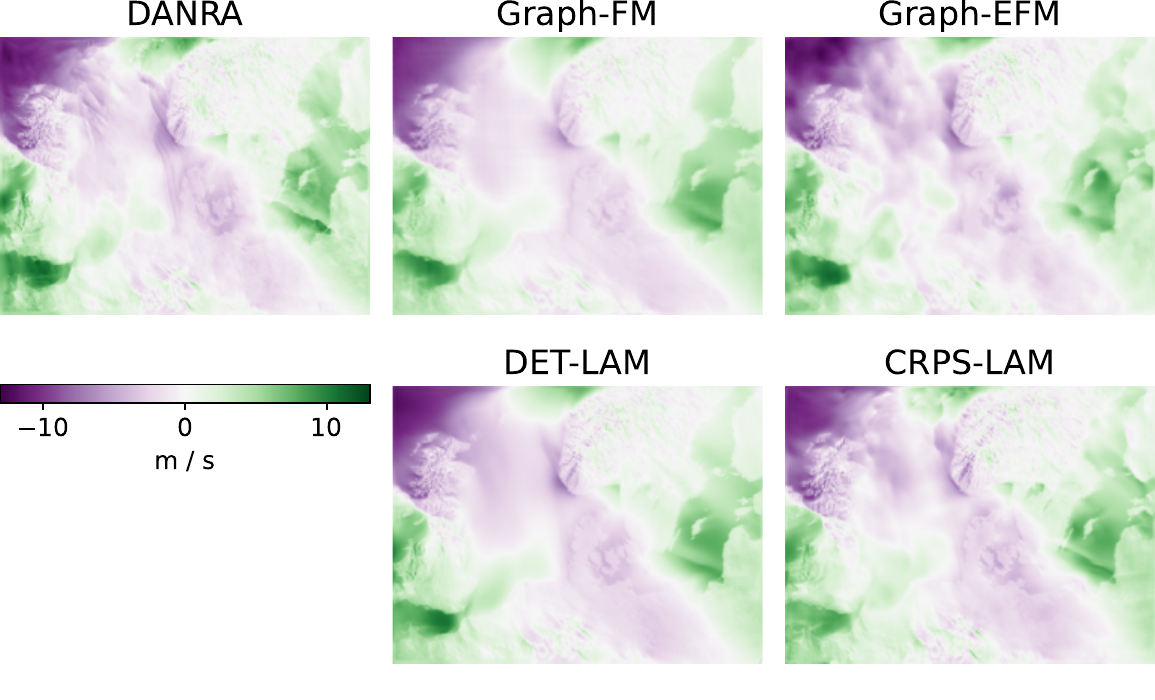}
    \caption{Forecasts at \SI{72}{h} lead time for the u wind component at \SI{10}{\meter} (\texttt{u10m}). For the probabilistic models (Graph-EFM and CRPS-LAM) we show one ensemble member.}
    \vspace{-1.0em} 
    \label{fig:danra_forecast}
\end{figure}
\begin{figure}[tbph]
    \centering
    \includegraphics[width=\textwidth]{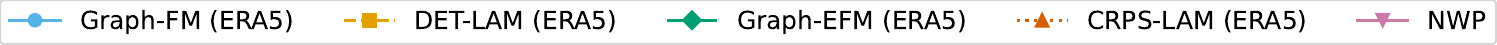}
    \begin{subfigure}[b]{0.33\textwidth}
        \centering
        \includegraphics[width=\textwidth]{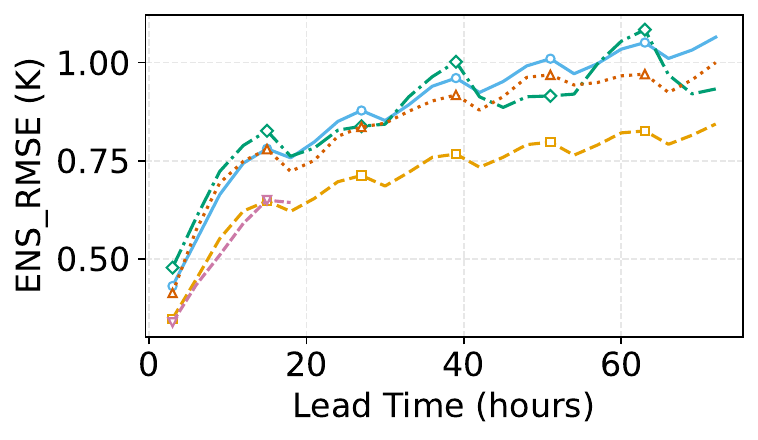}
        \caption{\gls{rmse} for \texttt{t2m}}
        \label{fig:t2m_rmse}
    \end{subfigure}%
    \hfill%
    \begin{subfigure}[b]{0.33\textwidth}
        \centering
        \includegraphics[width=\textwidth]{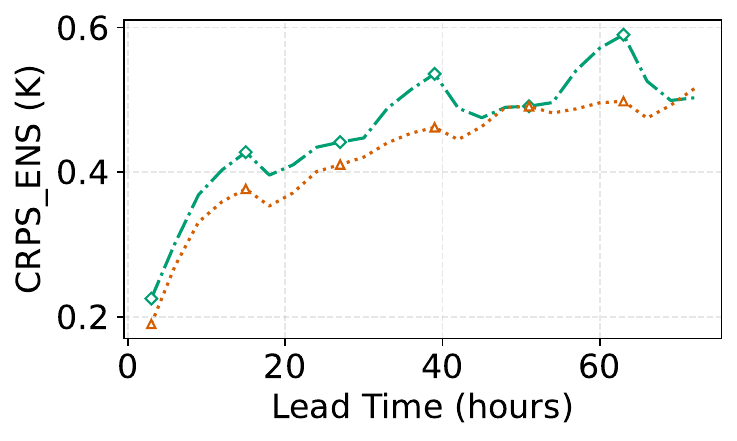}
        \caption{\gls{crps} for \texttt{t2m}}
        \label{fig:t2m_crps}
    \end{subfigure}%
    \hfill%
    \begin{subfigure}[b]{0.33\textwidth}
        \centering
        \includegraphics[width=\textwidth]{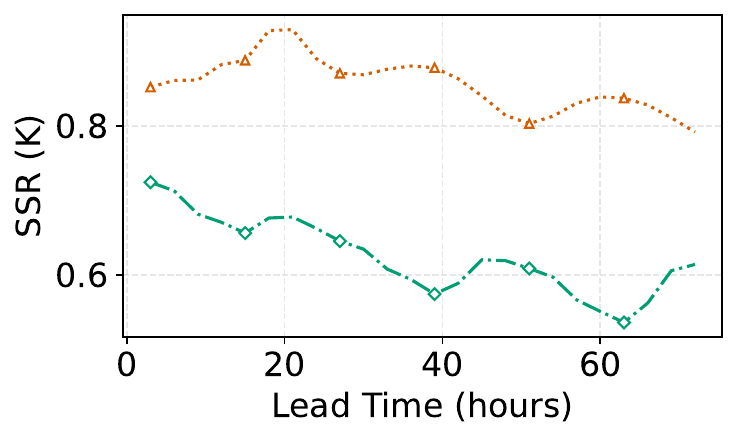}
        \caption{\gls{ssr} for \texttt{t2m}}
        \label{fig:t2m_ssr}
    \end{subfigure}
        \begin{subfigure}[b]{0.33\textwidth}
        \centering
        \includegraphics[width=\textwidth]{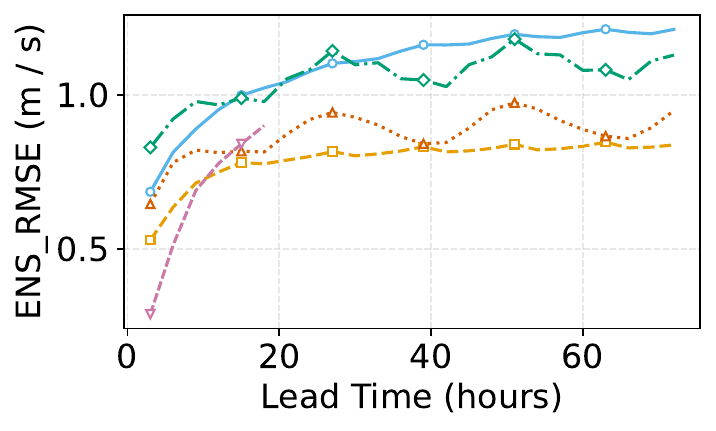}
        \caption{\gls{rmse} for \texttt{u10m}}
        \label{fig:u10m_rmse}
    \end{subfigure}%
    \hfill%
    \begin{subfigure}[b]{0.33\textwidth}
        \centering
        \includegraphics[width=\textwidth]{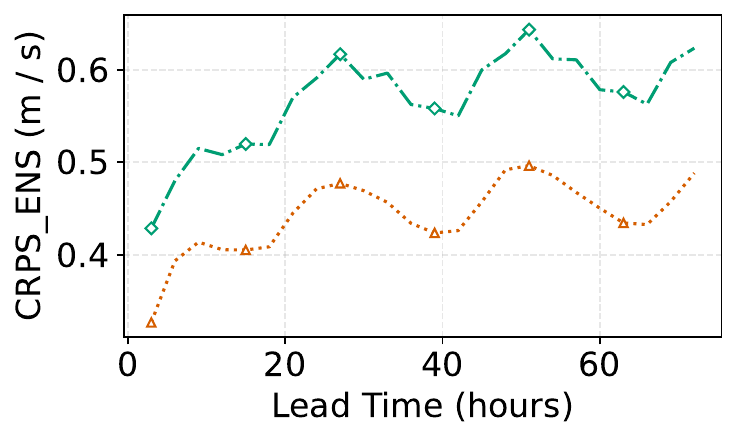}
        \caption{\gls{crps} for \texttt{u10m}}
        \label{fig:u10m_crps}
    \end{subfigure}%
    \hfill%
    \begin{subfigure}[b]{0.33\textwidth}
        \centering
        \includegraphics[width=\textwidth]{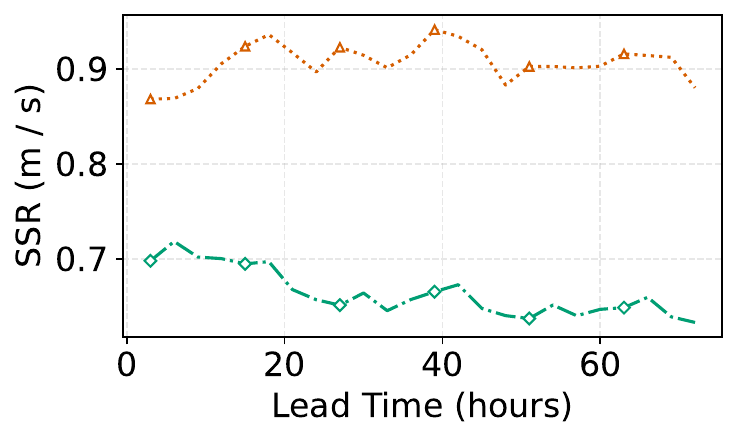}
        \caption{\gls{ssr} for \texttt{u10m}}
        \label{fig:u10m_ssr}
    \end{subfigure}%
    \caption{The \gls{rmse}, \gls{crps}, and \gls{ssr} for \SI{2}{\meter} temperature and the u wind component at \SI{10}{\meter} on the DANRA dataset with ERA5 boundary conditions.
    Note that the machine learning models have access to future information through the ERA5 boundary forcing, giving a major advantage compared to the \gls{nwp} model. The more fair comparison is in \cref{fig:quantitative_results_danra_IFS}.
    }
    \label{fig:quantitative_results_danra_ERA5}
\end{figure}

\paragraph{Gridded Evaluation Against Reanalysis.}
The forecast skill of the different models is first evaluated with ERA5 boundary forcing, corresponding to accurate (but coarse) boundary information. 
Then we replace this with imperfect IFS forecasts, and evaluate the models also in this setting.
We assess how well the models capture the mean of the predictive distribution using the \gls{rmse}. 
The \gls{rmse} is computed directly from the predictions for deterministic models, and from the ensemble mean for probabilistic ones.
As shown in \cref{fig:quantitative_results_danra_ERA5}, DET-LAM and CRPS-LAM consistently match or outperform their respective deterministic and probabilistic baselines, Graph-FM and Graph-EFM.
We do not necessarily expect the probabilistic models to achieve lower \gls{rmse} values, as this metric is directly optimized by the MSE objective used to train the deterministic models. In principle, the RMSE gap should decrease as the number of ensemble members increases, since the ensemble mean becomes a more accurate estimate of the predictive mean. However, if the primary objective is to maximize the skill of the mean forecast, a deterministic model may be the more appropriate choice. In contrast, when the goal is to characterize forecast uncertainty and provide probabilistic predictions, metrics such as \gls{crps} and \gls{ssr} are more informative. These metrics evaluate the quality and calibration of the full predictive distribution rather than only the accuracy of its mean, making them more relevant for assessing ensemble forecasting systems.

As shown in \cref{fig:quantitative_results_danra_ERA5}, CRPS-LAM remains slightly underdispersed, but its \gls{ssr} values are consistently closer to $1$ than those of Graph-EFM, indicating improved calibration and a better balance between ensemble spread and forecast skill. For the experiments using the IFS data, \cref{fig:quantitative_results_danra_IFS} shows that the \gls{ssr} decreases with increasing lead time. This behavior is not entirely unexpected given our experimental setup. Maintaining calibration across the full forecast horizon is challenging, since the models are trained on relatively short rollouts initialized from reanalysis data. At longer lead times, both the initial conditions and the boundary forcing become increasingly uncertain, introducing a distribution shift that is not explicitly represented during training.
We also assess the quality of the full predictive distribution using the \gls{crps}, computed between the ensemble forecasts and the DANRA reanalysis. In \cref{fig:quantitative_results_danra_ERA5,fig:quantitative_results_danra_IFS} we see that CRPS-LAM consistently outperforms or matches Graph-EFM in terms of CRPS across all lead times. This suggests that CRPS-LAM better captures the predictive distribution.

\begin{figure}[tbph]
    \centering
    \includegraphics[width=\textwidth]{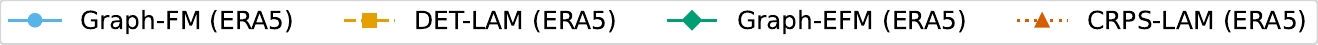}
    \begin{subfigure}[b]{0.33\textwidth}
        \centering
        \includegraphics[width=\textwidth]{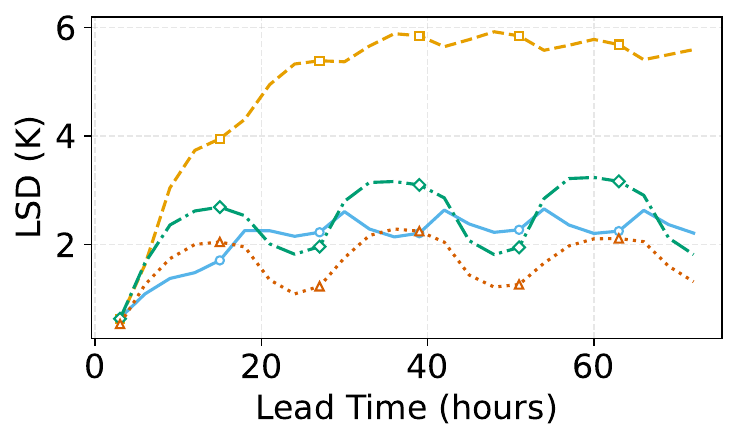}
        \caption{LSD for \texttt{t2m}}
        \label{fig:subfigure1}
    \end{subfigure}%
    \hfill%
    \begin{subfigure}[b]{0.33\textwidth}
        \centering
        \includegraphics[width=\textwidth]{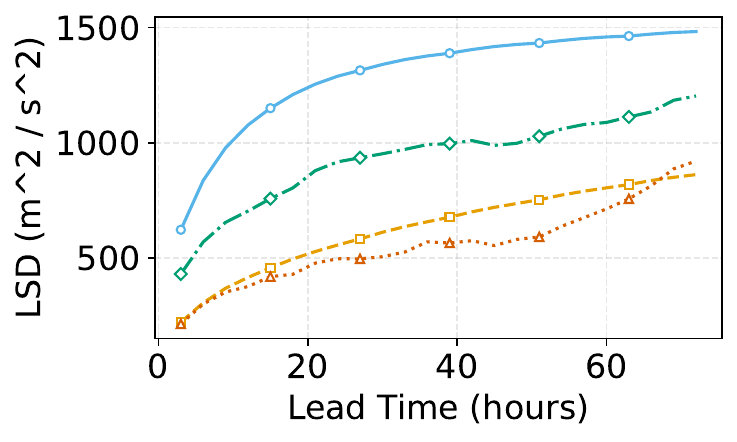}
        \caption{LSD for \texttt{z600}}
        \label{fig:subfigure1}
    \end{subfigure}%
    \hfill%
    \begin{subfigure}[b]{0.33\textwidth}
        \centering
        \includegraphics[width=\textwidth]{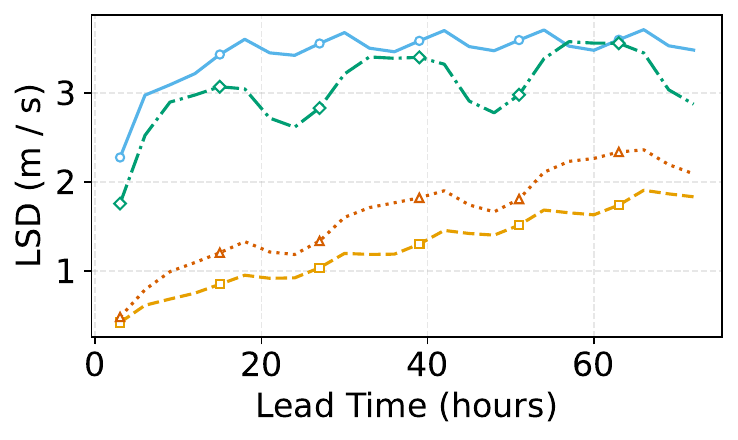}
        \caption{LSD for \texttt{u10m}}
        \label{fig:u10m_lsd}
    \end{subfigure}%
    \caption{The \gls{lsd} of the \SI{2}{\meter} temperature, geopotential at \SI{600}{\hecto\pascal}, and u wind component at \SI{10}{\meter} on the DANRA dataset with ERA5 boundary conditions.}
    \label{fig:spectra_danra}
    \vspace{-1.0em}
\end{figure}

We also evaluate the spatial structure over time using the \gls{lsd} at each forecast lead time. As shown in \cref{fig:spectra_danra}, DET-LAM and CRPS-LAM generally achieve lower \gls{lsd} than Graph-EFM, indicating better preservation of spatial structure. 
Graph-FM shows the largest difference, over-smoothing many of the detailed fields.
For \SI{2}{\meter} temperature, however, the deterministic DET-LAM model is outperformed by the probabilistic models, Graph-EFM and CRPS-LAM. Among these, CRPS-LAM still achieves the best performance, consistently yielding lower \gls{lsd} than Graph-EFM. We also note that, as shown in \cref{apx:danra_forecasts}, Graph-EFM occasionally exhibits visual artifacts in both its predictions and estimated standard deviations, potentially due to its graph-based architecture \citep{oskarsson2024probabilistic}.

\begin{figure}[tbph]
    \centering%
    \includegraphics[width=\textwidth]{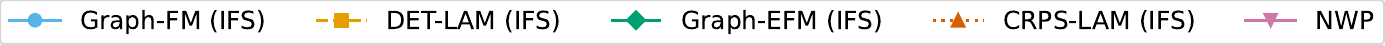}
    \begin{subfigure}[b]{0.33\textwidth}
        \centering
        \includegraphics[width=\textwidth]{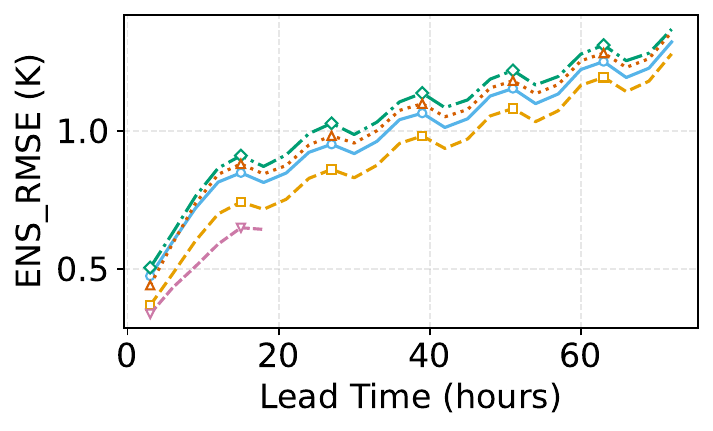}
        \caption{\gls{rmse} for \texttt{t2m}}
        \label{fig:t2m_rmse}
    \end{subfigure}%
    \hfill%
    \begin{subfigure}[b]{0.33\textwidth}
        \centering
        \includegraphics[width=\textwidth]{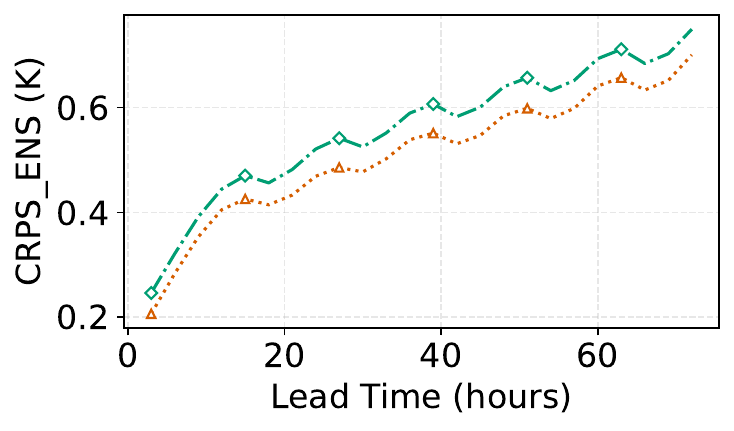}
        \caption{\gls{crps} for \texttt{t2m}}
        \label{fig:t2m_crps}
    \end{subfigure}%
    \hfill%
    \begin{subfigure}[b]{0.33\textwidth}
        \centering
        \includegraphics[width=\textwidth]{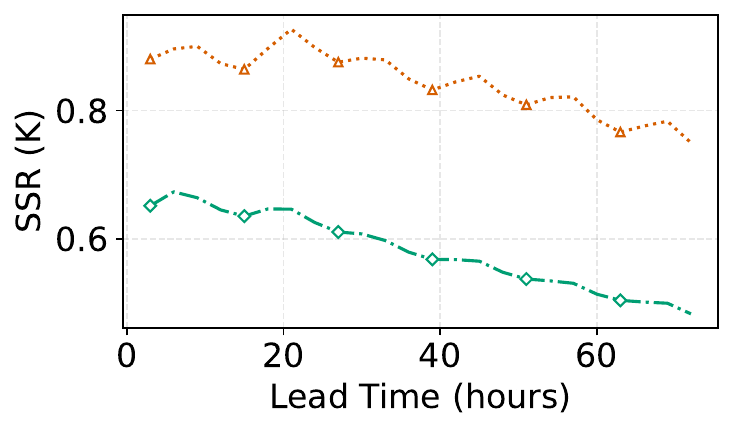}
        \caption{\gls{ssr} for \texttt{t2m}}
        \label{fig:t2m_ssr}
    \end{subfigure}
    \begin{subfigure}[b]{0.33\textwidth}
        \centering
        \includegraphics[width=\textwidth]{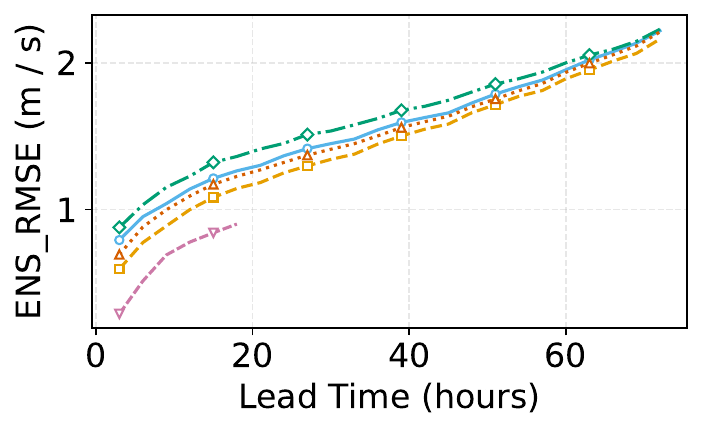}
        \caption{\gls{rmse} for \texttt{u10m}}
        \label{fig:u10m_rmse}
    \end{subfigure}%
    \hfill%
    \begin{subfigure}[b]{0.33\textwidth}
        \centering
        \includegraphics[width=\textwidth]{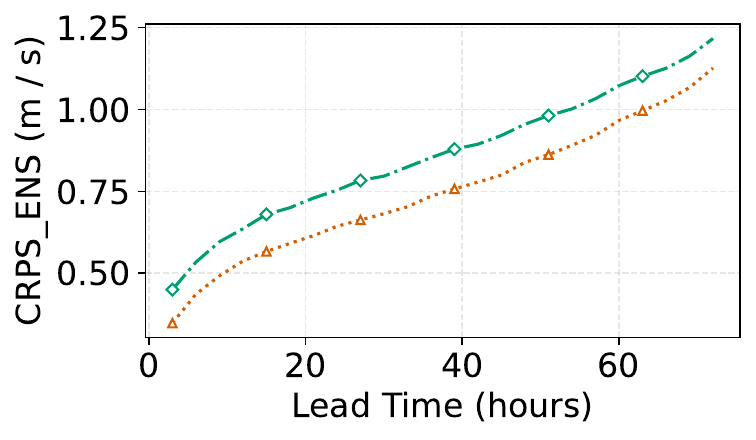}
        \caption{\gls{crps} for \texttt{u10m}}
        \label{fig:u10m_crps}
    \end{subfigure}%
    \hfill%
    \begin{subfigure}[b]{0.33\textwidth}
        \centering
        \includegraphics[width=\textwidth]{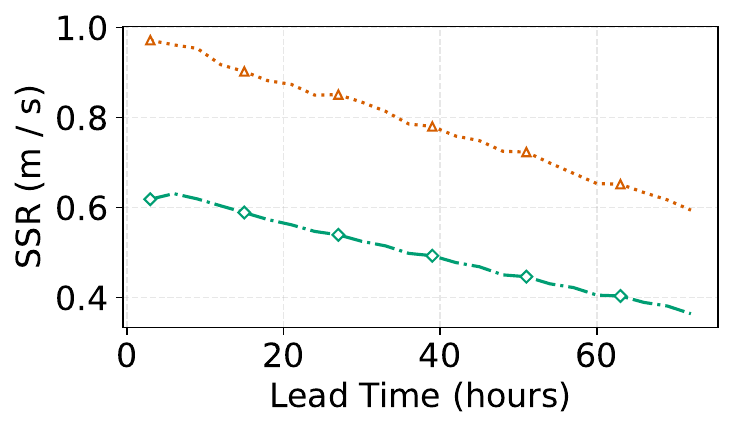}
        \caption{\gls{ssr} for \texttt{u10m}}
        \label{fig:u10m_ssr}
    \end{subfigure}%
    \caption{The \gls{rmse}, \gls{crps}, and \gls{ssr} for \SI{2}{\meter} temperature and the u wind component at \SI{10}{\meter} on the DANRA dataset with IFS boundary for the models finetuned with IFS boundary conditions.}
    \label{fig:quantitative_results_danra_IFS_finetune}
\end{figure}

When replacing the boundary conditions with IFS forecasts, the performance differences between the models become less pronounced at longer lead times, and all models exhibit higher errors. However, we still see a clear, consistent benefit of DET-LAM and CRPS-LAM over Graph-FM and Graph-EFM, for the 1-step \SI{3}{\hour} forecast. This suggests that the models may not be well-adapted to this setup and could likely benefit from additional fine-tuning using IFS boundary conditions. To mitigate this underdispersion and better adapt the models to the IFS boundary data, we performed a final fine-tuning step using the IFS dataset. The results after fine-tuning are shown in \cref{fig:quantitative_results_danra_IFS_finetune}. Compared to the results in \cref{fig:quantitative_results_danra_IFS,fig:quantitative_results_danra_finetuing_boundary_ablation}, both the predictive skill and the ensemble calibration improved, demonstrating the effectiveness of the fine-tuning procedure. While performance improves, it still falls short of the results achieved with ERA5 forcing. This highlights the challenge of learning a mapping to the next time step when the reliability of boundary information degrades at longer lead times. Ultimately, the ERA5 boundary conditions provide an upper bound for \gls{lam} performance under optimal conditions. This setup simplifies the problem, as ERA5 boundary conditions remain consistently reliable across all lead times, unlike the IFS forcing.

\begin{figure}[tbph]
    \centering%
    \includegraphics[width=\textwidth]{figures/legend_ifs_nwp.pdf}
    \begin{subfigure}[b]{0.33\textwidth}
        \centering
        \includegraphics[width=\textwidth]{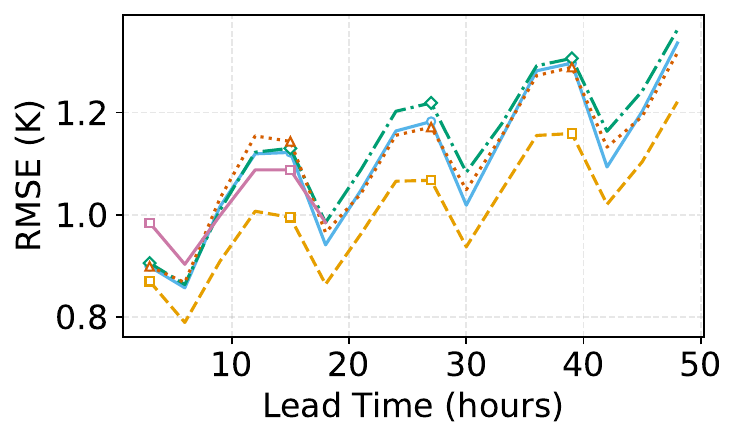}
        \caption{\gls{rmse} for \texttt{t2m}}
        \label{fig:t2m_rmse}
    \end{subfigure}%
    \hfill%
    \begin{subfigure}[b]{0.33\textwidth}
        \centering
        \includegraphics[width=\textwidth]{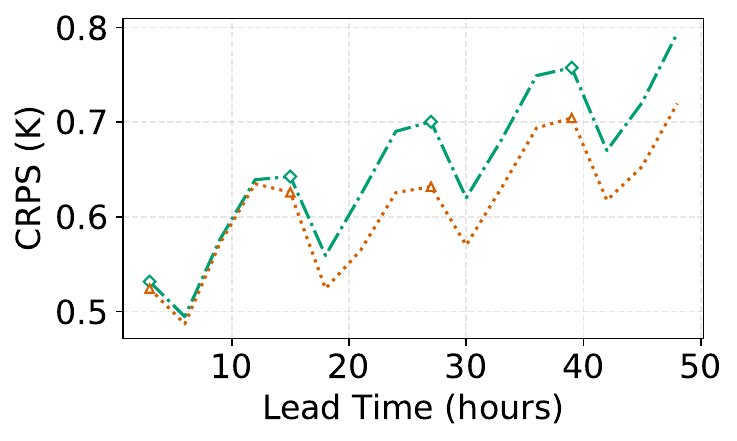}
        \caption{\gls{crps} for \texttt{t2m}}
        \label{fig:t2m_crps}
    \end{subfigure}%
    \hfill%
    \begin{subfigure}[b]{0.33\textwidth}
        \centering
        \includegraphics[width=\textwidth]{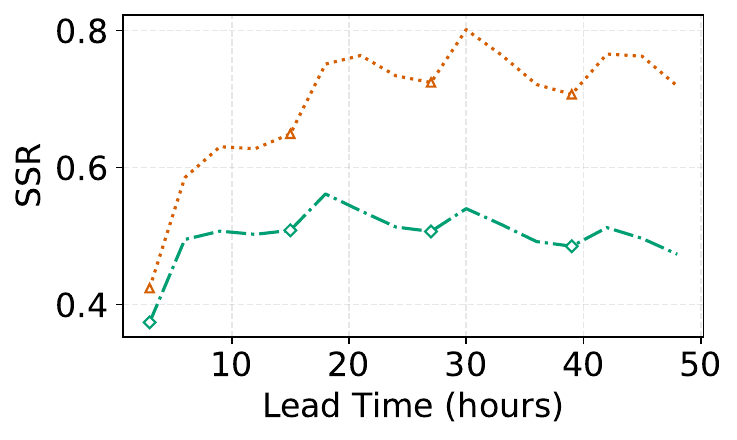}
        \caption{\gls{ssr} for \texttt{t2m}}
        \label{fig:t2m_ssr}
    \end{subfigure}
    \begin{subfigure}[b]{0.33\textwidth}
        \centering
        \includegraphics[width=\textwidth]{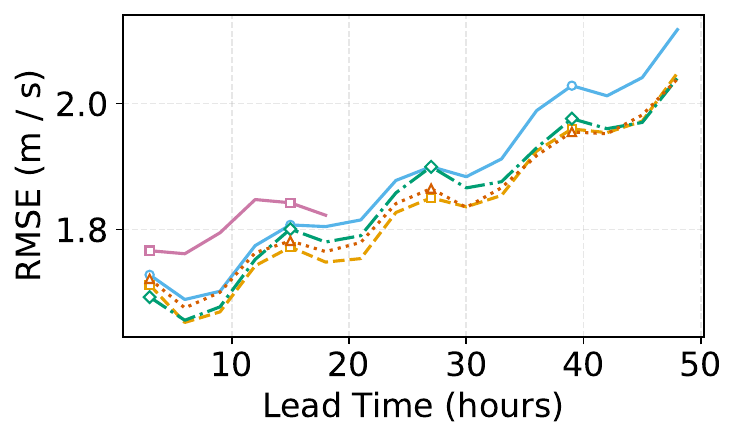}
        \caption{\gls{rmse} for \texttt{u10m}}
        \label{fig:u10m_rmse}
    \end{subfigure}%
    \hfill%
    \begin{subfigure}[b]{0.33\textwidth}
        \centering
        \includegraphics[width=\textwidth]{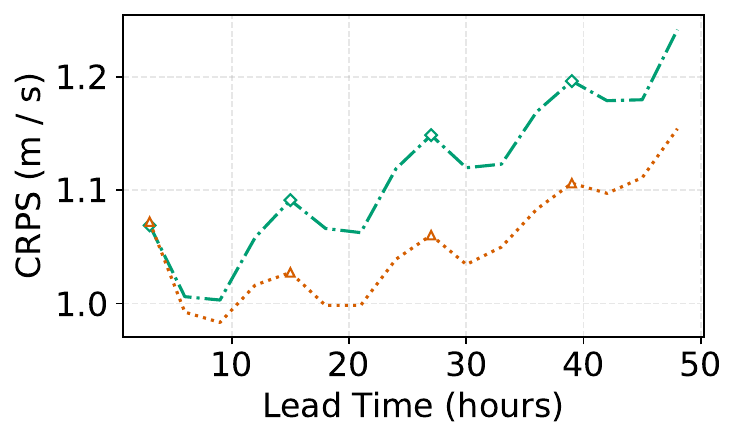}
        \caption{\gls{crps} for \texttt{u10m}}
        \label{fig:u10m_crps}
    \end{subfigure}%
    \hfill%
    \begin{subfigure}[b]{0.33\textwidth}
        \centering
        \includegraphics[width=\textwidth]{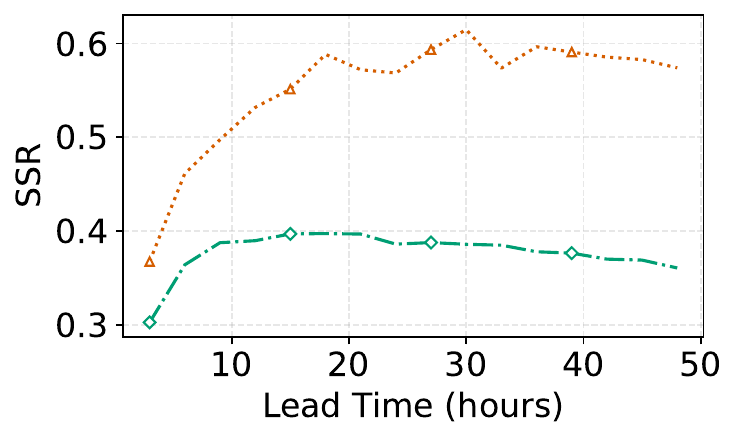}
        \caption{\gls{ssr} for \texttt{u10m}}
        \label{fig:u10m_ssr}
    \end{subfigure}%
    \caption{Results on the DANRA dataset with IFS boundary for the models finetuned with IFS boundary conditions compared against in-situ observations.}
    \label{fig:quantitative_results_danra_IFS_finetune_obs}
\end{figure}

\paragraph{Evaluation against in-situ observations.}
We further evaluate model performance against in-situ observations from weather stations in Denmark, at the center of the DANRA domain. We use measurements of \SI{2}{\meter} temperature and \SI{10}{\meter} wind from 50 stations over the test period, and compute the \gls{rmse}, \gls{ssr}, and \gls{crps} at these locations (see \cref{fig:station_locations}). As shown in \cref{fig:quantitative_results_danra_IFS_finetune_obs}, the \gls{mlwp} models are competitive with the \gls{nwp} model in terms of \gls{rmse}. This is particularly evident for the \SI{10}{\meter} u-wind component and for short-range forecasts (up to \SI{6}{\hour}) of \SI{2}{\meter} temperature. In terms of \gls{crps}, CRPS-LAM consistently outperforms Graph-EFM, except at shorter lead times where the two models exhibit very similar performance. The \gls{ssr} results indicate that neither model is well calibrated with respect to the observations. This is expected, as the observational distribution differs from the reanalysis data used as training target. 
The shift between reanalysis and observations leads to an increase in error, but the models are not trained to match this with an increase in spread.
This calibration gap could be reduced by finetuning with model outputs directly compared to observations.


\section{Discussion}\label{sec:discussion}
We introduce a new backbone architecture for \gls{lam} weather forecasting, combining \gls{gnn} and \gls{cnn} components.
This hybridization substantially reduces the memory footprint, allowing the model to scale more effectively and achieve improved performance.
Building on this backbone, we propose two models: DET-LAM, a deterministic forecasting model, and CRPS-LAM, a probabilistic forecasting model that produces skillful, well-calibrated ensemble forecasts at inference speeds an order of magnitude faster than flow-based models.
Experiments on high-resolution forecasting show how our approach achieves lower errors, and how the generative CRPS-LAM method produces more realistic fields.

\paragraph{Limitations.}
In our DANRA experiments, the ML models are initialized from DANRA reanalysis states rather than operational analysis states, giving them a slight advantage over the NWP baseline, though comparisons between ML models remain fair. We also restrict the models to a subset of atmospheric variables due to data availability and prior work \citep{adamov2025buildingmachinelearninglimited}; extending to a broader variable set is left for future work.

\paragraph{Future Work.}
In this work, we consider only two sources of boundary forcing, ERA5 and IFS. A direction for future work is to investigate the effects of using alternative sources of boundary information, including ML-based and ensemble forecasts, and to determine suitable adaptations or fine-tuning strategies in such settings. It would also be valuable to investigate cross-domain transfer, particularly whether models can generalize zero-shot or with minimal fine-tuning in regions where high-quality reanalysis data is limited.

\begin{ack}
This research is financially supported by the Swedish Research Council (grant no:
2024-05011) the Wallenberg AI, Autonomous Systems and Software Program (WASP) funded
by the Knut and Alice Wallenberg Foundation, and the Excellence Center at Linköping–Lund in
Information Technology (ELLIIT). 
This research was supported by the ETH AI Center through an ETH AI Center postdoctoral fellowship to Joel Oskarsson.
Our computations were enabled by the Berzelius resource at the
National Supercomputer Centre, provided by the Knut and Alice Wallenberg Foundation. We thank Xuan Gu at the National Supercomputer Centre for assistance with technical and implementation aspects of running the code on the multi-node Berzelius system, and Simon Adamov at ETHZ/MeteoSwiss for assistance with the Graph-FM baseline results. Landelius was funded by the Swedish Energy Agency and the European Union’s Horizon 2020 research and
innovation programme under grant agreement no. 883973–EnerDigit as well as by the Swedish Foundation for Strategic Research.
\end{ack}

\bibliographystyle{abbrvnat} 
\bibliography{references}

\begin{thebibliography}{43}
\providecommand{\natexlab}[1]{#1}
\providecommand{\url}[1]{\texttt{#1}}
\expandafter\ifx\csname urlstyle\endcsname\relax
  \providecommand{\doi}[1]{doi: #1}\else
  \providecommand{\doi}{doi: \begingroup \urlstyle{rm}\Url}\fi

\bibitem[Abdi et~al.(2026)Abdi, Jankov, Madden, Vargas, Smith, Frolov, Flora, and Potvin]{hrrrcast}
D.~Abdi, I.~Jankov, P.~Madden, V.~Vargas, T.~A. Smith, S.~Frolov, M.~Flora, and C.~Potvin.
\newblock Hrrrcast: A data-driven emulator for regional weather forecasting at convection-allowing scales.
\newblock \emph{Artificial Intelligence for the Earth Systems}, 5\penalty0 (2):\penalty0 250061, 2026.
\newblock \doi{10.1175/AIES-D-25-0061.1}.

\bibitem[Adamov et~al.(2025)Adamov, Oskarsson, Denby, Landelius, Hintz, Christiansen, Schicker, Osuna, Lindsten, Fuhrer, and Schemm]{adamov2025buildingmachinelearninglimited}
S.~Adamov, J.~Oskarsson, L.~Denby, T.~Landelius, K.~Hintz, S.~Christiansen, I.~Schicker, C.~Osuna, F.~Lindsten, O.~Fuhrer, and S.~Schemm.
\newblock Building machine learning limited area models: Kilometer-scale weather forecasting in realistic settings.
\newblock \emph{arXiv preprint arXiv:2504.09340}, 2025.
\newblock URL \url{https://arxiv.org/abs/2504.09340}.

\bibitem[Alet et~al.(2025)Alet, Price, El-Kadi, Masters, Markou, Andersson, Stott, Lam, Willson, Sanchez-Gonzalez, and Battaglia]{FGNalet2025skillfuljointprobabilisticweather}
F.~Alet, I.~Price, A.~El-Kadi, D.~Masters, S.~Markou, T.~R. Andersson, J.~Stott, R.~Lam, M.~Willson, A.~Sanchez-Gonzalez, and P.~Battaglia.
\newblock Skillful joint probabilistic weather forecasting from marginals.
\newblock \emph{arXiv preprint arXiv:2506.10772}, 2025.
\newblock URL \url{https://arxiv.org/abs/2506.10772}.

\bibitem[Andrae et~al.(2025)Andrae, Landelius, Oskarsson, and Lindsten]{andrae2024continuousensembleweatherforecasting}
M.~Andrae, T.~Landelius, J.~Oskarsson, and F.~Lindsten.
\newblock Continuous ensemble weather forecasting with diffusion models.
\newblock In \emph{International Conference on Learning Representations}, 2025.

\bibitem[Battaglia et~al.(2018)Battaglia, Hamrick, Bapst, Sanchez-Gonzalez, Zambaldi, Malinowski, Tacchetti, Raposo, Santoro, Faulkner, et~al.]{battaglia2018relational}
P.~W. Battaglia, J.~B. Hamrick, V.~Bapst, A.~Sanchez-Gonzalez, V.~Zambaldi, M.~Malinowski, A.~Tacchetti, D.~Raposo, A.~Santoro, R.~Faulkner, et~al.
\newblock Relational inductive biases, deep learning, and graph networks.
\newblock \emph{arXiv preprint arXiv:1806.01261}, 2018.

\bibitem[Bengtsson et~al.(2017)Bengtsson, Andrae, Aspelien, Batrak, Calvo, de~Rooy, Gleeson, Hansen-Sass, Homleid, Hortal, Ivarsson, Lenderink, Niemelä, Nielsen, Onvlee, Rontu, Samuelsson, Muñoz, Subias, Tijm, Toll, Yang, and Ødegaard Køltzow]{bengtsson2017}
L.~Bengtsson, U.~Andrae, T.~Aspelien, Y.~Batrak, J.~Calvo, W.~de~Rooy, E.~Gleeson, B.~Hansen-Sass, M.~Homleid, M.~Hortal, K.-I. Ivarsson, G.~Lenderink, S.~Niemelä, K.~P. Nielsen, J.~Onvlee, L.~Rontu, P.~Samuelsson, D.~S. Muñoz, A.~Subias, S.~Tijm, V.~Toll, X.~Yang, and M.~Ødegaard Køltzow.
\newblock The harmonie–arome model configuration in the aladin–hirlam nwp system.
\newblock \emph{Monthly Weather Review}, 145\penalty0 (5):\penalty0 1919 -- 1935, 2017.
\newblock \doi{10.1175/MWR-D-16-0417.1}.
\newblock URL \url{https://journals.ametsoc.org/view/journals/mwre/145/5/mwr-d-16-0417.1.xml}.

\bibitem[Bi et~al.(2023)Bi, Xie, Zhang, Chen, Gu, and Tian]{pangu}
K.~Bi, L.~Xie, H.~Zhang, X.~Chen, X.~Gu, and Q.~Tian.
\newblock Accurate medium-range global weather forecasting with 3d neural networks.
\newblock \emph{Nature}, 619\penalty0 (7970):\penalty0 533--538, 2023.

\bibitem[Bonev et~al.(2025)Bonev, Kurth, Mahesh, Bisson, Kossaifi, Kashinath, Anandkumar, Collins, Pritchard, and Keller]{fourcastnet3}
B.~Bonev, T.~Kurth, A.~Mahesh, M.~Bisson, J.~Kossaifi, K.~Kashinath, A.~Anandkumar, W.~D. Collins, M.~S. Pritchard, and A.~Keller.
\newblock {FourCastNet} 3: A geometric approach to probabilistic machine-learning weather forecasting at scale.
\newblock \emph{arXiv preprint arXiv:2507.12144}, \penalty0 ({arXiv}:2507.12144), 2025.
\newblock \doi{10.48550/arXiv.2507.12144}.
\newblock URL \url{http://arxiv.org/abs/2507.12144}.

\bibitem[Chen et~al.(2021)Chen, Tan, Li, Liu, Qin, Zhao, and Liu]{chen2021adaspeechadaptivetextspeech}
M.~Chen, X.~Tan, B.~Li, Y.~Liu, T.~Qin, S.~Zhao, and T.-Y. Liu.
\newblock Adaspeech: Adaptive text to speech for custom voice.
\newblock \emph{arXiv preprint arXiv:2103.00993}, 2021.
\newblock URL \url{https://arxiv.org/abs/2103.00993}.

\bibitem[Couairon et~al.(2026)Couairon, Singh, Charantonis, Lessig, and Monteleoni]{couairon2024archesweatherarchesweathergendeterministic}
G.~Couairon, R.~Singh, A.~Charantonis, C.~Lessig, and C.~Monteleoni.
\newblock Archesweathergen: Skillful and compute-efficient probabilistic weather forecasting with machine learning.
\newblock \emph{Science Advances}, 12\penalty0 (17):\penalty0 eadx2372, 2026.
\newblock \doi{10.1126/sciadv.adx2372}.
\newblock URL \url{https://www.science.org/doi/abs/10.1126/sciadv.adx2372}.

\bibitem[{ECMWF}(2024)]{ecmwf2024ifs}
{ECMWF}.
\newblock Integrated forecasting system.
\newblock \url{https://www.ecmwf.int/en/forecasts/documentation-and-support/changes-ecmwf-model}, 2024.
\newblock Accessed: 2026-05-07.

\bibitem[Ferro(2014)]{fairCRPS}
C.~A.~T. Ferro.
\newblock Fair scores for ensemble forecasts.
\newblock \emph{Quarterly Journal of the Royal Meteorological Society}, 140\penalty0 (683):\penalty0 1917--1923, 2014.
\newblock \doi{https://doi.org/10.1002/qj.2270}.
\newblock URL \url{https://rmets.onlinelibrary.wiley.com/doi/abs/10.1002/qj.2270}.

\bibitem[Fortin et~al.(2014)Fortin, Abaza, Anctil, and Turcotte]{fortin2014should}
V.~Fortin, M.~Abaza, F.~Anctil, and R.~Turcotte.
\newblock Why should ensemble spread match the rmse of the ensemble mean?
\newblock \emph{Journal of Hydrometeorology}, 15\penalty0 (4):\penalty0 1708--1713, 2014.

\bibitem[Gneiting and Raftery(2007)]{gneiting2007strictly}
T.~Gneiting and A.~E. Raftery.
\newblock Strictly proper scoring rules, prediction, and estimation.
\newblock \emph{Journal of the American Statistical Association}, pages 359--378, 2007.

\bibitem[Hu et~al.(2023)Hu, Chen, Wang, and Li]{SwinVRNN}
Y.~Hu, L.~Chen, Z.~Wang, and H.~Li.
\newblock Swinvrnn: A data‐driven ensemble forecasting model via learned distribution perturbation.
\newblock \emph{Journal of Advances in Modeling Earth Systems}, 15, 02 2023.
\newblock \doi{10.1029/2022MS003211}.

\bibitem[Karras et~al.(2022)Karras, Aittala, Aila, and Laine]{karras2022elucidating}
T.~Karras, M.~Aittala, T.~Aila, and S.~Laine.
\newblock Elucidating the design space of diffusion-based generative models.
\newblock In S.~Koyejo, S.~Mohamed, A.~Agarwal, D.~Belgrave, K.~Cho, and A.~Oh, editors, \emph{Advances in Neural Information Processing Systems}, volume~35, pages 26565--26577, 2022.

\bibitem[Lam et~al.(2023)Lam, Sanchez-Gonzalez, Willson, Wirnsberger, Fortunato, Alet, Ravuri, Ewalds, Eaton-Rosen, Hu, et~al.]{graphcast}
R.~Lam, A.~Sanchez-Gonzalez, M.~Willson, P.~Wirnsberger, M.~Fortunato, F.~Alet, S.~Ravuri, T.~Ewalds, Z.~Eaton-Rosen, W.~Hu, et~al.
\newblock Learning skillful medium-range global weather forecasting.
\newblock \emph{Science}, 382\penalty0 (6677):\penalty0 1416--1421, 2023.

\bibitem[Lang et~al.(2024)Lang, Alexe, Chantry, Dramsch, Pinault, Raoult, Clare, Lessig, Maier-Gerber, Magnusson, Bouallègue, Nemesio, Dueben, Brown, Pappenberger, and Rabier]{lang2024aifsecmwfsdatadriven}
S.~Lang, M.~Alexe, M.~Chantry, J.~Dramsch, F.~Pinault, B.~Raoult, M.~C.~A. Clare, C.~Lessig, M.~Maier-Gerber, L.~Magnusson, Z.~B. Bouallègue, A.~P. Nemesio, P.~D. Dueben, A.~Brown, F.~Pappenberger, and F.~Rabier.
\newblock {AIFS -- ECMWF}'s data-driven forecasting system.
\newblock \emph{arXiv preprint arXiv:2406.01465}, 2024.
\newblock URL \url{https://arxiv.org/abs/2406.01465}.

\bibitem[Lang et~al.(2025)Lang, Leutbecher, and Maciel]{lang_multi-scale}
S.~Lang, M.~Leutbecher, and P.~Maciel.
\newblock A multi-scale loss formulation for learning a probabilistic model with proper score optimisation.
\newblock \emph{arXiv preprint arXiv:2506.10868}, \penalty0 ({arXiv}:2506.10868), 2025.
\newblock \doi{10.48550/arXiv.2506.10868}.
\newblock URL \url{http://arxiv.org/abs/2506.10868}.

\bibitem[Lang et~al.(2026)Lang, Alexe, Clare, Roberts, Adewoyin, Ben~Bouall{\`e}gue, Chantry, Dramsch, Dueben, Hahner, et~al.]{lang2024aifscrpsensembleforecastingusing}
S.~Lang, M.~Alexe, M.~C. Clare, C.~Roberts, R.~Adewoyin, Z.~Ben~Bouall{\`e}gue, M.~Chantry, J.~Dramsch, P.~D. Dueben, S.~Hahner, et~al.
\newblock {AIFS-CRPS}: ensemble forecasting using a model trained with a loss function based on the continuous ranked probability score.
\newblock \emph{npj Artificial Intelligence}, 2\penalty0 (1):\penalty0 18, 2026.

\bibitem[Larsson et~al.(2025)Larsson, Oskarsson, Landelius, and Lindsten]{larsson2025diffusionlam}
E.~Larsson, J.~Oskarsson, T.~Landelius, and F.~Lindsten.
\newblock {Diffusion-LAM}: Probabilistic limited area weather forecasting with diffusion.
\newblock In \emph{ICLR 2025 Workshop on Tackling Climate Change with Machine Learning}, 2025.
\newblock URL \url{https://www.climatechange.ai/papers/iclr2025/36}.

\bibitem[Larsson et~al.(2026)Larsson, Fuentes-Franco, Ivanov, and Lindsten]{larsson2026climatedownscalingstochasticinterpolants}
E.~Larsson, R.~Fuentes-Franco, M.~Ivanov, and F.~Lindsten.
\newblock Climate downscaling with stochastic interpolants (cdsi).
\newblock \emph{arXiv preprint arXiv:2603.03838}, 2026.
\newblock URL \url{https://arxiv.org/abs/2603.03838}.

\bibitem[Loshchilov and Hutter(2019)]{adamw}
I.~Loshchilov and F.~Hutter.
\newblock Decoupled weight decay regularization.
\newblock In \emph{International Conference on Learning Representations}, 2019.
\newblock URL \url{https://openreview.net/forum?id=Bkg6RiCqY7}.

\bibitem[Mardani et~al.(2025)Mardani, Brenowitz, Cohen, Pathak, Chen, Liu, Vahdat, Nabian, Ge, Subramaniam, Kashinath, Kautz, and Pritchard]{CorrDiff}
M.~Mardani, N.~Brenowitz, Y.~Cohen, J.~Pathak, C.-Y. Chen, C.-C. Liu, A.~Vahdat, M.~A. Nabian, T.~Ge, A.~Subramaniam, K.~Kashinath, J.~Kautz, and M.~Pritchard.
\newblock Residual corrective diffusion modeling for km-scale atmospheric downscaling.
\newblock \emph{Commun Earth Environ}, 6:\penalty0 124, 2025.
\newblock \doi{10.1038/s43247-025-02042-5}.

\bibitem[M\"{u}ller et~al.(2017)M\"{u}ller, Homleid, Ivarsson, K{\o}ltzow, Lindskog, Midtb{\o}, Andrae, Aspelien, Berggren, Bj{\o}rge, Dahlgren, Kristiansen, Randriamampianina, Ridal, and Vignes]{arome_metcoop}
M.~M\"{u}ller, M.~Homleid, K.-I. Ivarsson, M.~A.~{\O}. K{\o}ltzow, M.~Lindskog, K.~H. Midtb{\o}, U.~Andrae, T.~Aspelien, L.~Berggren, D.~Bj{\o}rge, P.~Dahlgren, J.~Kristiansen, R.~Randriamampianina, M.~Ridal, and O.~Vignes.
\newblock {AROME}-{MetCoOp}: A nordic convective-scale operational weather prediction model.
\newblock \emph{Weather and Forecasting}, 2017.

\bibitem[Nipen et~al.(2026)Nipen, Haugen, Ingstad, Nordhagen, Salihi, Tedesco, Seierstad, Kristiansen, Lang, Alexe, et~al.]{stretch_grid_norway}
T.~N. Nipen, H.~H. Haugen, M.~S. Ingstad, E.~M. Nordhagen, A.~F.~S. Salihi, P.~Tedesco, I.~A. Seierstad, J.~Kristiansen, S.~Lang, M.~Alexe, et~al.
\newblock Regional data-driven weather modeling with a global stretched grid.
\newblock \emph{Artificial Intelligence for the Earth Systems}, 5\penalty0 (2):\penalty0 250001, 2026.

\bibitem[Nordhagen et~al.(2025)Nordhagen, Haugen, Salihi, Ingstad, Nipen, Seierstad, Frogner, Clare, Lang, Chantry, Dueben, and Kristiansen]{nordhagen2025highresolutionprobabilisticdatadrivenweather}
E.~M. Nordhagen, H.~H. Haugen, A.~F.~S. Salihi, M.~S. Ingstad, T.~N. Nipen, I.~A. Seierstad, I.-L. Frogner, M.~Clare, S.~Lang, M.~Chantry, P.~Dueben, and J.~Kristiansen.
\newblock High-resolution probabilistic data-driven weather modeling with a stretched-grid.
\newblock \emph{arXiv preprint arXiv:2511.23043}, 2025.
\newblock URL \url{https://arxiv.org/abs/2511.23043}.

\bibitem[Oskarsson et~al.(2023)Oskarsson, Landelius, and Lindsten]{oskarsson2023graph-lam}
J.~Oskarsson, T.~Landelius, and F.~Lindsten.
\newblock Graph-based neural weather prediction for limited area modeling.
\newblock In \emph{NeurIPS 2023 Workshop on Tackling Climate Change with Machine Learning}, 2023.

\bibitem[Oskarsson et~al.(2024)Oskarsson, Landelius, Deisenroth, and Lindsten]{oskarsson2024probabilistic}
J.~Oskarsson, T.~Landelius, M.~P. Deisenroth, and F.~Lindsten.
\newblock Probabilistic weather forecasting with hierarchical graph neural networks.
\newblock In \emph{Advances in Neural Information Processing Systems}, volume~37, 2024.

\bibitem[Pacchiardi et~al.(2024)Pacchiardi, Adewoyin, Dueben, and Dutta]{prob_fc_scoring_rules}
L.~Pacchiardi, R.~A. Adewoyin, P.~Dueben, and R.~Dutta.
\newblock Probabilistic forecasting with generative networks via scoring rule minimization.
\newblock \emph{Journal of Machine Learning Research}, 25\penalty0 (45):\penalty0 1--64, 2024.
\newblock ISSN 1533-7928.
\newblock URL \url{http://jmlr.org/papers/v25/23-0038.html}.

\bibitem[Pathak et~al.(2026)Pathak, Cohen, Garg, Harrington, Brenowitz, Durran, Mardani, Vahdat, Xu, Kashinath, and Pritchard]{stormCast}
J.~Pathak, Y.~Cohen, P.~Garg, P.~Harrington, N.~Brenowitz, D.~Durran, M.~Mardani, A.~Vahdat, S.~Xu, K.~Kashinath, and M.~Pritchard.
\newblock Kilometer-scale convection-allowing model emulation using generative diffusion modeling.
\newblock \emph{Science Advances}, 12\penalty0 (5):\penalty0 eadv0423, 2026.
\newblock \doi{10.1126/sciadv.adv0423}.
\newblock URL \url{https://www.science.org/doi/abs/10.1126/sciadv.adv0423}.

\bibitem[Price et~al.(2025)Price, Sanchez-Gonzalez, Alet, Andersson, El-Kadi, Masters, Ewalds, Stott, Mohamed, Battaglia, et~al.]{gencast}
I.~Price, A.~Sanchez-Gonzalez, F.~Alet, T.~R. Andersson, A.~El-Kadi, D.~Masters, T.~Ewalds, J.~Stott, S.~Mohamed, P.~Battaglia, et~al.
\newblock Probabilistic weather forecasting with machine learning.
\newblock \emph{Nature}, 637\penalty0 (8044):\penalty0 84--90, 2025.

\bibitem[Rasp et~al.(2024)Rasp, Hoyer, Merose, Langmore, Battaglia, Russell, Sanchez-Gonzalez, Yang, Carver, Agrawal, Chantry, Ben~Bouallegue, Dueben, Bromberg, Sisk, Barrington, Bell, and Sha]{rasp2023weatherbench}
S.~Rasp, S.~Hoyer, A.~Merose, I.~Langmore, P.~Battaglia, T.~Russell, A.~Sanchez-Gonzalez, V.~Yang, R.~Carver, S.~Agrawal, M.~Chantry, Z.~Ben~Bouallegue, P.~Dueben, C.~Bromberg, J.~Sisk, L.~Barrington, A.~Bell, and F.~Sha.
\newblock Weatherbench 2: A benchmark for the next generation of data-driven global weather models.
\newblock \emph{Journal of Advances in Modeling Earth Systems}, 16\penalty0 (6):\penalty0 e2023MS004019, 2024.
\newblock \doi{https://doi.org/10.1029/2023MS004019}.
\newblock URL \url{https://agupubs.onlinelibrary.wiley.com/doi/abs/10.1029/2023MS004019}.

\bibitem[Ridal et~al.(2024)Ridal, Bazile, Le~Moigne, Randriamampianina, Schimanke, Andrae, Berggren, Brousseau, Dahlgren, Edvinsson, El-Said, Glinton, Hagelin, Hopsch, Isaksson, Medeiros, Olsson, Unden, and Wang]{CERRA}
M.~Ridal, E.~Bazile, P.~Le~Moigne, R.~Randriamampianina, S.~Schimanke, U.~Andrae, L.~Berggren, P.~Brousseau, P.~Dahlgren, L.~Edvinsson, A.~El-Said, M.~Glinton, S.~Hagelin, S.~Hopsch, L.~Isaksson, P.~Medeiros, E.~Olsson, P.~Unden, and Z.~Q. Wang.
\newblock {CERRA}, the copernicus european regional reanalysis system.
\newblock \emph{Quarterly Journal of the Royal Meteorological Society}, 150\penalty0 (763):\penalty0 3385--3411, 2024.
\newblock \doi{https://doi.org/10.1002/qj.4764}.
\newblock URL \url{https://rmets.onlinelibrary.wiley.com/doi/abs/10.1002/qj.4764}.

\bibitem[Shi et~al.(2024)Shi, Jin, Han, Gopalakrishnan, and Narasimhan]{shi2024codicastconditionaldiffusionmodel}
J.~Shi, B.~Jin, J.~Han, S.~Gopalakrishnan, and G.~Narasimhan.
\newblock {CoDiCast}: Conditional diffusion model for global weather prediction with uncertainty quantification.
\newblock \emph{arXiv preprint arXiv:2409.05975}, 2024.

\bibitem[Siddiqui et~al.(2024)Siddiqui, Kossaifi, Bonev, Choy, Kautz, Krueger, and Azizzadenesheli]{siddiqui2024exploringdesignspacedeeplearningbased}
S.~A. Siddiqui, J.~Kossaifi, B.~Bonev, C.~Choy, J.~Kautz, D.~Krueger, and K.~Azizzadenesheli.
\newblock Exploring the design space of deep-learning-based weather forecasting systems.
\newblock \emph{arXiv preprint arXiv:2410.07472}, 2024.
\newblock URL \url{https://arxiv.org/abs/2410.07472}.

\bibitem[Song et~al.(2021)Song, Sohl-Dickstein, Kingma, Kumar, Ermon, and Poole]{song2020score}
Y.~Song, J.~Sohl-Dickstein, D.~P. Kingma, A.~Kumar, S.~Ermon, and B.~Poole.
\newblock Score-based generative modeling through stochastic differential equations.
\newblock In \emph{International Conference on Learning Representations}, 2021.

\bibitem[Srivastava et~al.(2024)Srivastava, Yang, Kerrigan, Dresdner, McGibbon, Bretherton, and Mandt]{srivastava2024precipitation}
P.~Srivastava, R.~Yang, G.~Kerrigan, G.~Dresdner, J.~McGibbon, C.~Bretherton, and S.~Mandt.
\newblock Precipitation downscaling with spatiotemporal video diffusion.
\newblock \emph{Advances in Neural Information Processing Systems}, 37:\penalty0 56374--56400, 2024.

\bibitem[Stock et~al.(2025)Stock, Arcomano, and Kotamarthi]{stock2025swiftautoregressiveconsistencymodel}
J.~Stock, T.~Arcomano, and R.~Kotamarthi.
\newblock Swift: An autoregressive consistency model for efficient weather forecasting.
\newblock \emph{arXiv preprint arXiv:2509.25631}, 2025.
\newblock URL \url{https://arxiv.org/abs/2509.25631}.

\bibitem[Wijnands et~al.(2025)Wijnands, Van~Ginderachter, Fran{\c{c}}ois, Buurman, Termonia, and Bleeken]{wijnands2025comparison}
J.~S. Wijnands, M.~Van~Ginderachter, B.~Fran{\c{c}}ois, S.~Buurman, P.~Termonia, and D.~V.~d. Bleeken.
\newblock A comparison of stretched-grid and limited-area modelling for data-driven regional weather forecasting.
\newblock \emph{arXiv preprint arXiv:2507.18378}, 2025.

\bibitem[Xu et~al.(2025)Xu, Zheng, Gao, Wang, Yin, Zhang, Zhang, Luo, Wang, Zhang, et~al.]{xu2024yinglongskillfulhighresolution}
P.~Xu, X.~Zheng, T.~Gao, Y.~Wang, J.~Yin, J.~Zhang, X.~Zhang, S.~Luo, Z.~Wang, Z.~Zhang, et~al.
\newblock An artificial intelligence-based limited area model for forecasting of surface meteorological variables.
\newblock \emph{Communications Earth \& Environment}, 6\penalty0 (1):\penalty0 372, 2025.

\bibitem[Yang et~al.(2026)Yang, Peralta, Amstrup, Hintz, Thorsen, Denby, Kamuk~Christiansen, Schulz, Pelt, and Schreiner]{DANRA}
X.~Yang, C.~Peralta, B.~Amstrup, K.~S. Hintz, S.~B. Thorsen, L.~Denby, S.~Kamuk~Christiansen, H.~Schulz, S.~Pelt, and M.~Schreiner.
\newblock {DANRA}: the kilometer-scale danish regional atmospheric reanalysis.
\newblock \emph{Earth System Science Data}, 18\penalty0 (3):\penalty0 2251--2264, 2026.
\newblock \doi{10.5194/essd-18-2251-2026}.
\newblock URL \url{https://essd.copernicus.org/articles/18/2251/2026/}.

\bibitem[Zamo and Naveau(2018)]{zamo2018estimation}
M.~Zamo and P.~Naveau.
\newblock Estimation of the continuous ranked probability score with limited information and applications to ensemble weather forecasts.
\newblock \emph{Mathematical Geosciences}, 50\penalty0 (2):\penalty0 209--234, 2018.

\end{thebibliography}


\appendix
\crefalias{section}{appendix}
\crefalias{subsection}{appendix}
\newpage
\section[CRPS as a training objective]{\gls{crps} as a training objective}\label{apx:crps_intro}
Scoring rules provide a principled framework for evaluating probabilistic forecasts. Given a predictive distribution $p$ and an observation $x \sim q$, a scoring rule $s(p, x)$ measures the agreement between $p$ and the observation \citep{gneiting2007strictly}. Defining $s(p,q) = \mathbb{E}_{q}[s(p,x)]$, a scoring rule is \emph{proper} if $s(q, q) \leq s(p, q)$ for all $p$ and $q$, and \emph{strictly proper} if equality holds only when $p = q$. Thus, strictly proper scoring rules are minimized uniquely by the true data-generating distribution, making them well-suited as training objectives for probabilistic models.

A widely used strictly proper scoring rule is the \acrlong{crps} \citep{gneiting2007strictly}, which measures the discrepancy between the predictive \gls{cdf} $F$ and the empirical CDF of the observation. For a univariate distribution, the \gls{crps} is defined as
\begin{equation}
    \text{CRPS}(F, x) = \int_{\mathbb{R}} \big(F(\hat{x}) - \mathds{1}(\hat{x} \geq x)\big)^2 \, \mathrm{d}\hat{x}
    =
    \mathbb{E}_p\big[|\hat{x} - x|\big] - \frac{1}{2}\mathbb{E}_{p,p}\big[|\hat{x} - \hat{x}'|\big],
\end{equation}
where $\hat{x}, \hat{x}' \sim p$ are independent samples. The second expression is particularly useful for empirical estimation.

Given an ensemble forecast $\{\hat{x}_n\}_{n=1}^N \sim p$, the \emph{fair} \gls{crps} estimator
\begin{equation}
    \text{CRPS}_{\text{fair}} =
    \frac{1}{N} \sum_{n=1}^{N} \lvert \hat{x}_{n} - x \rvert 
    - \frac{1}{2N(N-1)} 
    \sum_{n=1}^{N} \sum_{n^*=1}^{N} 
    \lvert \hat{x}_{n} - \hat{x}_{n^*} \rvert
\end{equation}
provides an unbiased estimate with respect to the ensemble size \citep{fairCRPS}. However, this estimator can lead to unstable training dynamics \citep{lang2024aifscrpsensembleforecastingusing,fourcastnet3}. To mitigate this, we adopt the almost-fair variant \citep{lang2024aifscrpsensembleforecastingusing},
\begin{equation}
    \text{CRPS}_{\alpha\text{-fair}} = \alpha \text{CRPS}_{\text{fair}} + (1-\alpha)\text{CRPS},
\end{equation}
which interpolates between the unbiased and standard estimators.

\section{Implementation details}\label{apx:implementation_details}
Both Graph-EFM and CRPS-LAM require training with multiple ensemble members for the \gls{crps} loss. This could be done by distributing each ensemble member on a different A100 GPU but we choose to use H200 GPUs instead for simplicity where we could fit two ensemble members on the same GPU. The code and implementation details will be made publicly available upon acceptance of this paper.

\subsection{Models}
For all machine learning models we use the same training budget as in \citep{adamov2025buildingmachinelearninglimited} with 80 epochs of pre-training for single step prediction, followed by 3 epochs of training with four autoregressive steps. Then we finetuned the models on the IFS boundary data for 7 epochs with four autoregressive steps followed by 3 epochs with 12 autoregressive steps. We use the AdamW optimizer \citep{adamw} with $\beta_1=0.9, \beta_2=0.95$ and weight decay $0.01$. All models where required to be possible to train with A100 GPUs, thus not exceeding \SI{80}{\giga\byte} of memory.

\paragraph{Numerical Model.}
To compare our models with an operational numerical baseline, we use forecasts produced by the HARMONIE-AROME NWP forecasting system \citep{bengtsson2017}, run by the Danish Meteorological Institute and initialized from operational analysis. Unfortunately, we only have access to a very limited set of variables and relatively short lead times of up to \SI{18}{\hour}.

\paragraph{Graph-FM.}
We use the exact model setup for the best performing model in \citep{adamov2025buildingmachinelearninglimited} with a four-level hierarchical graph and overlapping boundary. The hidden dimension encoder/decoder was set to 150, and the hidden dimension for the processor was set to 300. This model uses a modified graph structure for the boundary, applying a \SI{400}{\kilo\meter} buffer around the domain rather than extending \SI{400}{\kilo\meter} in each direction to construct a regular grid.

\paragraph{Graph-EFM.}
For the Graph-EFM model, we use a three-level hierarchical graph, after initial experimentation with four hierarchical levels showed that it did not perform well.
The encoder/decoder hidden dimension was set to 150, and the processor's hidden dimension was set to 300. We follow a training curriculum similar to \citep{oskarsson2024probabilistic} shown in \cref{tab:training_curr_graph_efm}.

\begin{table}[h]
    \centering
    \begin{tabular}{cccccc}
        Epochs & KL $\beta$ & CRPS Weight & Learning Rate & AR Steps & Boundary\\
        \hline
         35 & 0 & 0 & 0.001 & 1 & ERA5 \\
         45 & 1 & 0 & 0.001 & 1 & ERA5 \\
         3 & 1 & 0.00001 & 0.0005 & 4 & ERA5\\
         7 & 1 & 0.00001 & 0.0005 & 4 & IFS \\
         3 & 1 & 0.00001 & 0.0005 & 12 & IFS \\
    \end{tabular}
    \caption{Training Curriculum for Graph-EFM}
    \label{tab:training_curr_graph_efm}
\end{table}

\paragraph{DET-LAM (ours).}
We set the hidden dimension of the encoder, processor, and decoder to 300. We train the model with a \gls{mse} objective
\begin{equation}
    \mathcal{L} =\frac{1}{T|G|} \sum_{t=0}^{T-1} \sum_{g \in G} \sum_{d \in D} \omega_d \sigma_d (\hat{X}_{t+1, g, d} - X_{t+1, g, d})^2,
\end{equation}
where
\begin{itemize}
    \item $T$ is the number of autoregressive forecast steps during training.
    \item $G \subset \Omega^I$ is the set of interior grid points.
    \item $D$ is the set of variables that we forecast.
    \item $\sigma_d$ is the inverse variance of the time differences for variable $d$, following \citep{graphcast}. This weighting accounts for variables varying at different temporal scales.
    \item $\omega_d$ is a manually choosen variable-specific weighting. We follow \citep{adamov2025buildingmachinelearninglimited} and use the same weighting scheme.
    \item $\hat{X}_{t+1, g, d}$ is the prediction at gridpoint $g$, for variable $d$ at time step $t$ with the corresponding ground truth $X_{t+1, g, d}$.
\end{itemize}

For the first 80 epochs we use a cosine learning rate scheduler with initial learning rate $10^{-4}$ and end learning rate $10^{-5}$. For the autoregressive training we use a learning rate of $10^{-5}$. 

\paragraph{CRPS-LAM (ours).}
Training with two ensemble members introduces additional memory overhead. To accommodate this within a single H200 GPU, we reduce the hidden dimension of the encoder and decoder to 270, while increasing the processor hidden dimension to 330. This adjustment allows us to fit the full model in memory while simplifying the training setup.

For the first 80 epochs we use a cosine learning rate scheduler with initial learning rate $10^{-4}$ and end learning rate $10^{-5}$. For the autoregressive training we use a learning rate of $10^{-5}$. We use $N=2$ ensemble members, and $\alpha=0.95$ when training CRPS-LAM.

\section{Metrics}\label{apx:metrics}
\label{apx:metrics}
Given an ensemble forecast $\{\hat{X}_{n}\}_{n=1}^N$ with $N$ ensemble members we define the \gls{rmse} of variable $d \in D$ at step $t \in T$ for the ensemble mean $\bar{\hat{X}}_{t, g, d}$ at the spatial position $g \in G \subset \Omega^{I}$ as
$$
    \text{RMSE}_{t, d} =
    \sqrt{
    \frac{1}{|G|N}
    \sum^N_{n=1}
    \sum_{g \in G}
    (\bar{\hat{X}}_{t, g, d,, n} - X_{t, g, d, n})^2
    },
$$
where
$$
\bar{\hat{X}}_{t, g, d} = \frac{1}{N} \sum^N_{n=1}  \hat{X}_{t, g, d, n},
$$
where $\hat{X}_{t, g, d, n}$ is the prediction of ensemble member $n$ with a total number of $N$ ensemble members. Note, we follow the standard convention and the WeatherBench 2 benchmark \citet{rasp2023weatherbench} and apply the square root after sample averaging.

To measure the calibration of the uncertainty in the forecasts we use the bias corrected \gls{ssr} for variable $d$ at step $t$ as
$$
\text{SSR}_{t, d} = \sqrt{\frac{N + 1}{N} \frac{\text{Spread}_{t, d}}{\text{RMSE}_{t, d}}}
$$
where
$$
\text{Spread}_{t, d} =
\sqrt{\frac{1}{|G|N} \sum^{N}_{n=1} \sum_{g \in G} (\bar{\hat{X}}_{t, g, d} - \hat{X}_{t, g, d, n})^2}.
$$
If the uncertainty in the forecasts is well calibrated $\text{SSR}^t_d \approx 1$ \citep{fortin2014should}.

We also compute\gls{crps} \citep{gneiting2007strictly} for variable $d$ at step $t$ 
$$
\text{CRPS}_{t, d} = 
\frac{1}{|G|N} \sum_{g \in G} (\sum^{N}_{n=1} \lvert \hat{X}_{t,d,g,n} - X^{s, t}_{t,g,d} \rvert 
- \frac{1}{2(N-1)}
\sum^{N}_{n=1}\sum^{N}_{n^*=1}\lvert \hat{X}_{t,d,g,n} - \hat{X}_{t,d,g,n^*} \rvert).
$$
Note, we follow the convention of \citet{oskarsson2024probabilistic} and compute the \gls{crps} as a finite sample estimate \citep{zamo2018estimation} over all ensemble members without accounting for any covariance structure.

When calculating metrics for each individual variable separately, we first unnormalize the predictions before comparing them to the ground truth. However, when evaluating the mean performance across all variables, we compute the metrics using normalized data and forecasts. In this case, we normalize the ground truth and compare it to the normalized predictions. The mean normalized score is then obtained by averaging the metric values (\gls{rmse}, \gls{crps}, \gls{ssr}) across all variables.

To evaluate how well the predicted fields preserve the spatial spectral characteristics of the target over different lead times, we compute the \gls{lsd}, defined as
\begin{equation}
\text{LSD}_{t,d}
=
\sqrt{
\frac{1}{\Lambda}
\sum_{\lambda=1}^{\Lambda}
\left(
\log_{10} S(\hat{X}_{t,d})_{\lambda}
-
\log_{10} S(X_{t,d})_{\lambda}
\right)^2
}.
\end{equation}
Here, $S(X)_{\lambda}$ denotes the power spectrum of field $X$ at wavenumber $\lambda$, where $\lambda = 1, \dots, \Lambda$, and $\Lambda$ is the total number of spectral components considered. Lower LSD values indicate better agreement between the predicted and target spectra, and therefore better preservation of spatial structure across scales.

\section{DANRA Dataset details}\label{apx:danra_details}
\begin{figure}[p]
    \centering
    \includegraphics[width=0.9\linewidth]{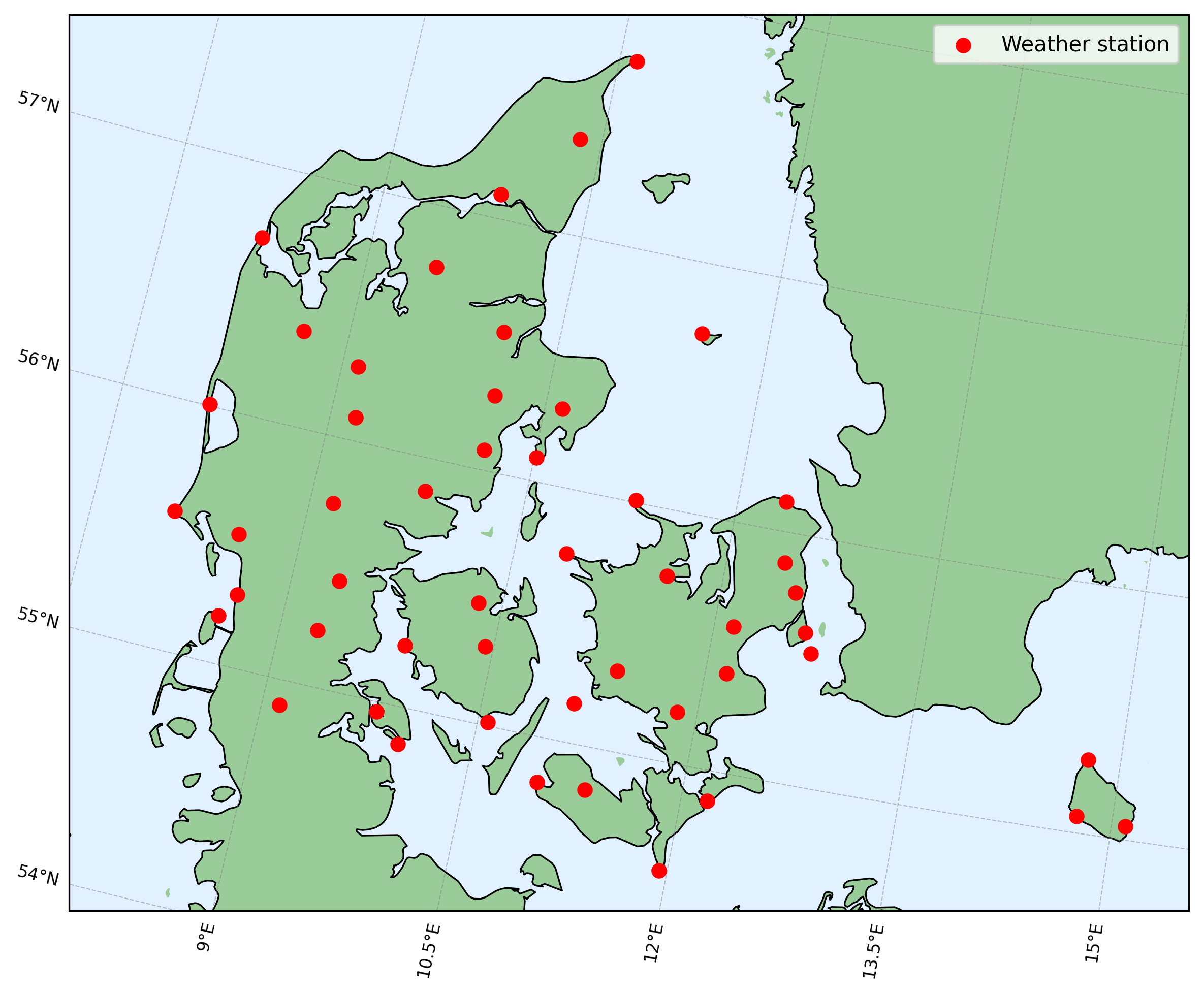}
    \caption{The locations of the in-situ observations in Denmark.}
    \label{fig:station_locations}
\end{figure}

A list of the data used in our experiments over the DANRA\footnote{The DANRA dataset (\url{https://dmidk.github.io/danradocs/}) is openly available under a CC BY 4.0 license.} domain is given in \cref{tab:danra_datasets}
All variables used as model input and forecasted in the DANRA experiments are listed in \cref{tab:danra_variables}. Following \citet{adamov2025buildingmachinelearninglimited}, we split the data into three splits, training (2000-01-01 to 2018-10-29), validation (2018-11-05 to 2019-10-22) and test (2019-10-29 to 2020-10-29). The locations of the in-situ observations in Denmark are shown in \cref{fig:station_locations}.
\newcommand{\wvar}[1]{\texttt{#1}}
\newcommand{\varsepline}{\arrayrulecolor{lightgray}\hdashline[5pt/10pt]}
\setlength\extrarowheight{2pt}

\begin{table}[H]
\centering
\caption{Variables used in the DANRA experimental setup. The ERA5/IFS column describes the boundary forcing. Adapted from \citet{adamov2025buildingmachinelearninglimited}.}
\label{tab:danra_variables}
\begin{tabular}{@{}lclcc@{}}
\toprule
\textbf{Variable} & \textbf{Unit} & \textbf{Abbrev.} & DANRA & ERA5 / IFS \\ \midrule
\textbf{Surface variables} &  &  &  &  \\ \midrule
\SI{10}{\meter} u-component of wind & \si{\metre\per\second} & \wvar{u10m} & \checkmark & \checkmark \\ \varsepline
\SI{10}{\meter} v-component of wind & \si{\metre\per\second} & \wvar{v10m} & \checkmark & \checkmark \\ \varsepline
\SI{2}{\meter} temperature & \si{\kelvin} & \wvar{t2m} & \checkmark & \checkmark \\ \varsepline
Mean sea level pressure & \si{\pascal} & \wvar{msl} & \checkmark & \checkmark \\ \varsepline
Surface pressure & \si{\pascal} & \wvar{sp} & \checkmark & \checkmark \\ \varsepline
Net short wave radiation flux & \si{\watt\per\metre\squared} & \wvar{swavr0m} & \checkmark &  \\ \varsepline
Net long wave radiation flux & \si{\watt\per\metre\squared} & \wvar{lwavr0m} & \checkmark &  \\ \arrayrulecolor{black}\midrule
\textbf{Variables on vertical levels} &  &  &  &  \\ \midrule
Geopotential & \si{\meter\squared\per\second\squared} & \wvar{z} & \checkmark & \checkmark \\ \varsepline
Temperature & \si{\kelvin} & \wvar{t} & \checkmark & \checkmark \\ \varsepline
u-component of wind & \si{\metre\per\second} & \wvar{u} & \checkmark & \checkmark \\ \varsepline
v-component of wind & \si{\metre\per\second} & \wvar{v} & \checkmark & \checkmark \\ \varsepline
Vertical velocity & \si{\pascal\per\second} & \wvar{w} & \checkmark & \checkmark \\ \varsepline
Relative humidity & - & \wvar{r} & \checkmark &  \\ \varsepline
Specific humidity & \si{\kilo\gram\per\kilo\gram} & \wvar{q} &  & \checkmark \\ \arrayrulecolor{black}\midrule
\textbf{Forcing} &  &  &  &  \\ \midrule
Top of atmosphere radiation flux & \si{\watt\per\metre\squared} & \wvar{toarf} & \checkmark & \checkmark \\ \varsepline
$\sin$-encoded time of day & - & \wvar{sin\_d} & \checkmark & \checkmark \\ \varsepline
$\cos$-encoded time of day & - & \wvar{cos\_d} & \checkmark & \checkmark \\ \varsepline
$\sin$-encoded day of year & - & \wvar{sin\_y} & \checkmark & \checkmark \\ \varsepline
$\cos$-encoded day of year & - & \wvar{cos\_y} & \checkmark & \checkmark \\ \arrayrulecolor{black}\midrule
\textbf{Static fields} &  &  &  &  \\ \midrule
Land-sea mask & - & \wvar{lsm} & \checkmark & \checkmark \\ \varsepline
Surface geopotential (orography) & \si{\meter\squared\per\second\squared} & \wvar{oro} & \checkmark & \checkmark \\ \arrayrulecolor{black}\bottomrule
\end{tabular}
\end{table}

\setlength\extrarowheight{0pt}
\begin{table}[H]
\centering
\caption{Overview of data products used for DANRA experiments.}
\label{tab:danra_datasets}
\begin{tabular}{@{}llllll@{}}
\toprule
\textbf{Dataset} & \textbf{Type} & \textbf{Use} & \textbf{Resolution} & \textbf{Time step} & \textbf{Length} \\ \midrule
DANRA & Reanalysis & Training & \SI{2.5}{\kilo\meter} & \SI{3}{\hour} & 21 years \\ \midrule
ERA5 & Reanalysis & \multirow{2}{*}{\centering Boundary} & \SI{0.25}{\degree} & \SI{6}{\hour} & 21 years \\
IFS HRES & Forecasts & & \SI{0.25}{\degree} & \SI{6}{\hour} & 2 years \\ \midrule
DMI SYNOP & Observations & \multirow{2}{*}{\centering Evaluation} & \centering 50 Stations & \centering 1 h & 1 year \\
DANRA Forecasts & Forecast & & \SI{2.5}{\kilo\meter} & \SI{3}{\hour} & 1 year \\ \bottomrule
\end{tabular}
\end{table}

\section{Additional Experiments on the DANRA dataset}\label{apx:danra_exp}
\begin{figure}[H]
    \centering
    \includegraphics[width=\textwidth]{figures/legend_era5_nwp.pdf}
    
    \begin{subfigure}[t]{0.33\textwidth}
        \centering
        \includegraphics[width=\textwidth]{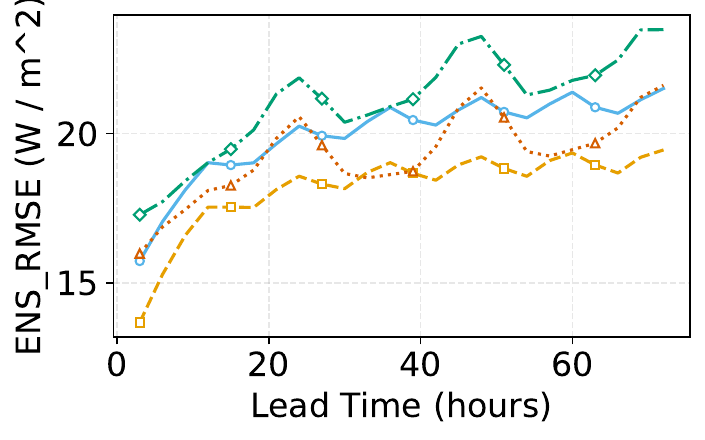}
        \caption{\gls{rmse} for \texttt{lwavr0m}}
    \end{subfigure}%
    \hfill%
    \begin{subfigure}[t]{0.33\textwidth}
        \centering
        \includegraphics[width=\textwidth]{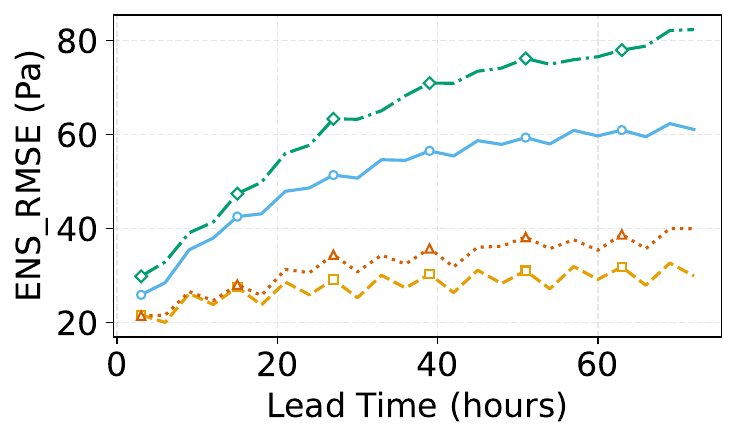}
        \caption{\gls{rmse} for \texttt{msl}}
    \end{subfigure}%
    \hfill%
    \begin{subfigure}[t]{0.33\textwidth}
        \centering
        \includegraphics[width=\textwidth]{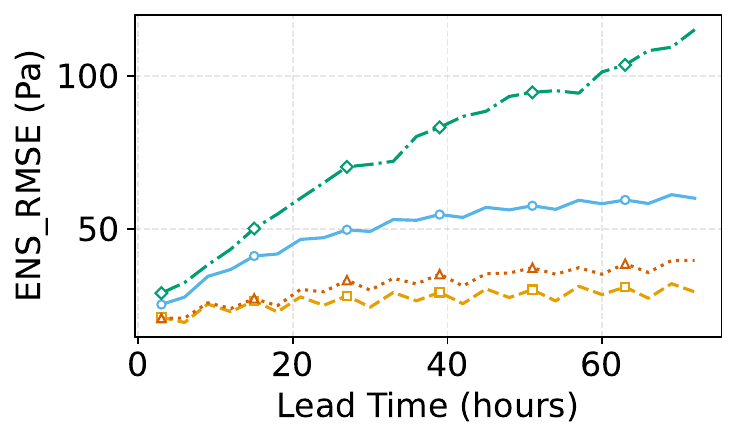}
        \caption{\gls{rmse} for \texttt{sp}}
    \end{subfigure}
    
    \vspace{0.5em}
    
    \begin{subfigure}[t]{0.33\textwidth}
        \centering
        \includegraphics[width=\textwidth]{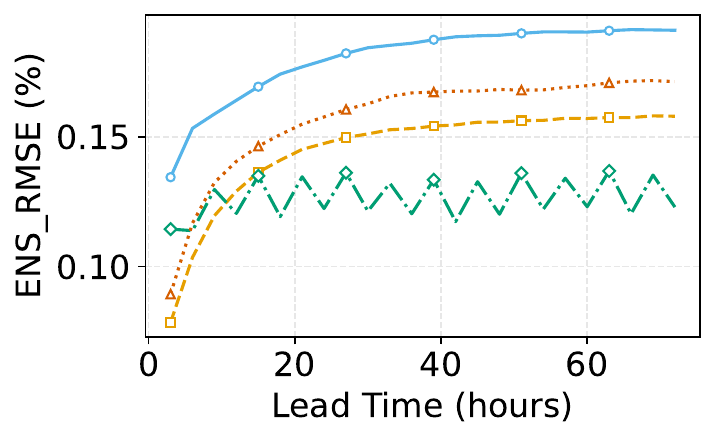}
        \caption{\gls{rmse} for \texttt{r600}}
    \end{subfigure}%
    \hfill%
    \begin{subfigure}[t]{0.33\textwidth}
        \centering
        \includegraphics[width=\textwidth]{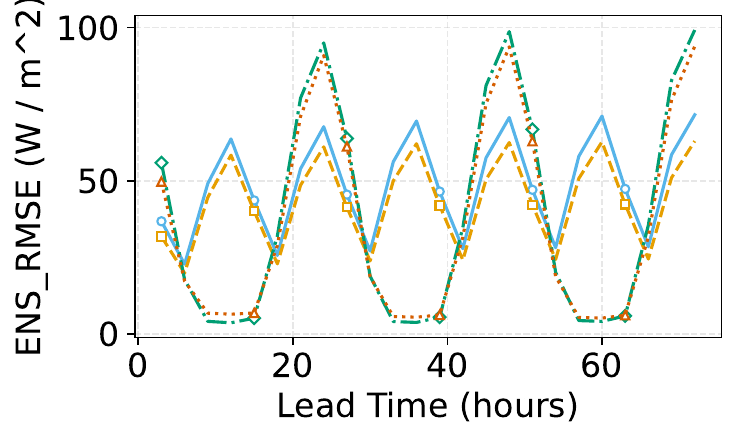}
        \caption{\gls{rmse} for \texttt{swavr0m}}
    \end{subfigure}%
    \hfill%
    \begin{subfigure}[t]{0.33\textwidth}
        \centering
        \includegraphics[width=\textwidth]{danra_plots_era5/t2m_ens_rmse.pdf}
        \caption{\gls{rmse} for \texttt{t2m}}
    \end{subfigure}%

    \vspace{0.5em}
    
    \begin{subfigure}[t]{0.33\textwidth}
        \centering
        \includegraphics[width=\textwidth]{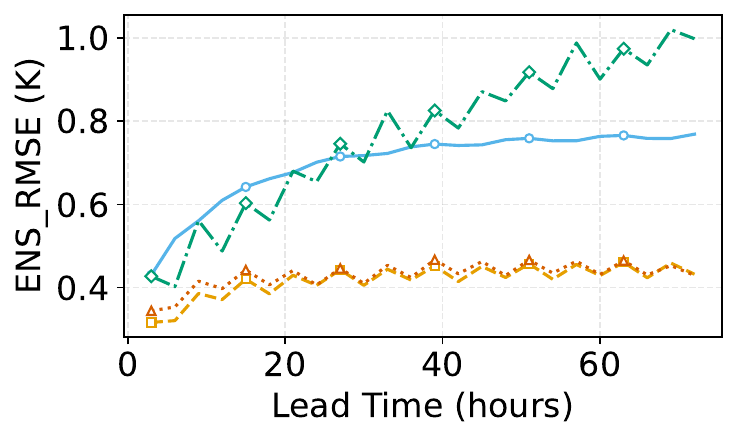}
        \caption{\gls{rmse} for \texttt{t600}}
    \end{subfigure}%
    \hfill%
    \begin{subfigure}[t]{0.33\textwidth}
        \centering
        \includegraphics[width=\textwidth]{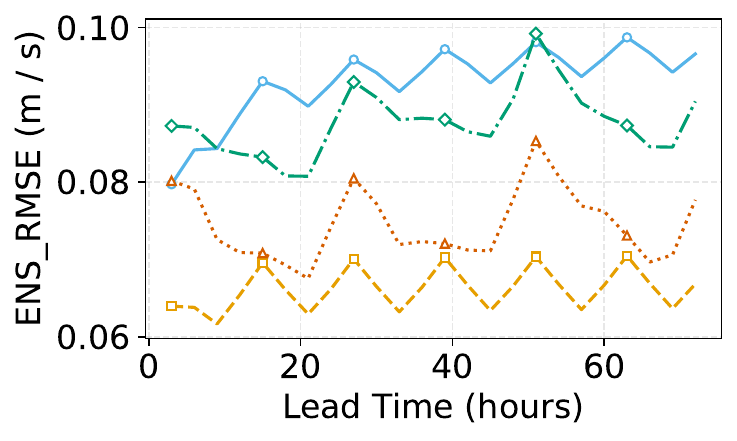}
        \caption{\gls{rmse} for \texttt{tw600}}
    \end{subfigure}
    \hfill%
    \begin{subfigure}[t]{0.33\textwidth}
        \centering
        \includegraphics[width=\textwidth]{danra_plots_era5/u10m_ens_rmse.pdf}
        \caption{\gls{rmse} for \texttt{u10m}}
    \end{subfigure}%

    \vspace{0.5em}
    
    \begin{subfigure}[t]{0.33\textwidth}
        \centering
        \includegraphics[width=\textwidth]{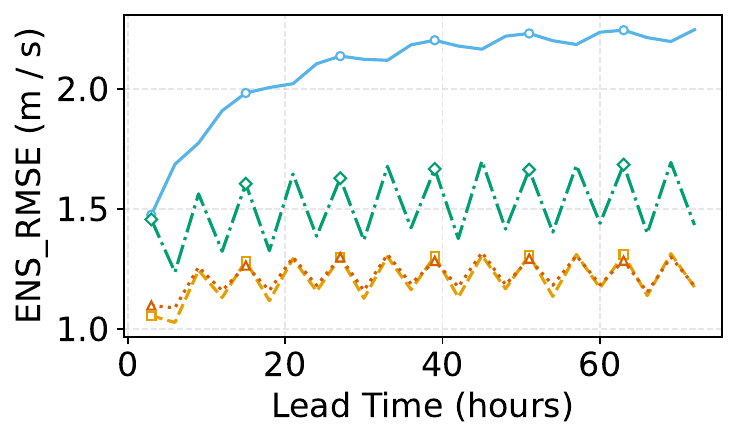}
        \caption{\gls{rmse} for \texttt{u600}}
    \end{subfigure}%
    \hfill%
    \begin{subfigure}[t]{0.33\textwidth}
        \centering
        \includegraphics[width=\textwidth]{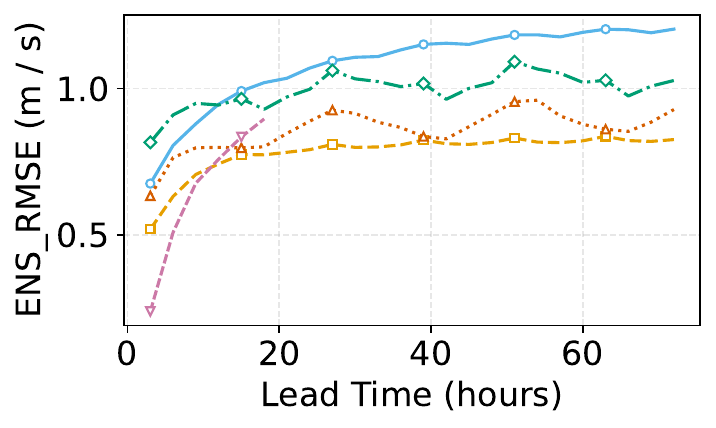}
        \caption{\gls{rmse} for \texttt{v10m}}
    \end{subfigure}%
    \hfill%
    \begin{subfigure}[t]{0.33\textwidth}
        \centering
        \includegraphics[width=\textwidth]{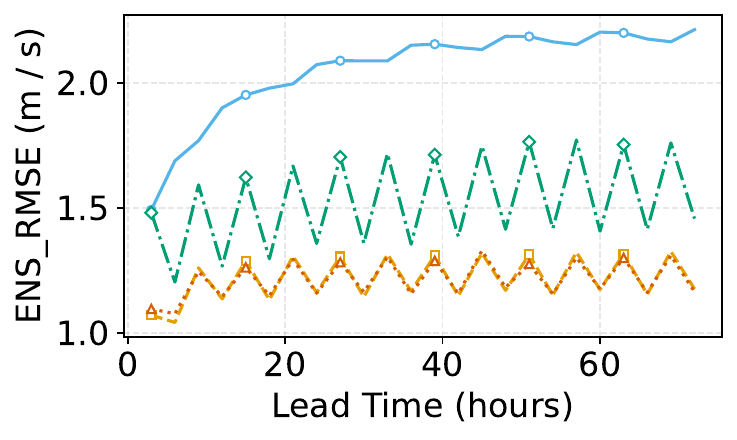}
        \caption{\gls{rmse} for \texttt{v600}}
    \end{subfigure}%

    \vspace{0.5em}
    
    \begin{subfigure}[t]{0.33\textwidth}
        \centering
        \includegraphics[width=\textwidth]{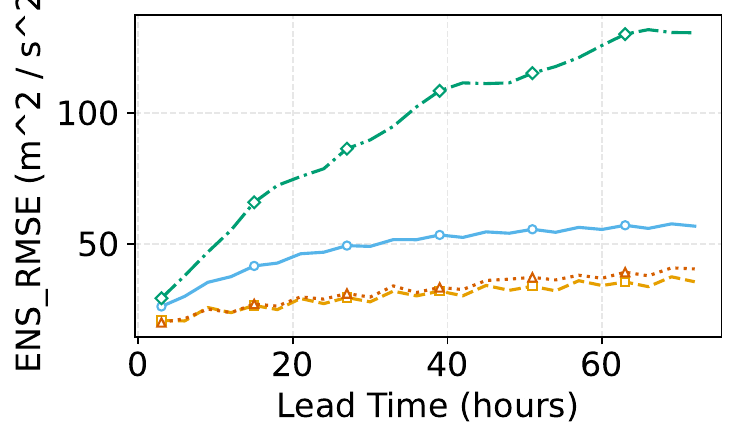}
        \caption{\gls{rmse} for \texttt{z600}}
    \end{subfigure}%
    
    \caption{The \gls{rmse} the DANRA dataset with ERA5 boundary conditions.}
    \label{fig:DANRA_rmse_full}
\end{figure}

\begin{figure}[H]
    \centering
    \includegraphics[width=\textwidth]{figures/legend_era5.pdf}
    
    \begin{subfigure}[t]{0.33\textwidth}
        \centering
        \includegraphics[width=\textwidth]{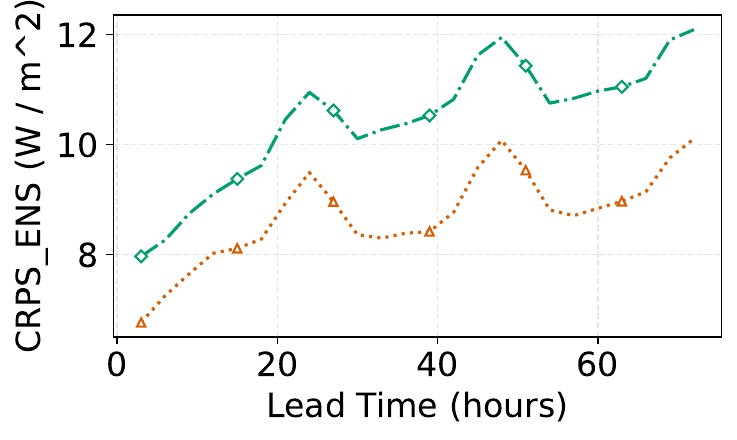}
        \caption{\gls{crps} for \texttt{lwavr0m}}
    \end{subfigure}%
    \hfill%
    \begin{subfigure}[t]{0.33\textwidth}
        \centering
        \includegraphics[width=\textwidth]{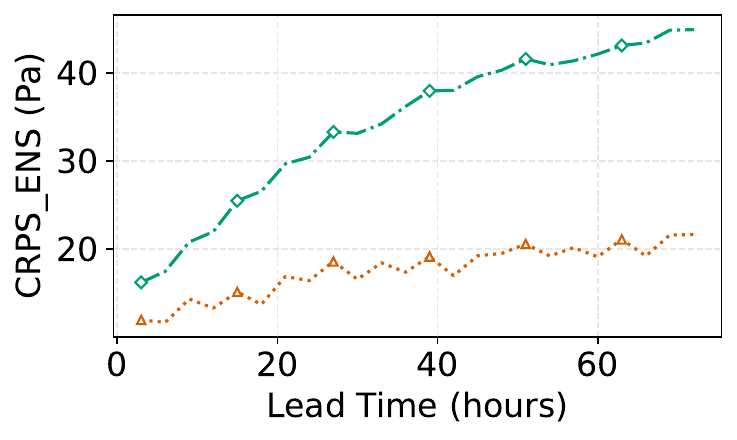}
        \caption{\gls{crps} for \texttt{msl}}
    \end{subfigure}%
    \hfill%
    \begin{subfigure}[t]{0.33\textwidth}
        \centering
        \includegraphics[width=\textwidth]{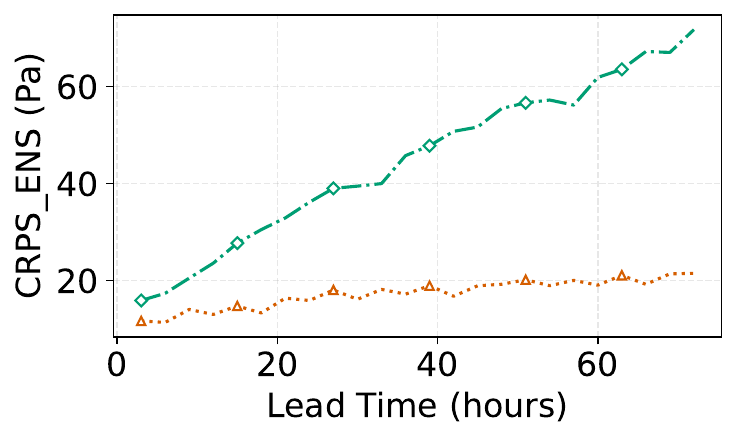}
        \caption{\gls{crps} for \texttt{sp}}
    \end{subfigure}
    
    \vspace{0.5em}
    
    \begin{subfigure}[t]{0.33\textwidth}
        \centering
        \includegraphics[width=\textwidth]{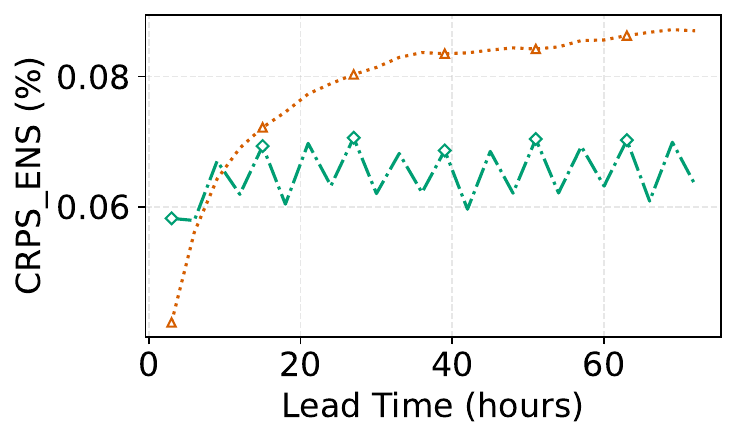}
        \caption{\gls{crps} for \texttt{r600}}
    \end{subfigure}%
    \hfill%
    \begin{subfigure}[t]{0.33\textwidth}
        \centering
        \includegraphics[width=\textwidth]{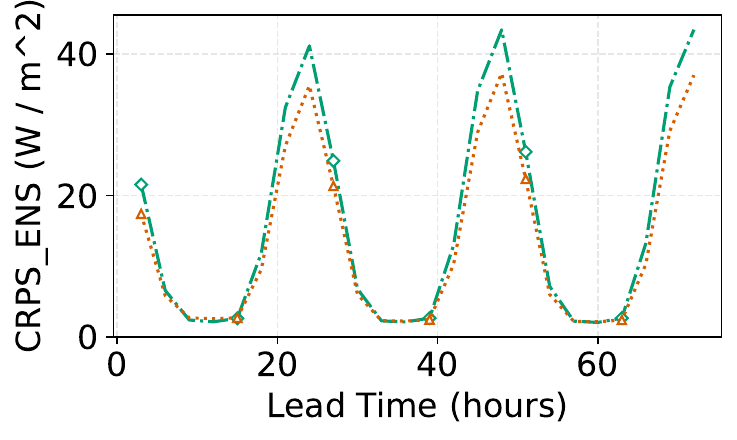}
        \caption{\gls{crps} for \texttt{swavr0m}}
    \end{subfigure}%
    \hfill%
    \begin{subfigure}[t]{0.33\textwidth}
        \centering
        \includegraphics[width=\textwidth]{danra_plots_era5/t2m_crps_ens.pdf}
        \caption{\gls{crps} for \texttt{t2m}}
    \end{subfigure}%

    \vspace{0.5em}
    
    \begin{subfigure}[t]{0.33\textwidth}
        \centering
        \includegraphics[width=\textwidth]{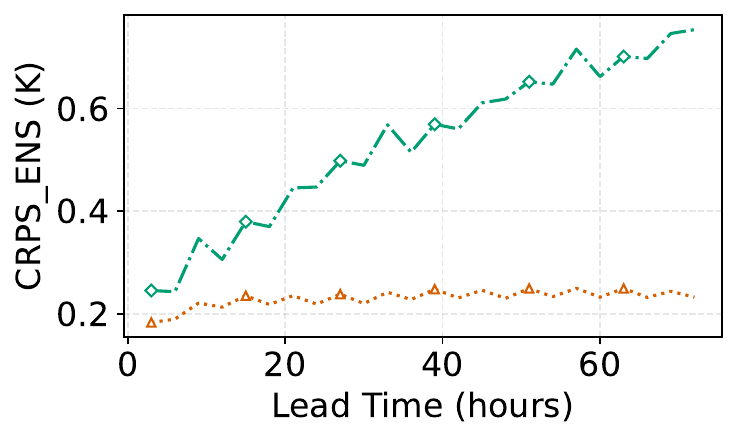}
        \caption{\gls{crps} for \texttt{t600}}
    \end{subfigure}%
    \hfill%
    \begin{subfigure}[t]{0.33\textwidth}
        \centering
        \includegraphics[width=\textwidth]{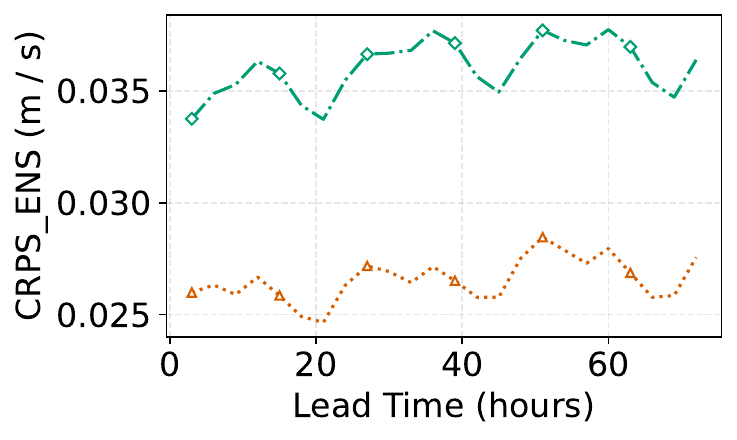}
        \caption{\gls{crps} for \texttt{tw600}}
    \end{subfigure}
    \hfill%
    \begin{subfigure}[t]{0.33\textwidth}
        \centering
        \includegraphics[width=\textwidth]{danra_plots_era5/u10m_crps_ens.pdf}
        \caption{\gls{crps} for \texttt{u10m}}
    \end{subfigure}%

    \vspace{0.5em}
    
    \begin{subfigure}[t]{0.33\textwidth}
        \centering
        \includegraphics[width=\textwidth]{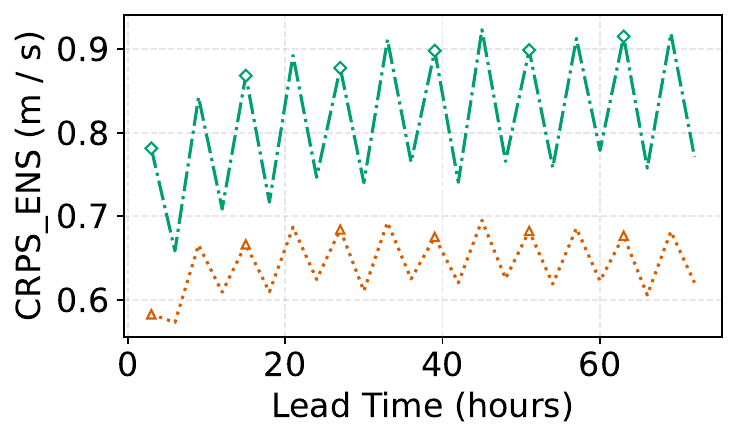}
        \caption{\gls{crps} for \texttt{u600}}
    \end{subfigure}%
    \hfill%
    \begin{subfigure}[t]{0.33\textwidth}
        \centering
        \includegraphics[width=\textwidth]{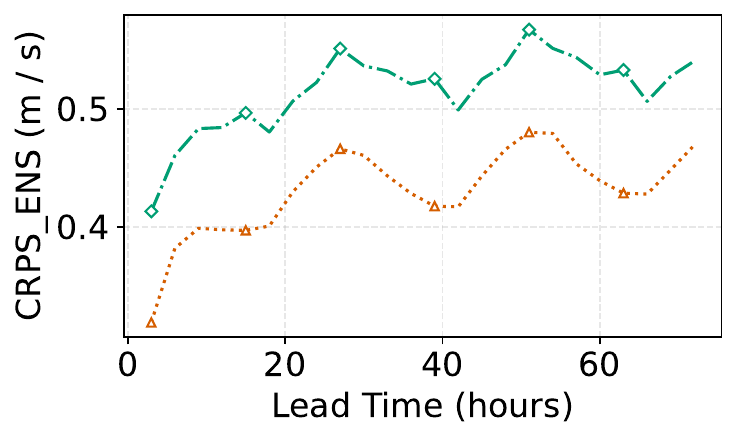}
        \caption{\gls{crps} for \texttt{v10m}}
    \end{subfigure}%
    \hfill%
    \begin{subfigure}[t]{0.33\textwidth}
        \centering
        \includegraphics[width=\textwidth]{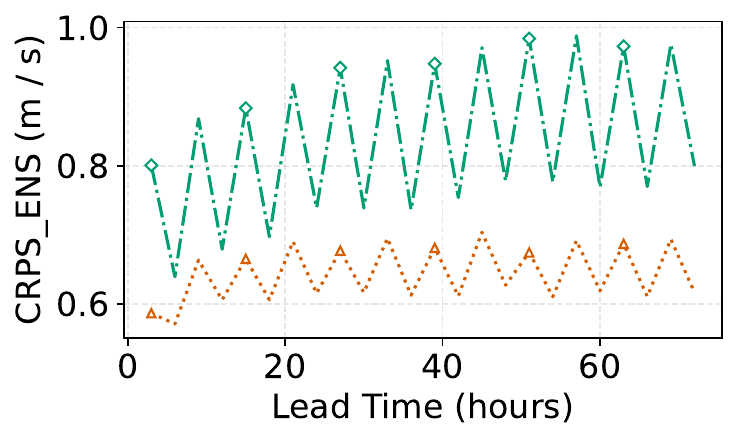}
        \caption{\gls{crps} for \texttt{v600}}
    \end{subfigure}%

    \vspace{0.5em}
    
    \begin{subfigure}[t]{0.33\textwidth}
        \centering
        \includegraphics[width=\textwidth]{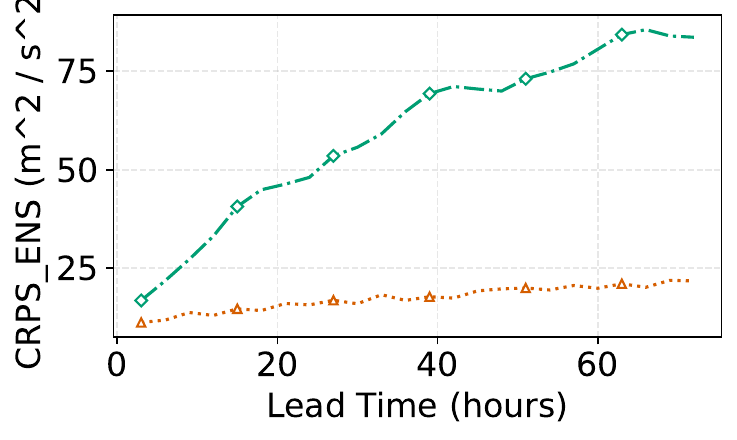}
        \caption{\gls{crps} for \texttt{z600}}
    \end{subfigure}%
    
    \caption{The \gls{crps} the DANRA dataset with ERA5 boundary conditions.}
    \label{fig:DANRA_crps_full}
\end{figure}

\begin{figure}[H]
    \centering
    \includegraphics[width=\textwidth]{figures/legend_era5.pdf}
    
    \begin{subfigure}[t]{0.33\textwidth}
        \centering
        \includegraphics[width=\textwidth]{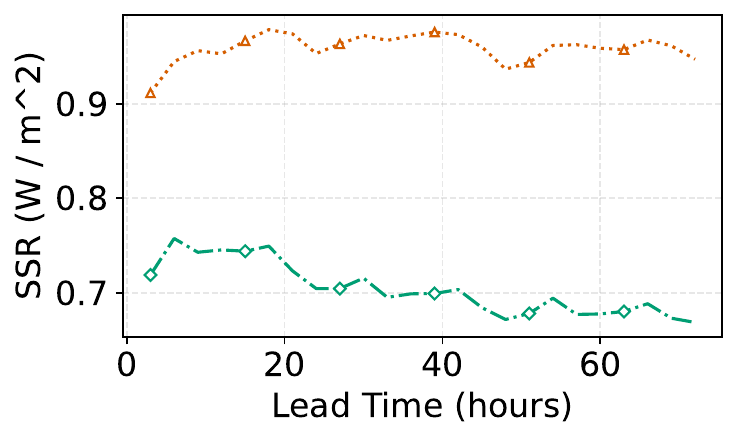}
        \caption{\gls{ssr} for \texttt{lwavr0m}}
    \end{subfigure}%
    \hfill%
    \begin{subfigure}[t]{0.33\textwidth}
        \centering
        \includegraphics[width=\textwidth]{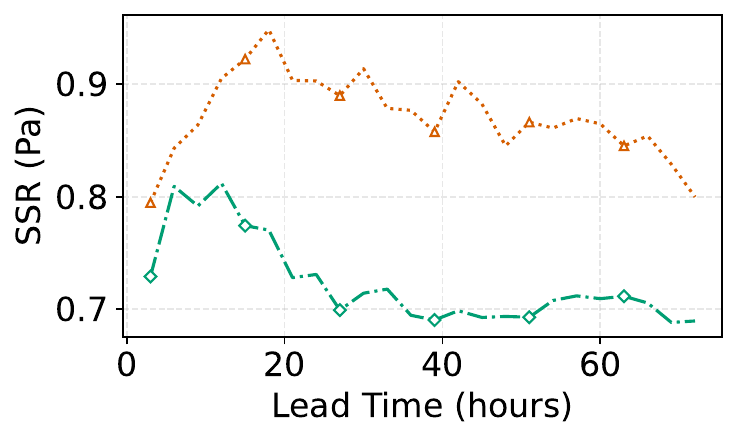}
        \caption{\gls{ssr} for \texttt{msl}}
    \end{subfigure}%
    \hfill%
    \begin{subfigure}[t]{0.33\textwidth}
        \centering
        \includegraphics[width=\textwidth]{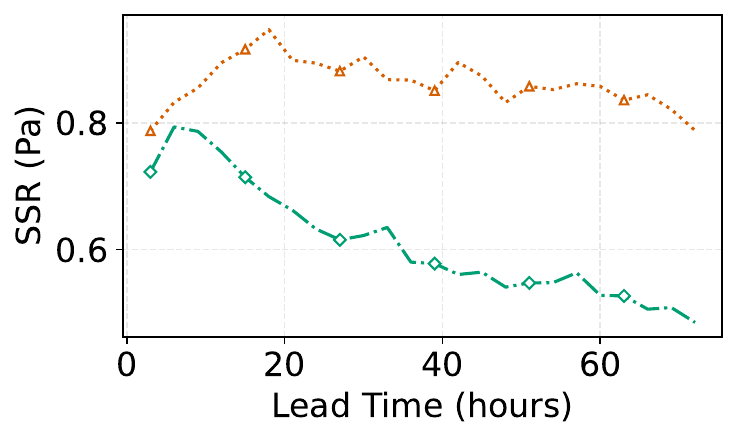}
        \caption{\gls{ssr} for \texttt{sp}}
    \end{subfigure}
    
    \vspace{0.5em}
    
    \begin{subfigure}[t]{0.33\textwidth}
        \centering
        \includegraphics[width=\textwidth]{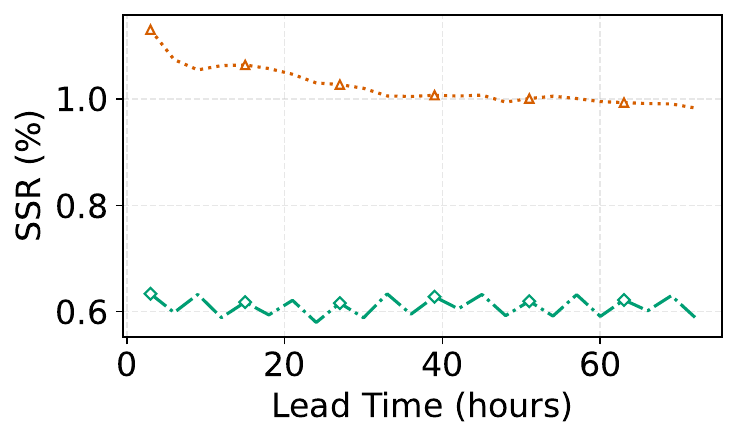}
        \caption{\gls{ssr} for \texttt{r600}}
    \end{subfigure}%
    \hfill%
    \begin{subfigure}[t]{0.33\textwidth}
        \centering
        \includegraphics[width=\textwidth]{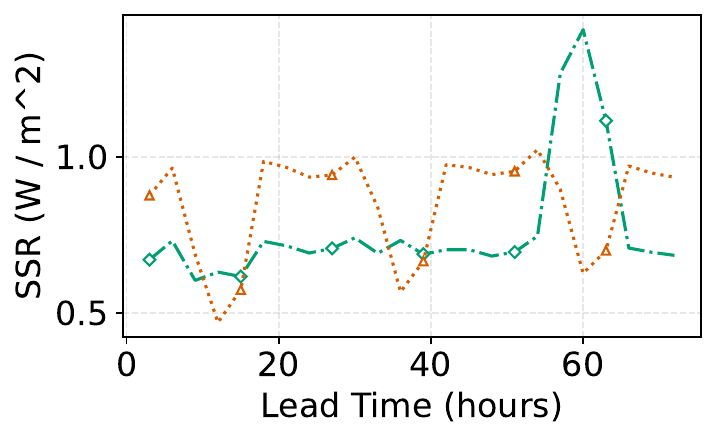}
        \caption{\gls{ssr} for \texttt{swavr0m}}
    \end{subfigure}%
    \hfill%
    \begin{subfigure}[t]{0.33\textwidth}
        \centering
        \includegraphics[width=\textwidth]{danra_plots_era5/t2m_ssr.pdf}
        \caption{\gls{ssr} for \texttt{t2m}}
    \end{subfigure}%

    \vspace{0.5em}
    
    \begin{subfigure}[t]{0.33\textwidth}
        \centering
        \includegraphics[width=\textwidth]{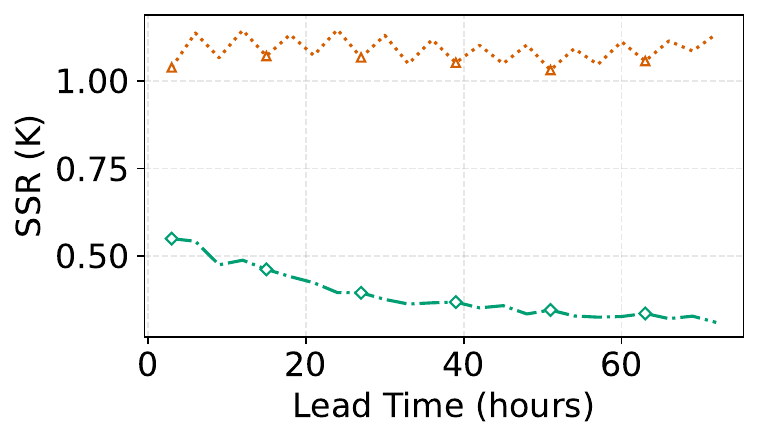}
        \caption{\gls{ssr} for \texttt{t600}}
    \end{subfigure}%
    \hfill%
    \begin{subfigure}[t]{0.33\textwidth}
        \centering
        \includegraphics[width=\textwidth]{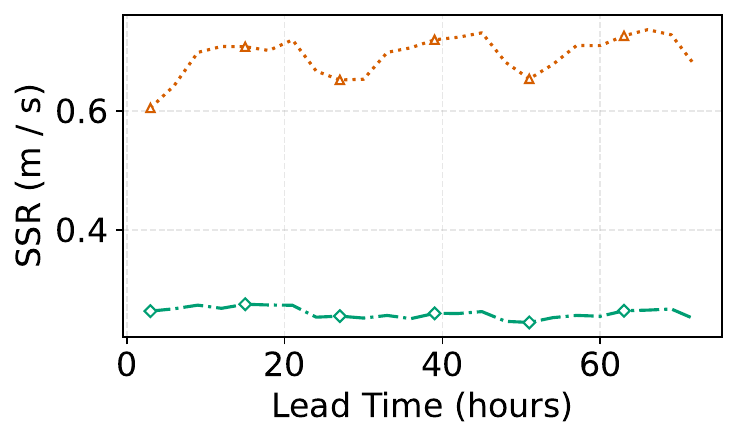}
        \caption{\gls{ssr} for \texttt{tw600}}
    \end{subfigure}
    \hfill%
    \begin{subfigure}[t]{0.33\textwidth}
        \centering
        \includegraphics[width=\textwidth]{danra_plots_era5/u10m_ssr.pdf}
        \caption{\gls{ssr} for \texttt{u10m}}
    \end{subfigure}%

    \vspace{0.5em}
    
    \begin{subfigure}[t]{0.33\textwidth}
        \centering
        \includegraphics[width=\textwidth]{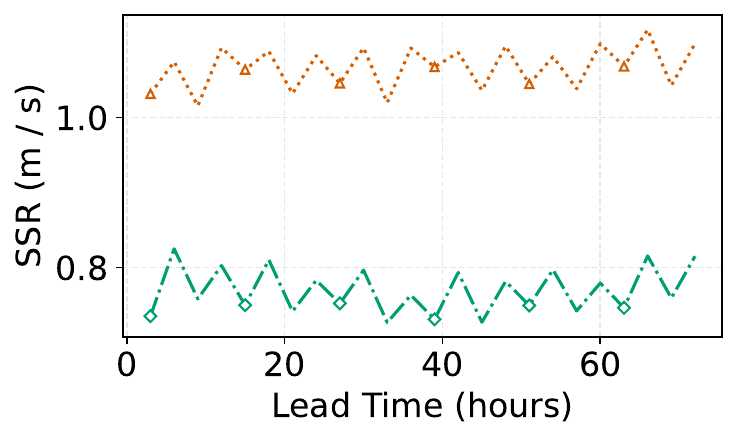}
        \caption{\gls{ssr} for \texttt{u600}}
    \end{subfigure}%
    \hfill%
    \begin{subfigure}[t]{0.33\textwidth}
        \centering
        \includegraphics[width=\textwidth]{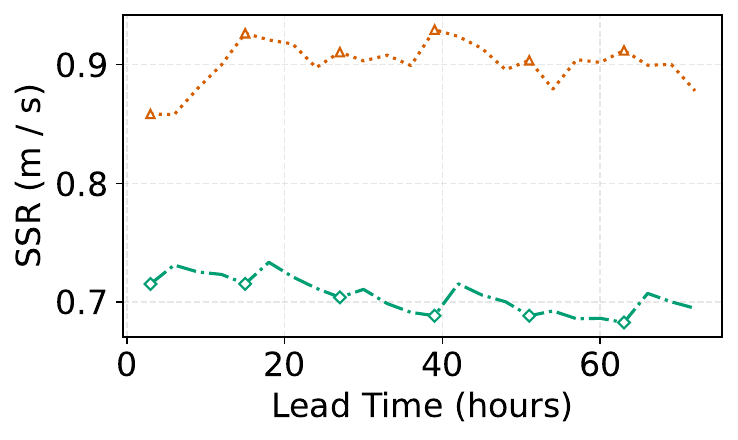}
        \caption{\gls{ssr} for \texttt{v10m}}
    \end{subfigure}%
    \hfill%
    \begin{subfigure}[t]{0.33\textwidth}
        \centering
        \includegraphics[width=\textwidth]{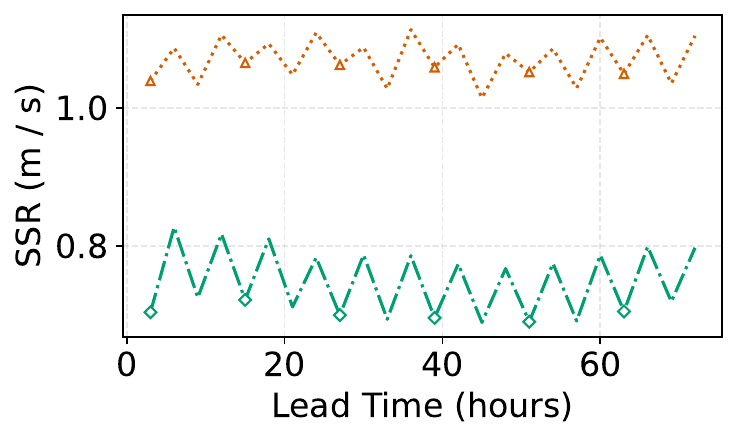}
        \caption{\gls{ssr} for \texttt{v600}}
    \end{subfigure}%

    \vspace{0.5em}
    
    \begin{subfigure}[t]{0.33\textwidth}
        \centering
        \includegraphics[width=\textwidth]{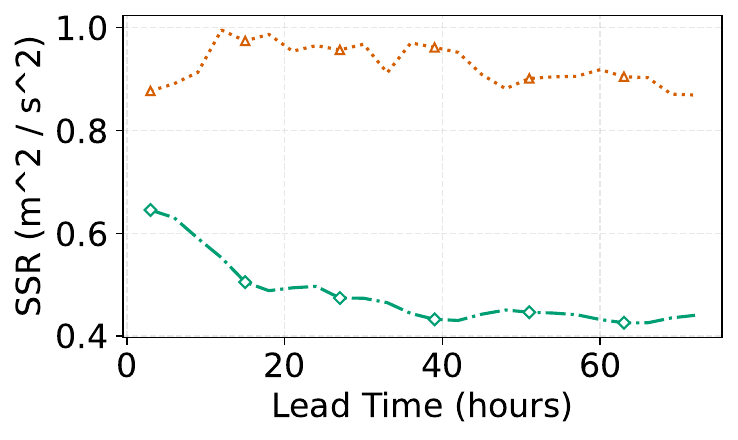}
        \caption{\gls{ssr} for \texttt{z600}}
    \end{subfigure}%
    
    \caption{The \gls{ssr} the DANRA dataset with ERA5 boundary conditions.}
    \label{fig:DANRA_ssr_full}
\end{figure}

\begin{figure}[H]
    \centering
    \includegraphics[width=\textwidth]{figures/legend_era5.pdf}
    
    \begin{subfigure}[t]{0.33\textwidth}
        \centering
        \includegraphics[width=\textwidth]{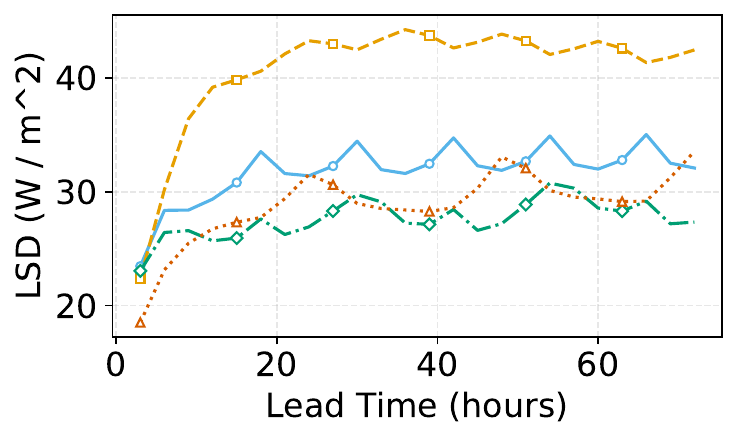}
        \caption{\gls{lsd} for \texttt{lwavr0m}}
    \end{subfigure}%
    \hfill%
    \begin{subfigure}[t]{0.33\textwidth}
        \centering
        \includegraphics[width=\textwidth]{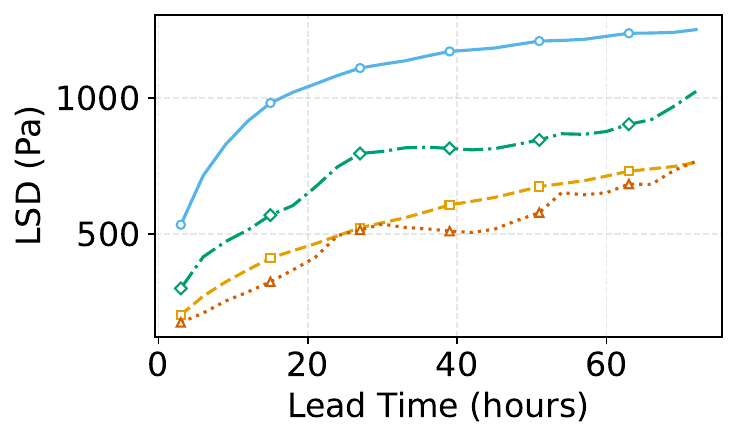}
        \caption{\gls{lsd} for \texttt{msl}}
    \end{subfigure}%
    \hfill%
    \begin{subfigure}[t]{0.33\textwidth}
        \centering
        \includegraphics[width=\textwidth]{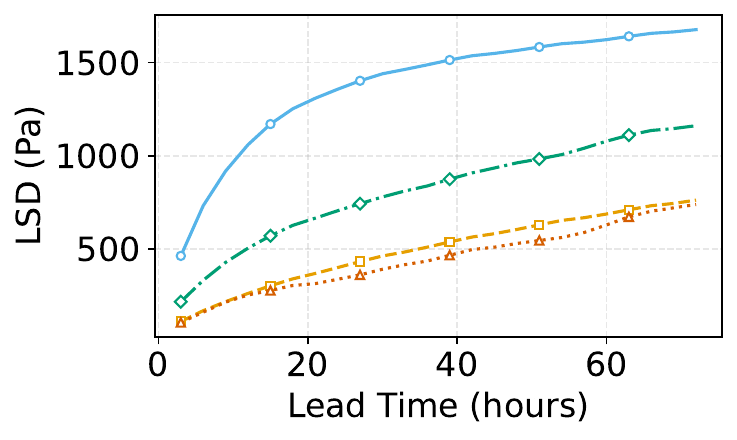}
        \caption{\gls{lsd} for \texttt{sp}}
    \end{subfigure}
    
    \vspace{0.5em}
    
    \begin{subfigure}[t]{0.33\textwidth}
        \centering
        \includegraphics[width=\textwidth]{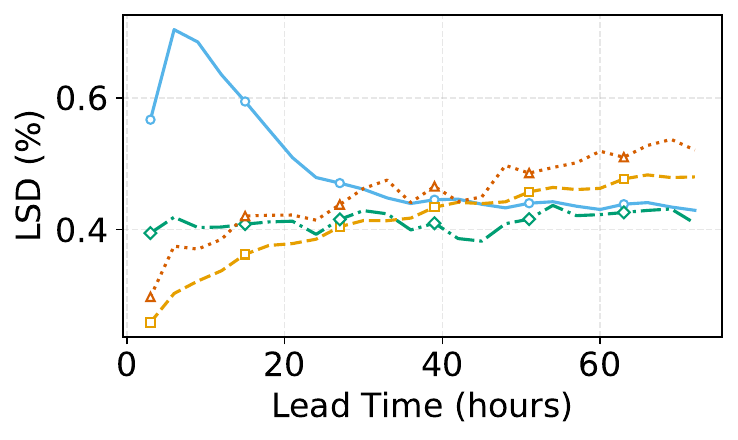}
        \caption{\gls{lsd} for \texttt{r600}}
    \end{subfigure}%
    \hfill%
    \begin{subfigure}[t]{0.33\textwidth}
        \centering
        \includegraphics[width=\textwidth]{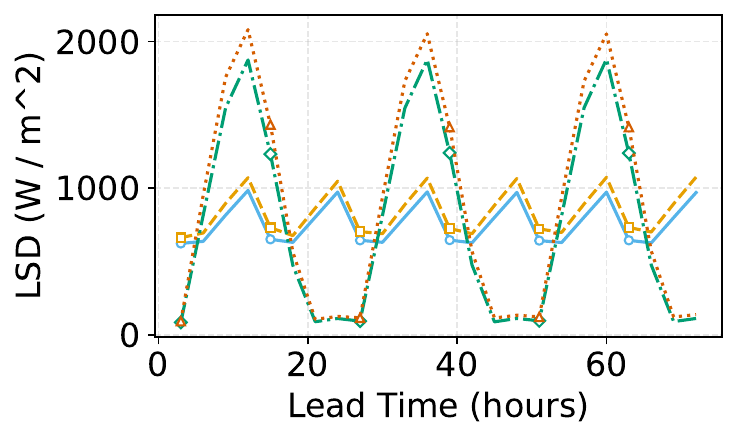}
        \caption{\gls{lsd} for \texttt{swavr0m}}
    \end{subfigure}%
    \hfill%
    \begin{subfigure}[t]{0.33\textwidth}
        \centering
        \includegraphics[width=\textwidth]{danra_plots_era5/t2m_lsd.pdf}
        \caption{\gls{lsd} for \texttt{t2m}}
    \end{subfigure}%

    \vspace{0.5em}
    
    \begin{subfigure}[t]{0.33\textwidth}
        \centering
        \includegraphics[width=\textwidth]{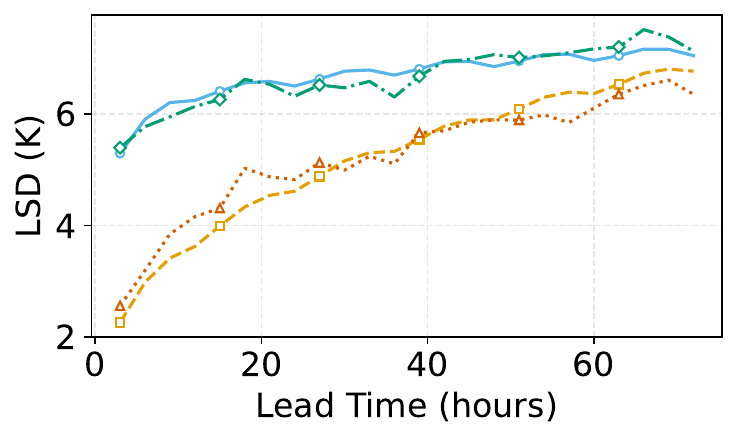}
        \caption{\gls{lsd} for \texttt{t600}}
    \end{subfigure}%
    \hfill%
    \begin{subfigure}[t]{0.33\textwidth}
        \centering
        \includegraphics[width=\textwidth]{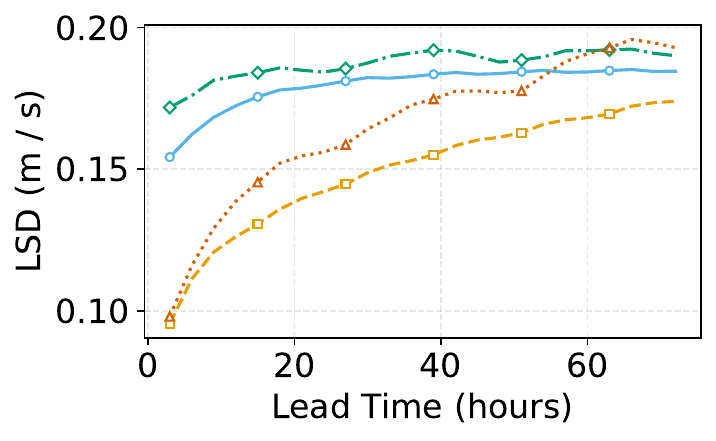}
        \caption{\gls{lsd} for \texttt{tw600}}
    \end{subfigure}
    \hfill%
    \begin{subfigure}[t]{0.33\textwidth}
        \centering
        \includegraphics[width=\textwidth]{danra_plots_era5/u10m_lsd.pdf}
        \caption{\gls{lsd} for \texttt{u10m}}
    \end{subfigure}%

    \vspace{0.5em}
    
    \begin{subfigure}[t]{0.33\textwidth}
        \centering
        \includegraphics[width=\textwidth]{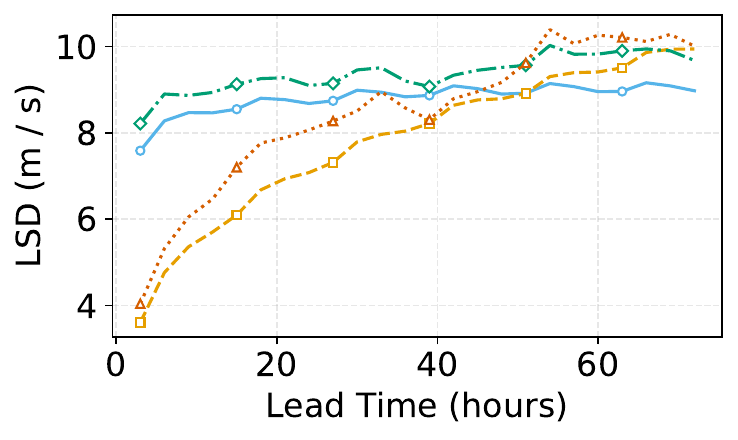}
        \caption{\gls{lsd} for \texttt{u600}}
    \end{subfigure}%
    \hfill%
    \begin{subfigure}[t]{0.33\textwidth}
        \centering
        \includegraphics[width=\textwidth]{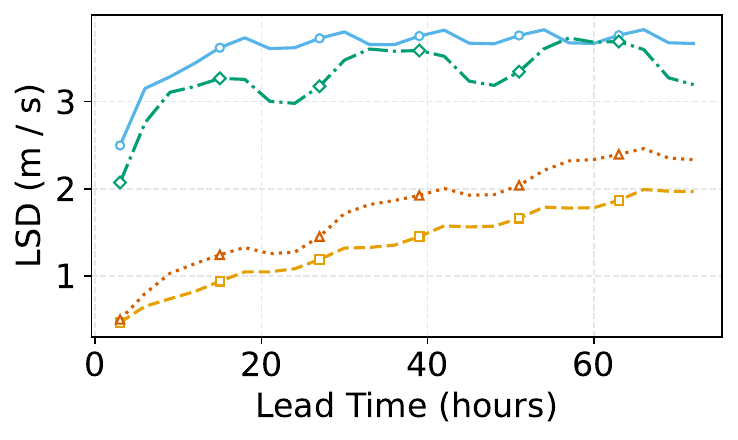}
        \caption{\gls{lsd} for \texttt{v10m}}
    \end{subfigure}%
    \hfill%
    \begin{subfigure}[t]{0.33\textwidth}
        \centering
        \includegraphics[width=\textwidth]{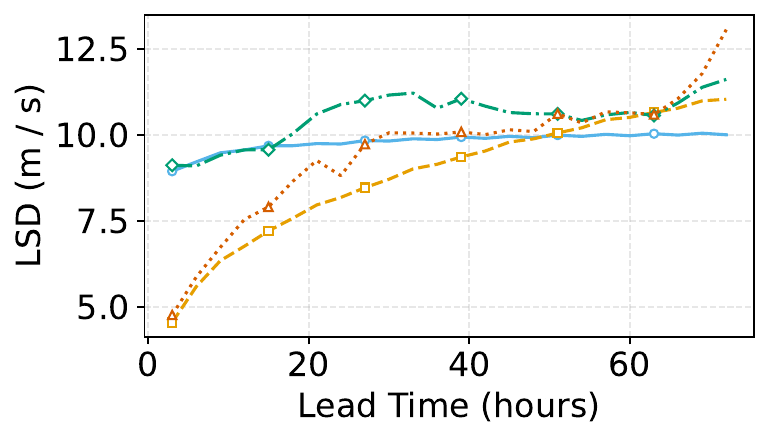}
        \caption{\gls{lsd} for \texttt{v600}}
    \end{subfigure}%

    \vspace{0.5em}
    
    \begin{subfigure}[t]{0.33\textwidth}
        \centering
        \includegraphics[width=\textwidth]{danra_plots_era5/z600_lsd.pdf}
        \caption{\gls{lsd} for \texttt{z600}}
    \end{subfigure}%
    
    \caption{The \gls{lsd} the DANRA dataset with ERA5 boundary conditions.}
    \label{fig:DANRA_lsd_full}
\end{figure}

\begin{figure}[tbph]
    \centering%
    \includegraphics[width=\textwidth]{figures/legend_ifs_nwp.pdf}
    \begin{subfigure}[b]{0.33\textwidth}
        \centering
        \includegraphics[width=\textwidth]{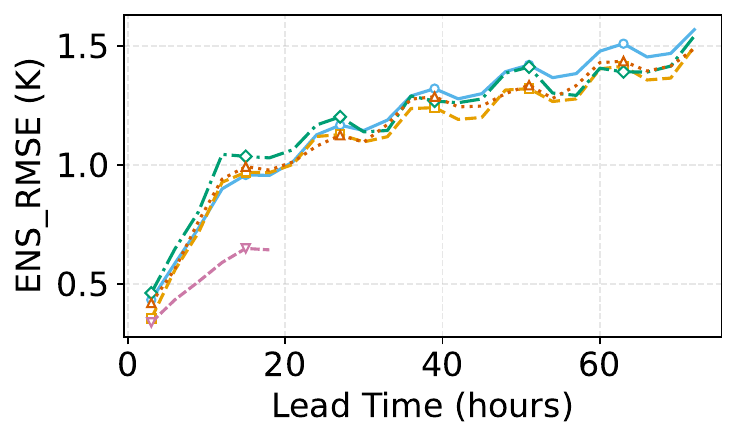}
        \caption{\gls{rmse} for \texttt{t2m}}
        \label{fig:t2m_rmse}
    \end{subfigure}%
    \hfill%
    \begin{subfigure}[b]{0.33\textwidth}
        \centering
        \includegraphics[width=\textwidth]{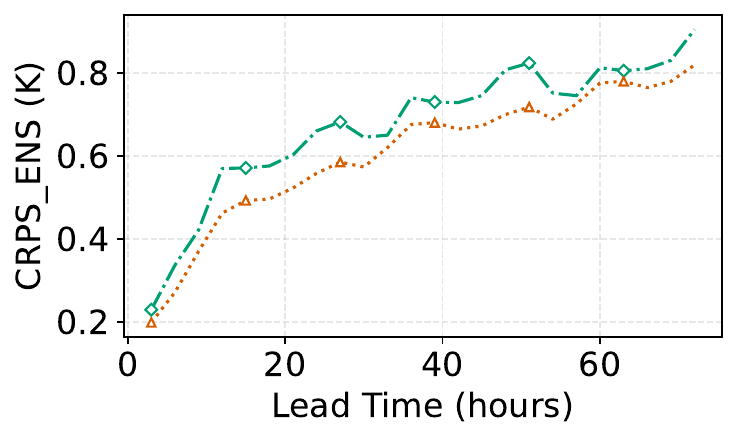}
        \caption{\gls{crps} for \texttt{t2m}}
        \label{fig:t2m_crps}
    \end{subfigure}%
    \hfill%
    \begin{subfigure}[b]{0.33\textwidth}
        \centering
        \includegraphics[width=\textwidth]{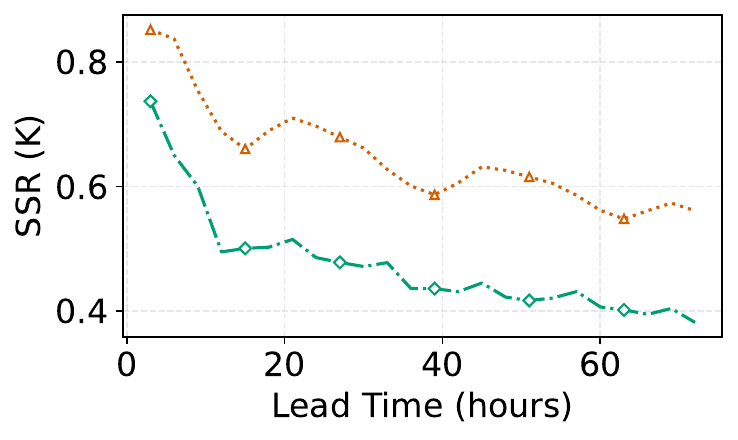}
        \caption{\gls{ssr} for \texttt{t2m}}
        \label{fig:t2m_ssr}
    \end{subfigure}
    \begin{subfigure}[b]{0.33\textwidth}
        \centering
        \includegraphics[width=\textwidth]{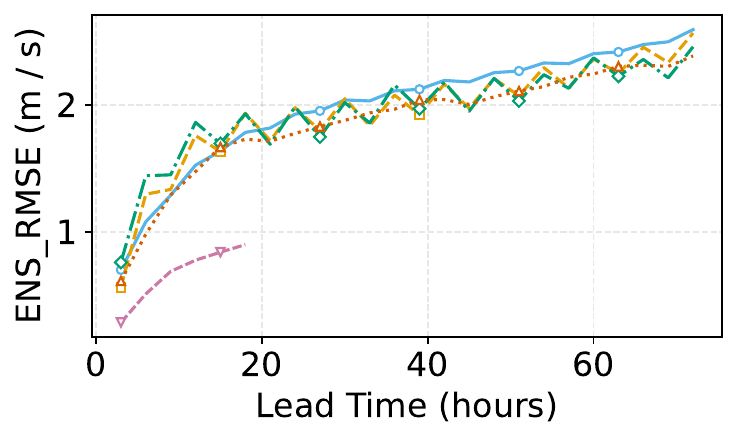}
        \caption{\gls{rmse} for \texttt{u10m}}
        \label{fig:u10m_rmse}
    \end{subfigure}%
    \hfill%
    \begin{subfigure}[b]{0.33\textwidth}
        \centering
        \includegraphics[width=\textwidth]{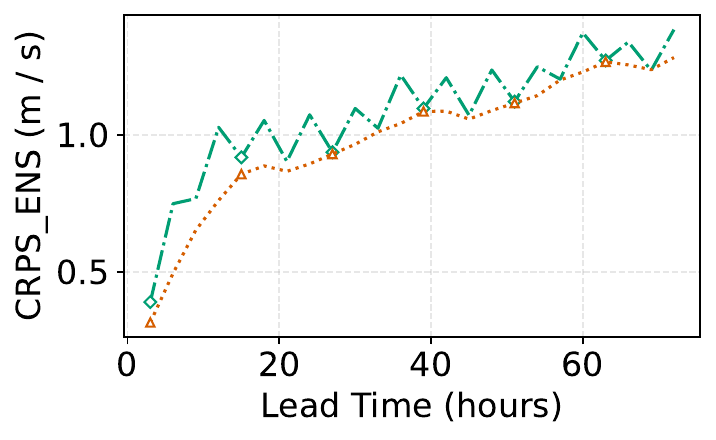}
        \caption{\gls{crps} for \texttt{u10m}}
        \label{fig:u10m_crps}
    \end{subfigure}%
    \hfill%
    \begin{subfigure}[b]{0.33\textwidth}
        \centering
        \includegraphics[width=\textwidth]{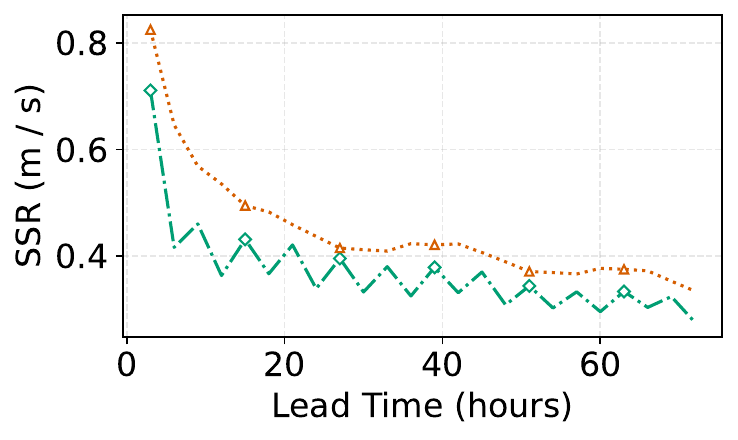}
        \caption{\gls{ssr} for \texttt{u10m}}
        \label{fig:u10m_ssr}
    \end{subfigure}%
    \caption{The \gls{rmse}, \gls{crps}, and \gls{ssr} for \SI{2}{\meter} temperature and the u wind component at \SI{10}{\meter} on the DANRA dataset with IFS boundary conditions.}
    \label{fig:quantitative_results_danra_IFS}
\end{figure}

Here, we provide a more detailed evaluation of the models on the DANRA dataset. We evaluate the \gls{rmse}, \gls{crps}, \gls{ssr}, and \gls{lsd} for all surface variables, along with all variables at the \SI{600}{\hecto\pascal} pressure level, is shown in \cref{fig:DANRA_rmse_full,fig:DANRA_crps_full,fig:DANRA_ssr_full,fig:DANRA_lsd_full}, respectively. We also evaluate the models trained solely with ERA5 boundary conditioning using IFS forecast boundaries in \cref{fig:quantitative_results_danra_IFS}. The resulting performance degradation motivated the additional fine-tuning experiments using IFS forecast boundaries. In \cref{apx:danra_forecasts}, we present \SI{72}{\hour} lead time forecasts for only a few variables to limit the length of the appendix.

\begin{figure}[tbph]
    \centering%
    \includegraphics[width=\textwidth]{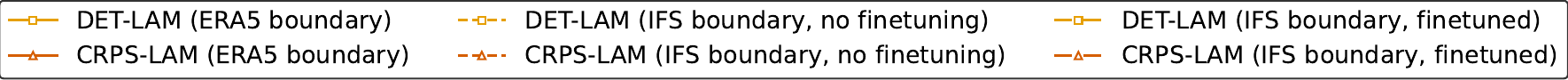}
    \begin{subfigure}[b]{0.33\textwidth}
        \centering
        \includegraphics[width=\textwidth]{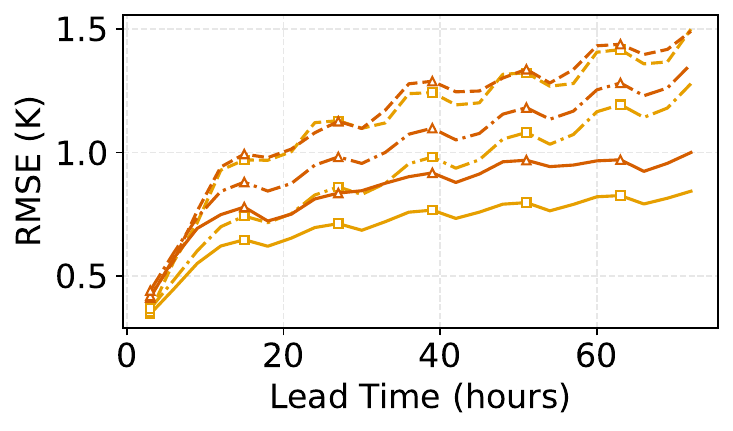}
        \caption{\gls{rmse} for \texttt{t2m}}
        \label{fig:t2m_rmse}
    \end{subfigure}%
    \hfill%
    \begin{subfigure}[b]{0.33\textwidth}
        \centering
        \includegraphics[width=\textwidth]{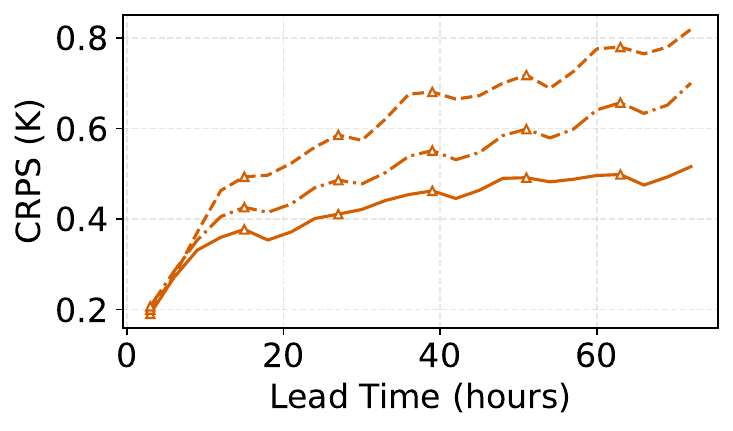}
        \caption{\gls{crps} for \texttt{t2m}}
        \label{fig:t2m_crps}
    \end{subfigure}%
    \hfill%
    \begin{subfigure}[b]{0.33\textwidth}
        \centering
        \includegraphics[width=\textwidth]{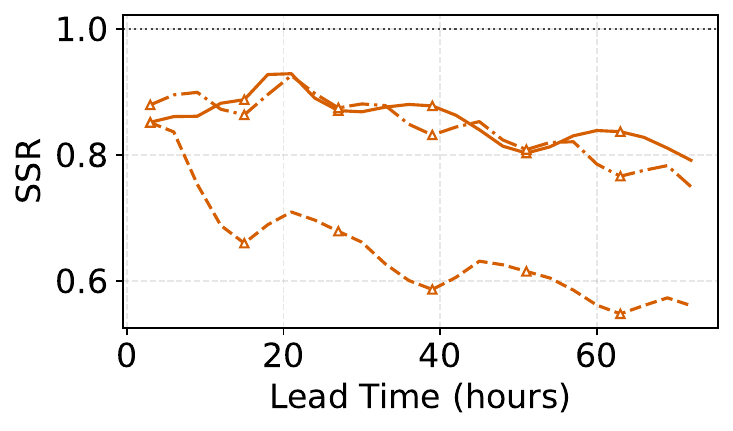}
        \caption{\gls{ssr} for \texttt{t2m}}
        \label{fig:t2m_ssr}
    \end{subfigure}
    \begin{subfigure}[b]{0.33\textwidth}
        \centering
        \includegraphics[width=\textwidth]{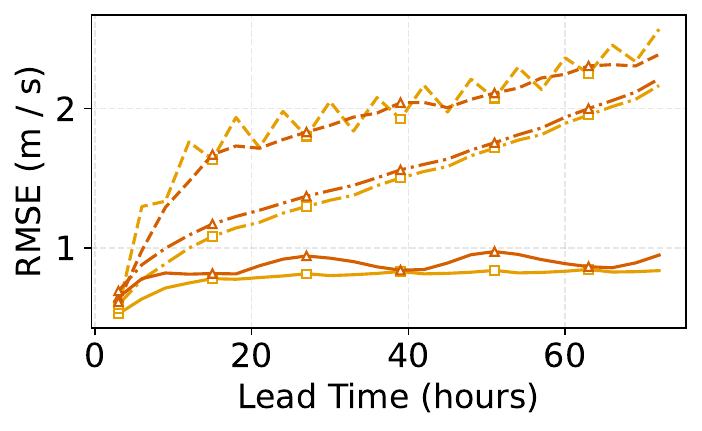}
        \caption{\gls{rmse} for \texttt{u10m}}
        \label{fig:u10m_rmse}
    \end{subfigure}%
    \hfill%
    \begin{subfigure}[b]{0.33\textwidth}
        \centering
        \includegraphics[width=\textwidth]{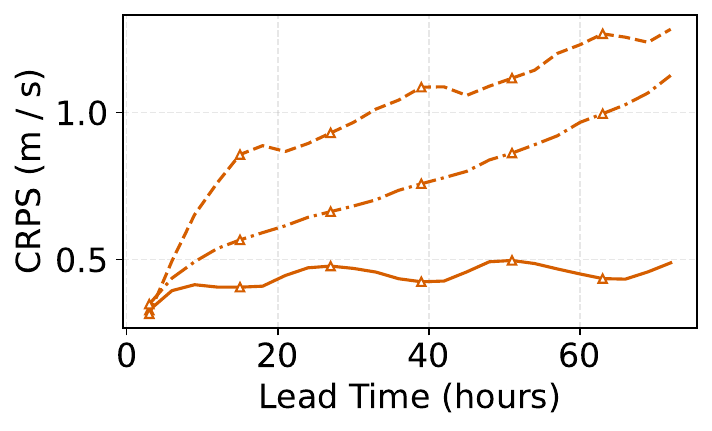}
        \caption{\gls{crps} for \texttt{u10m}}
        \label{fig:u10m_crps}
    \end{subfigure}%
    \hfill%
    \begin{subfigure}[b]{0.33\textwidth}
        \centering
        \includegraphics[width=\textwidth]{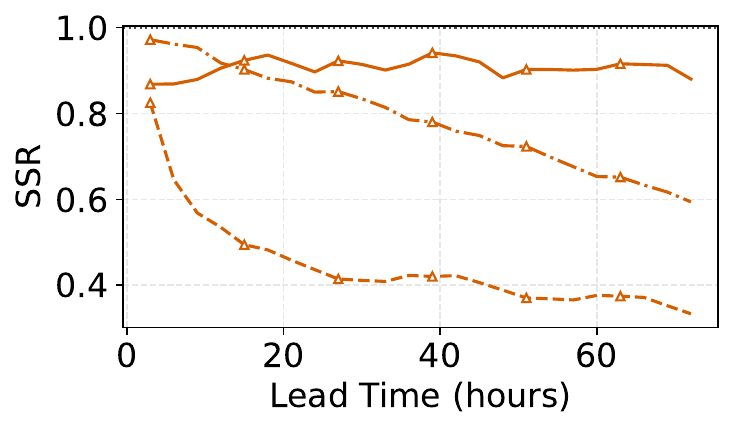}
        \caption{\gls{ssr} for \texttt{u10m}}
        \label{fig:u10m_ssr}
    \end{subfigure}%
    \caption{\Gls{rmse}, \gls{crps}, and \gls{ssr} for \SI{2}{\meter} temperature and the \SI{10}{\meter} u-wind component on the DANRA dataset. Results are shown for DET-LAM and CRPS-LAM under different boundary-condition configurations, both with and without fine-tuning.}
    \label{fig:quantitative_results_danra_finetuing_boundary_ablation}
\end{figure}

\subsection{Forecasts on the DANRA dataset}\label{apx:danra_forecasts}
\begin{figure}[H]
    \centering
    \includegraphics[width=1.0\linewidth]{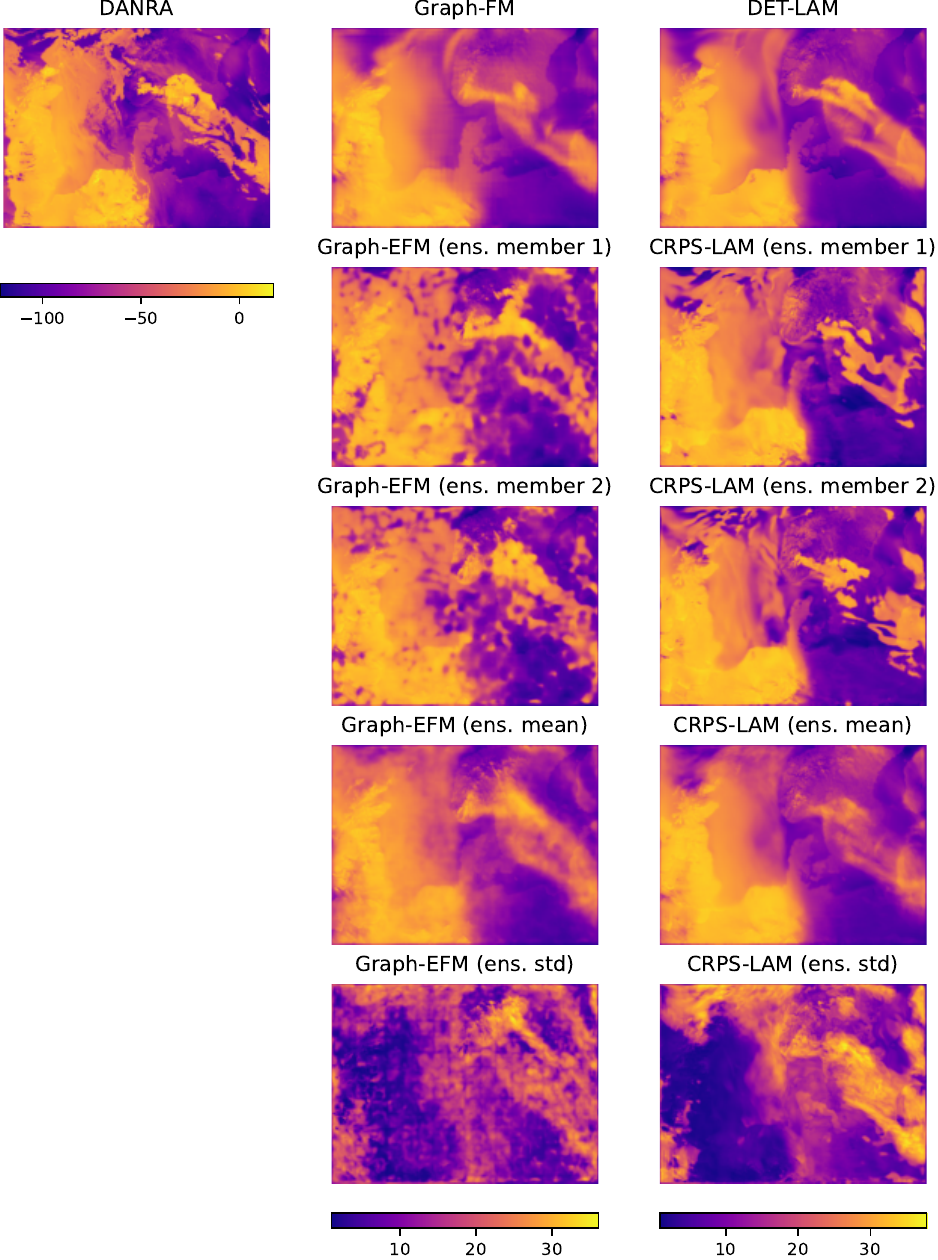}
    \caption{Forecasts of net long wave radiation flux at \SI{72}{\hour} lead time.}
    \label{fig:DANRA_forecast_apx_lwavr0m_24}
\end{figure}
\begin{figure}[H]
    \centering
    \includegraphics[width=1.0\linewidth]{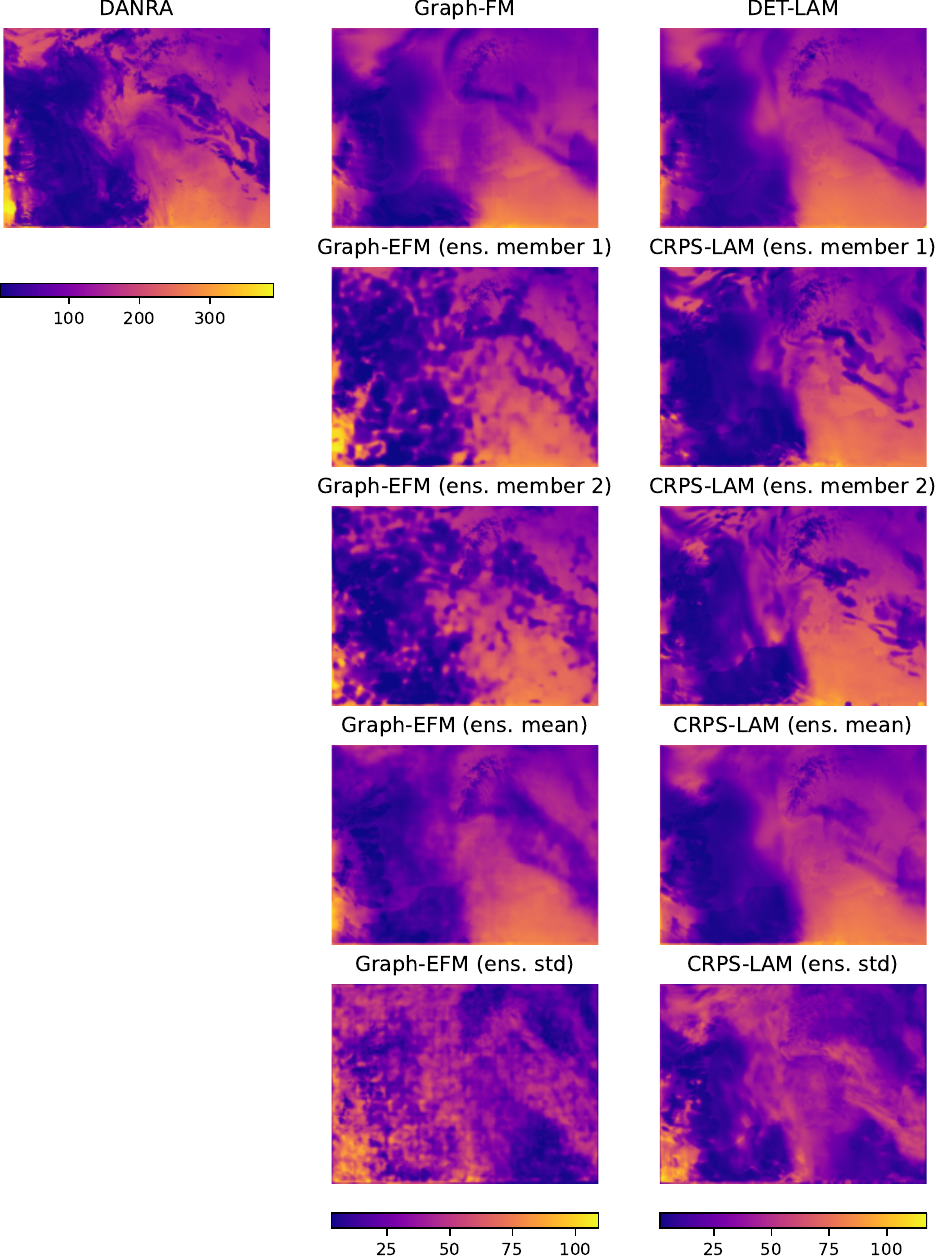}
    \caption{Forecasts of net short wave radiation flux at \SI{72}{\hour} lead time.}
    \label{fig:DANRA_forecast_apx_swavr0m_24}
\end{figure}
\begin{figure}[H]
    \centering
    \includegraphics[width=1.0\linewidth]{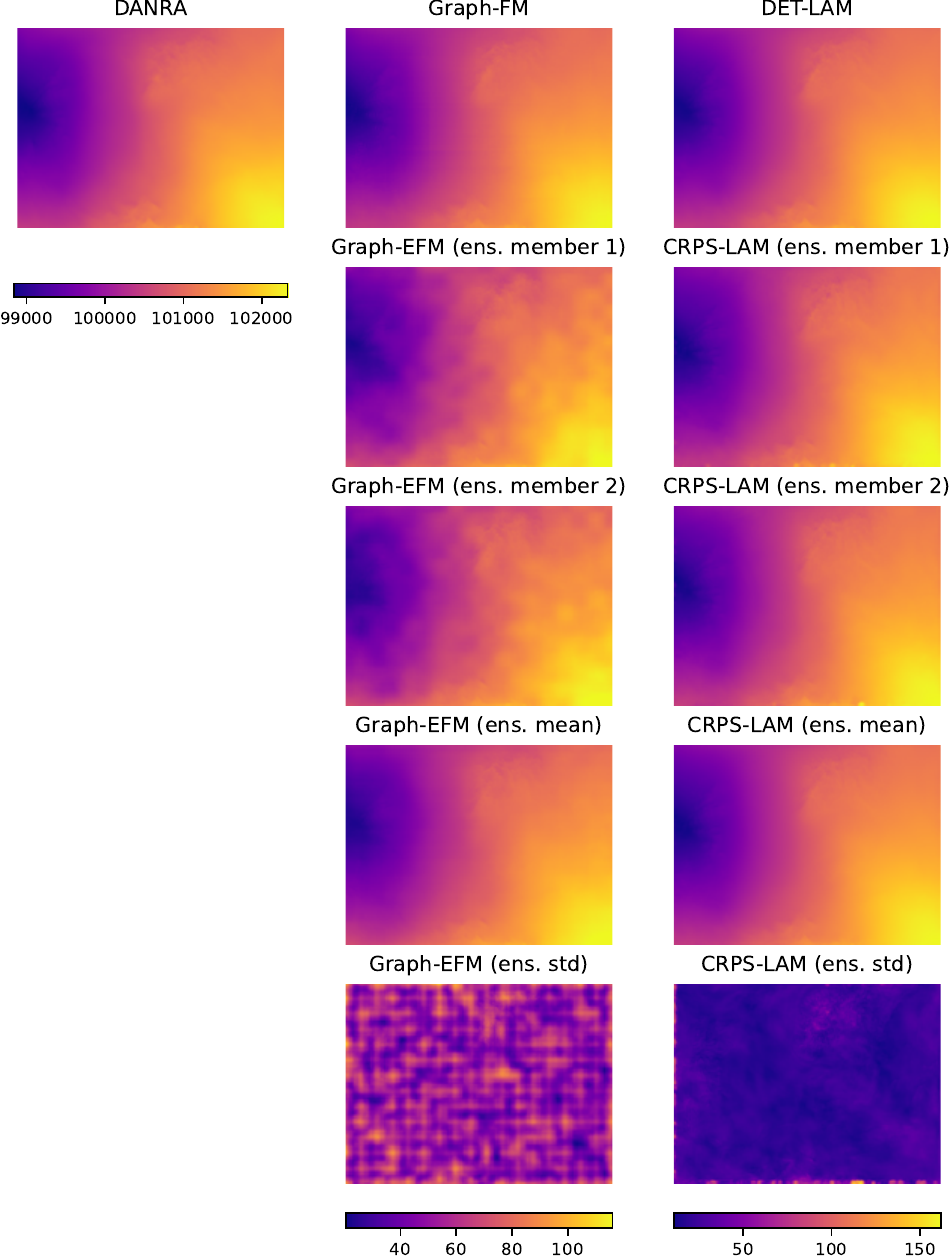}
    \caption{Forecasts of mean sea level pressure at \SI{72}{\hour} lead time.}
    \label{fig:DANRA_forecast_apx_pres_seasurface_24}
\end{figure}
\begin{figure}[H]
    \centering
    \includegraphics[width=1.0\linewidth]{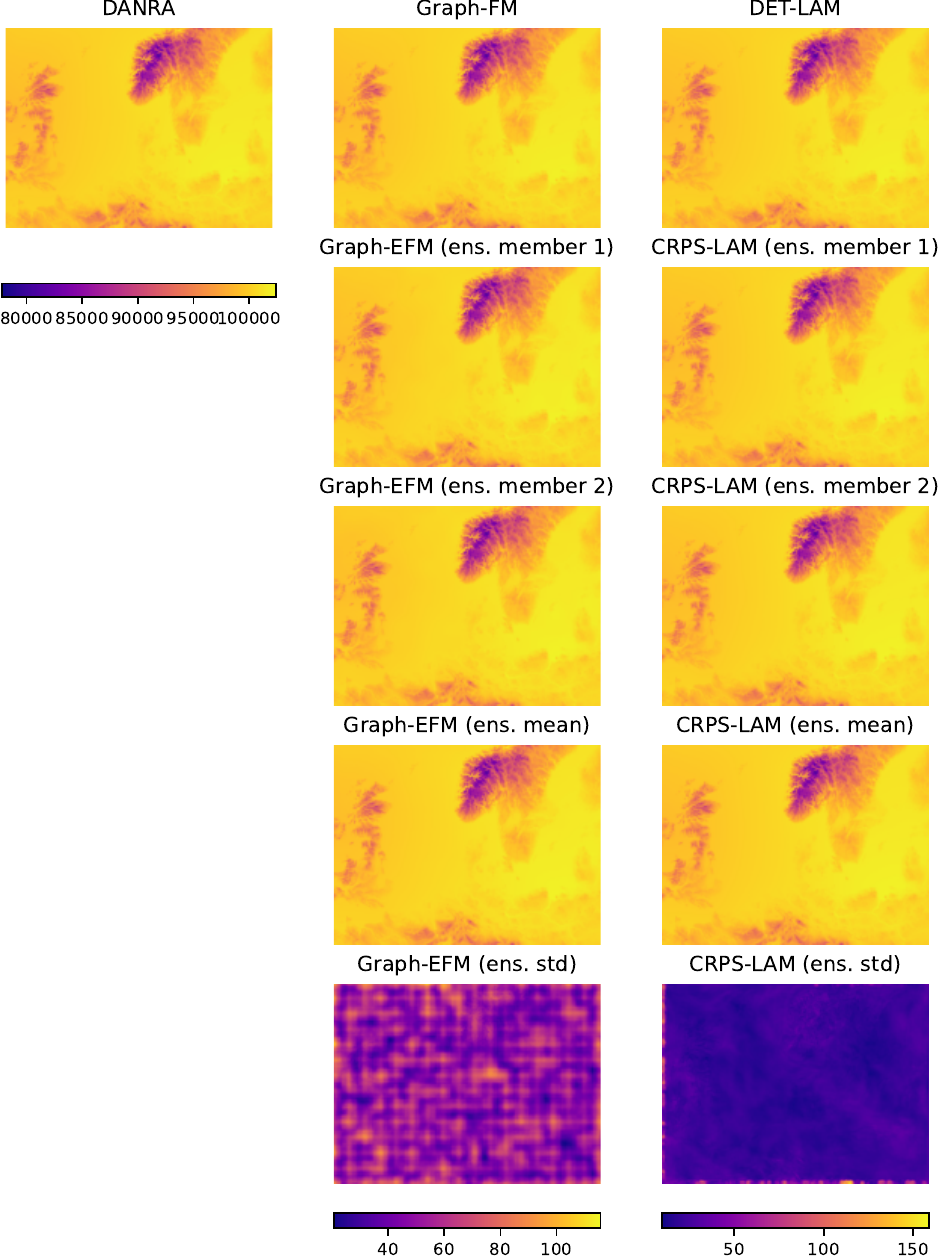}
    \caption{Forecasts of surface pressure at \SI{72}{\hour} lead time.}
    \label{fig:DANRA_forecast_apx_pres0m_24}
\end{figure}
\begin{figure}[H]
    \centering
    \includegraphics[width=1.0\linewidth]{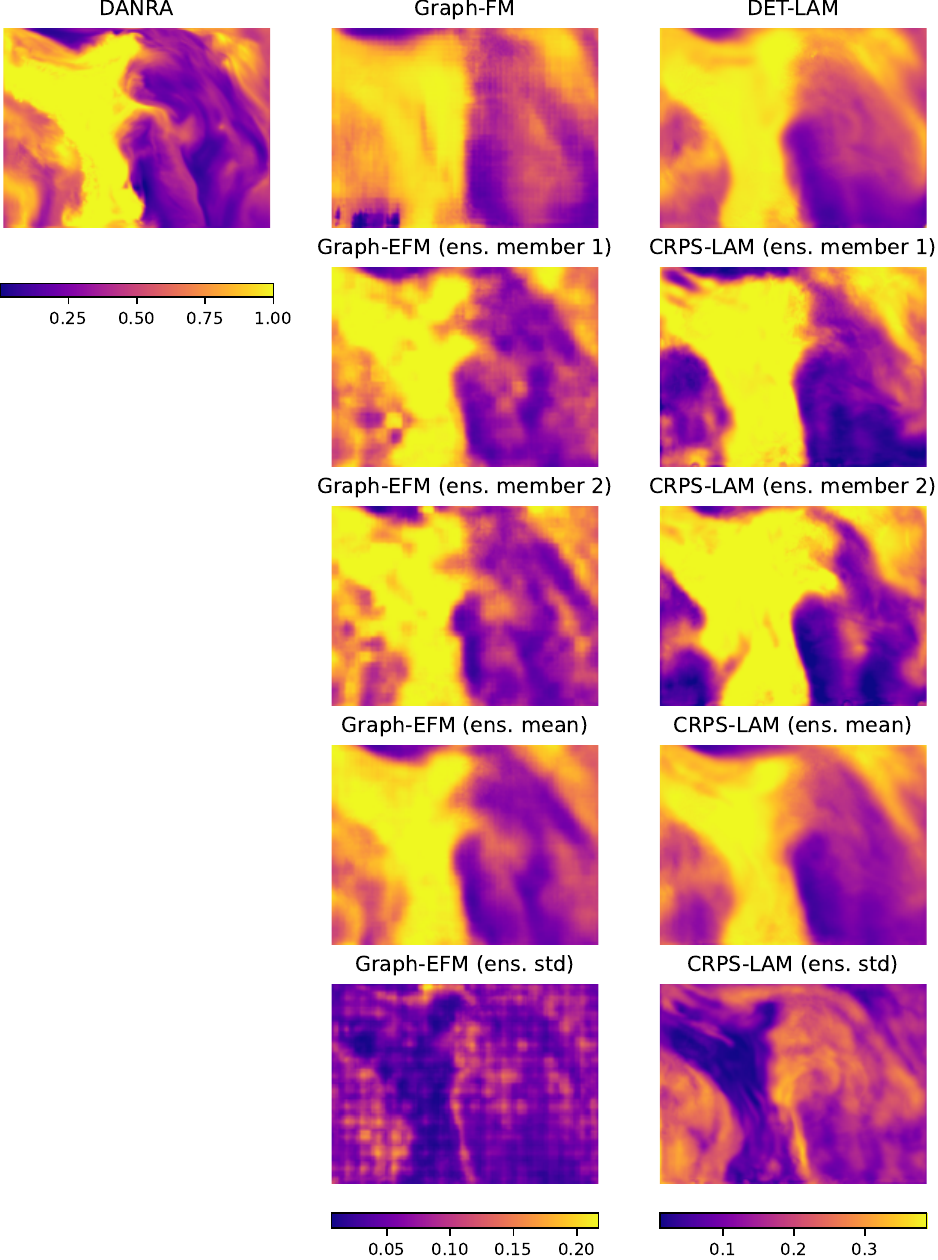}
    \caption{Forecasts of relative humidity at \SI{600}{\hecto\pascal} at \SI{72}{\hour} lead time.}
    \label{fig:DANRA_forecast_apx_r600_24}
\end{figure}
\begin{figure}[H]
    \centering
    \includegraphics[width=1.0\linewidth]{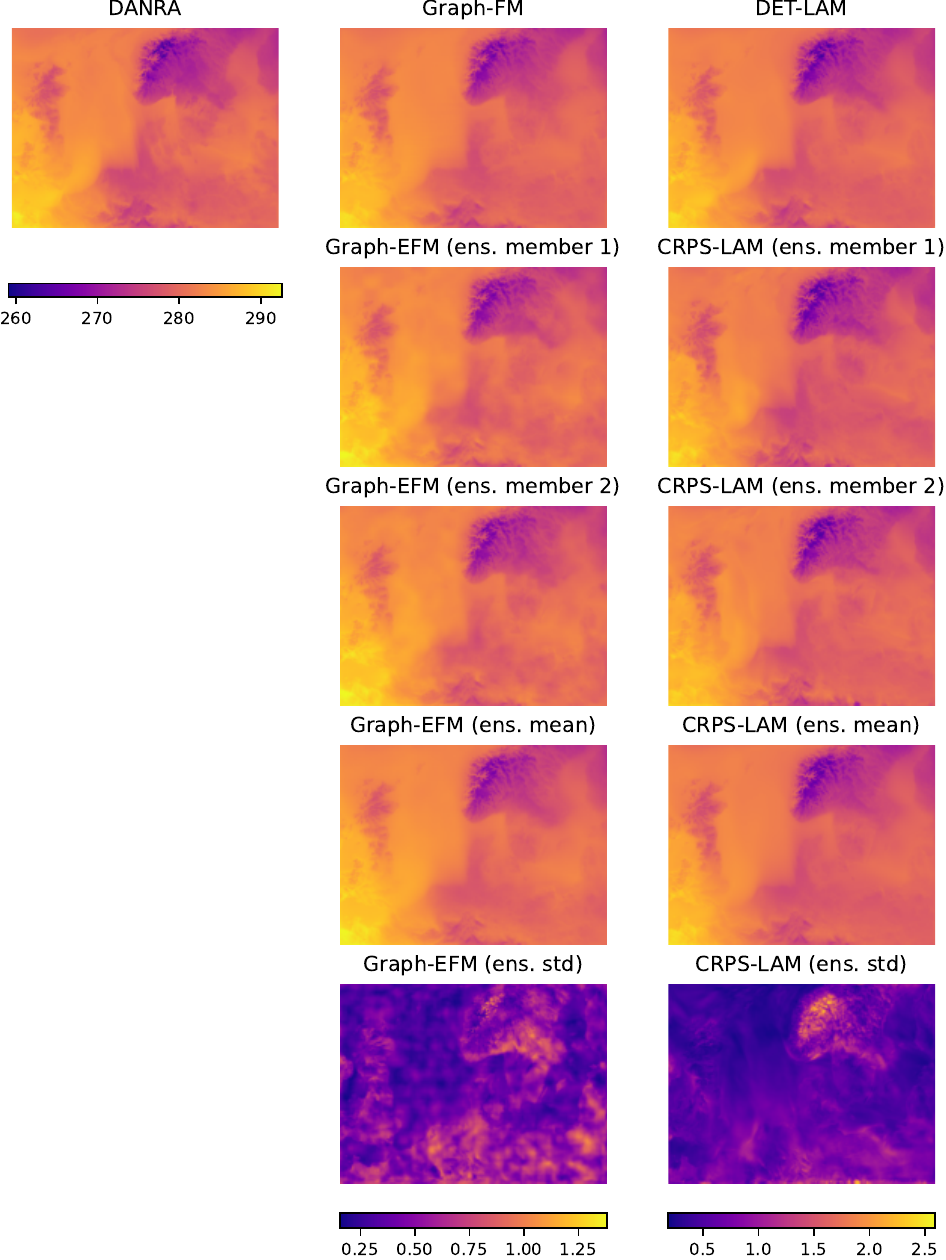}
    \caption{Forecasts of \SI{2}{\meter} temperature at \SI{72}{\hour} lead time.}
    \label{fig:DANRA_forecast_apx_t2m_24}
\end{figure}
\begin{figure}[H]
    \centering
    \includegraphics[width=1.0\linewidth]{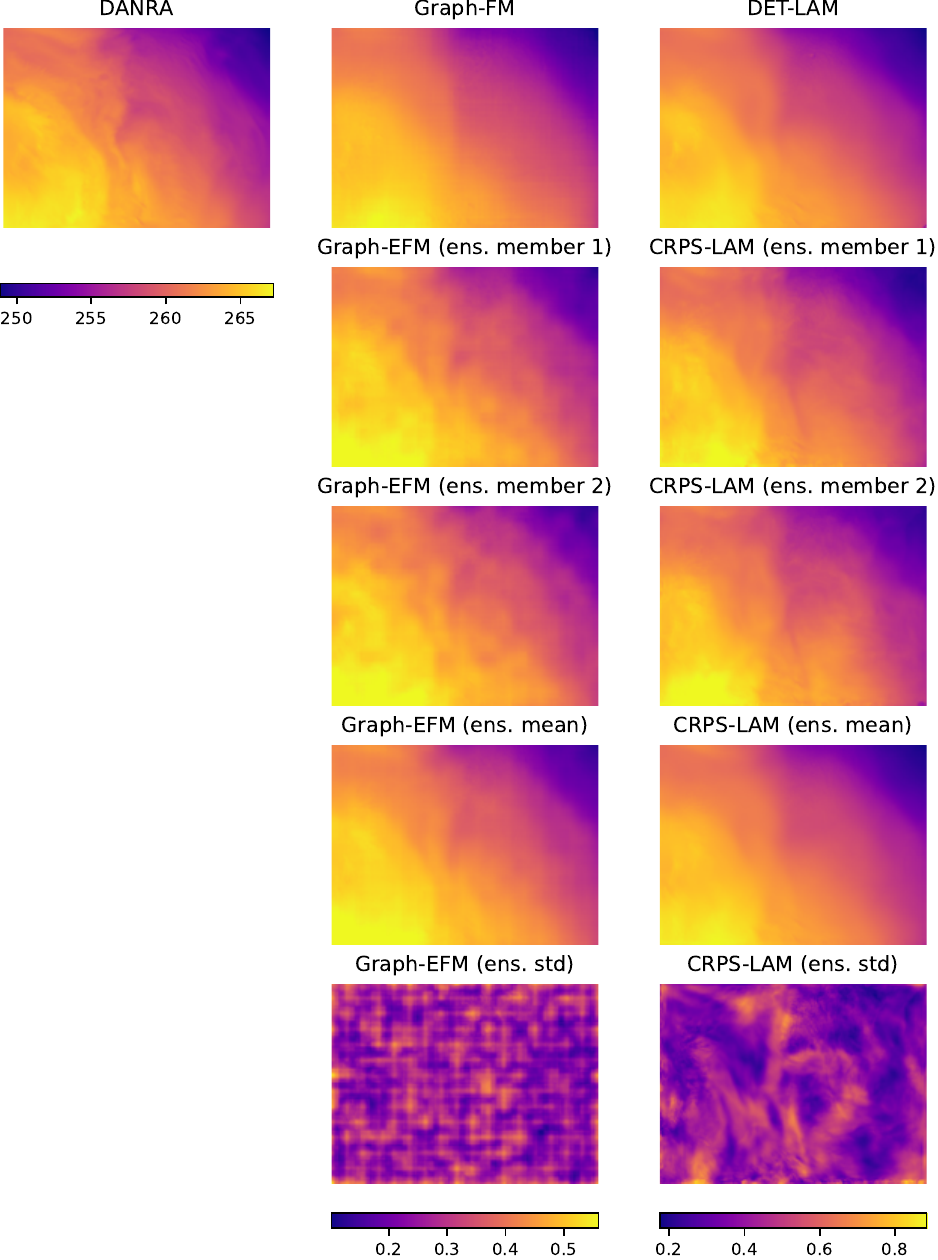}
    \caption{Forecasts of \SI{2}{\meter} temperature at \SI{600}{\hecto\pascal} at \SI{72}{\hour} lead time.}
    \label{fig:DANRA_forecast_apx_t600_24}
\end{figure}
\begin{figure}[H]
    \centering
    \includegraphics[width=1.0\linewidth]{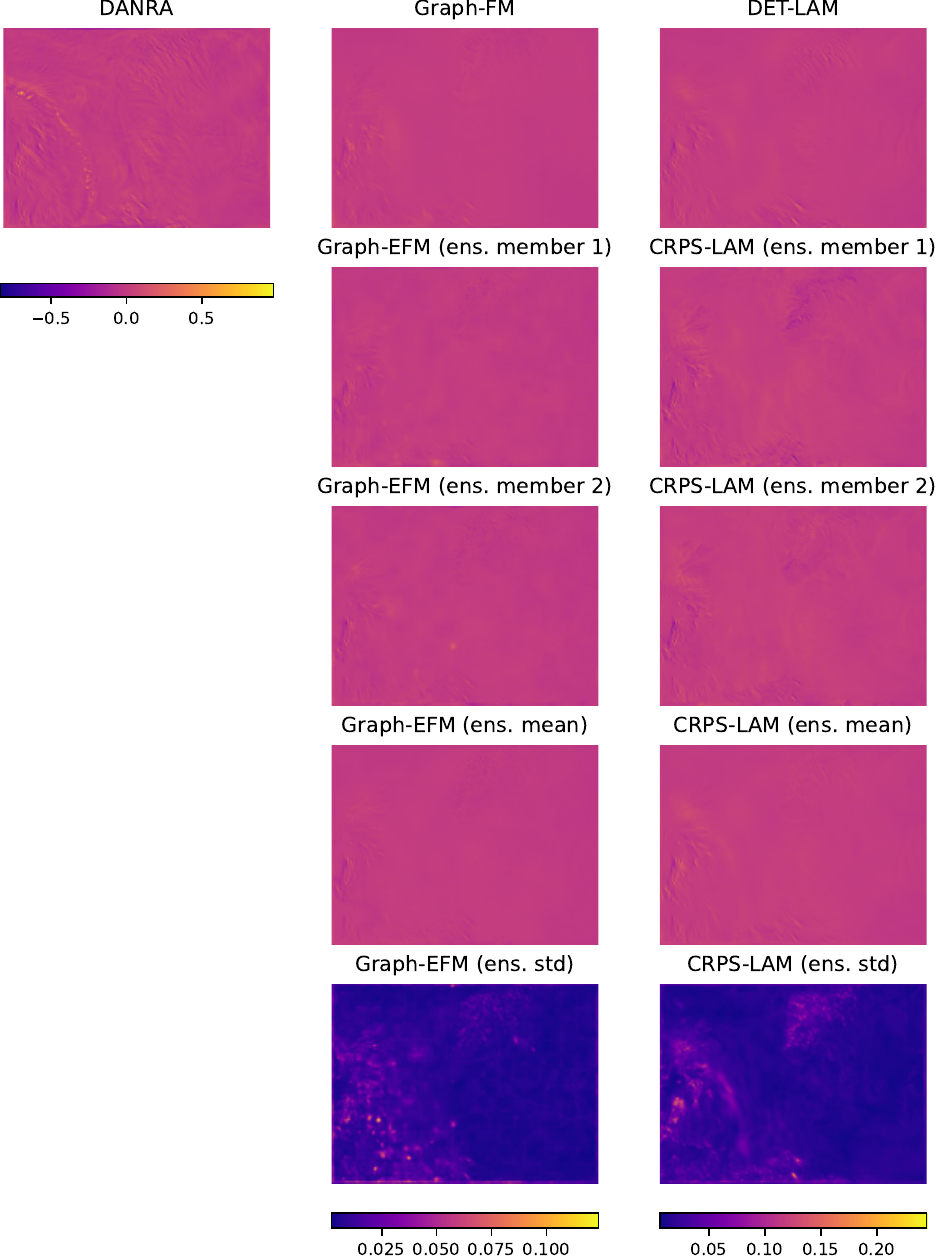}
    \caption{Forecasts of vertical velocity at \SI{600}{\hecto\pascal} at \SI{72}{\hour} lead time.}
    \label{fig:DANRA_forecast_apx_tw600_24}
\end{figure}
\begin{figure}[H]
    \centering
    \includegraphics[width=1.0\linewidth]{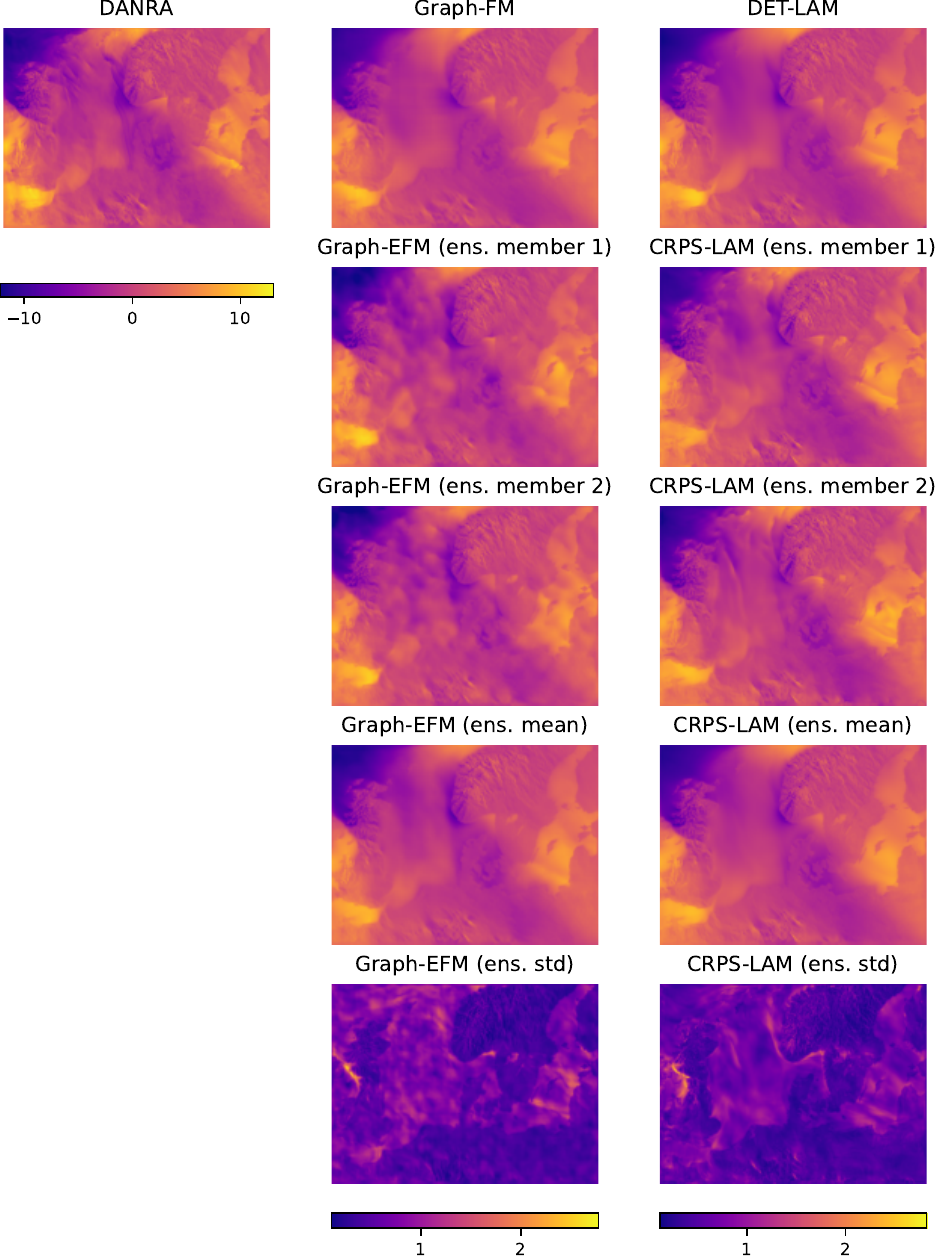}
    \caption{Forecasts of u wind component at \SI{10}{\meter} at \SI{72}{\hour} lead time.}
    \label{fig:DANRA_forecast_apx_u10m_24}
\end{figure}
\begin{figure}[H]
    \centering
    \includegraphics[width=1.0\linewidth]{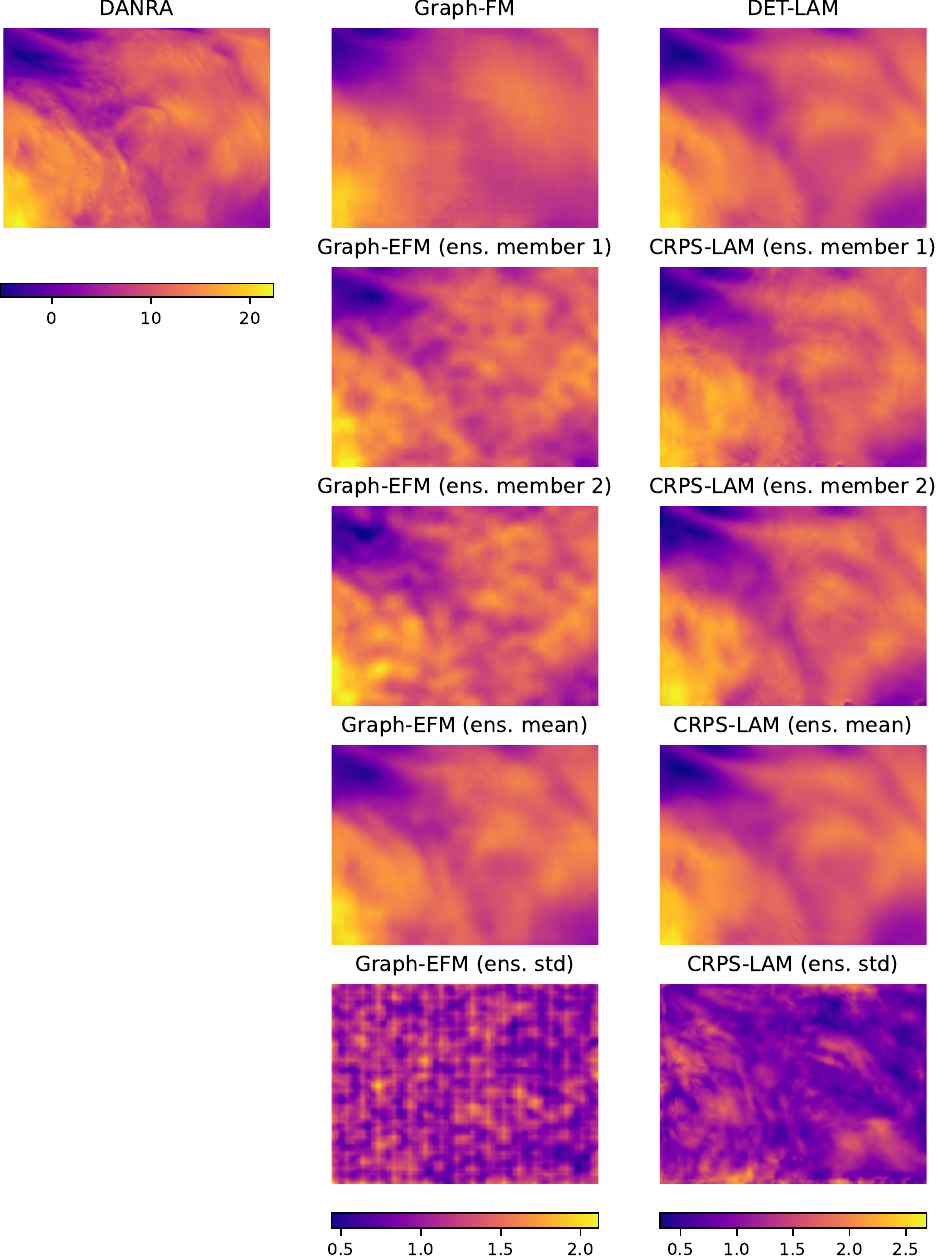}
    \caption{Forecasts of u wind component at \SI{600}{\hecto\pascal} at \SI{72}{\hour} lead time.}
    \label{fig:DANRA_forecast_apx_u600_24}
\end{figure}
\begin{figure}[H]
    \centering
    \includegraphics[width=1.0\linewidth]{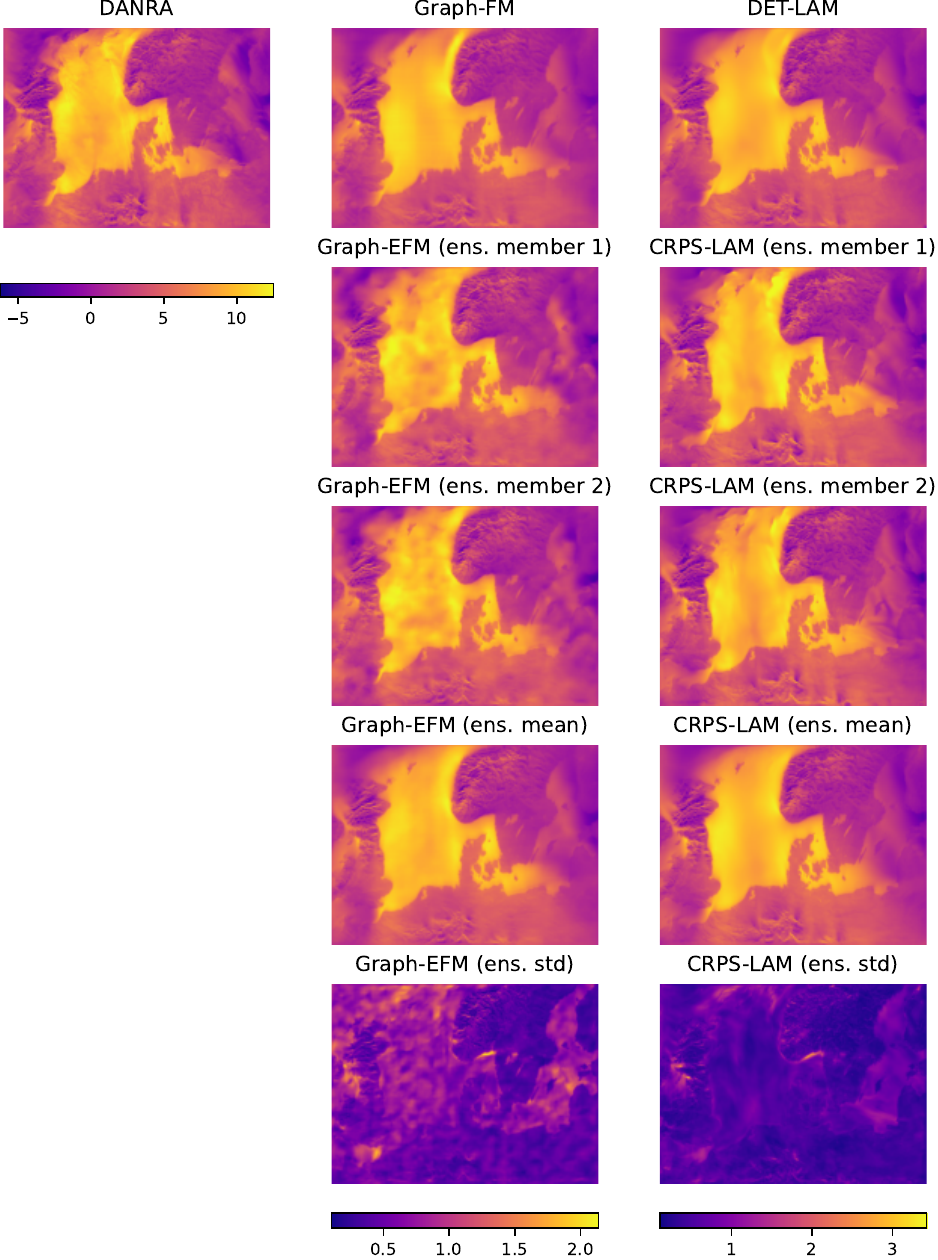}
    \caption{Forecasts of v wind component at \SI{10}{\meter} at \SI{72}{\hour} lead time.}
    \label{fig:DANRA_forecast_apx_v10m_24}
\end{figure}
\begin{figure}[H]
    \centering
    \includegraphics[width=1.0\linewidth]{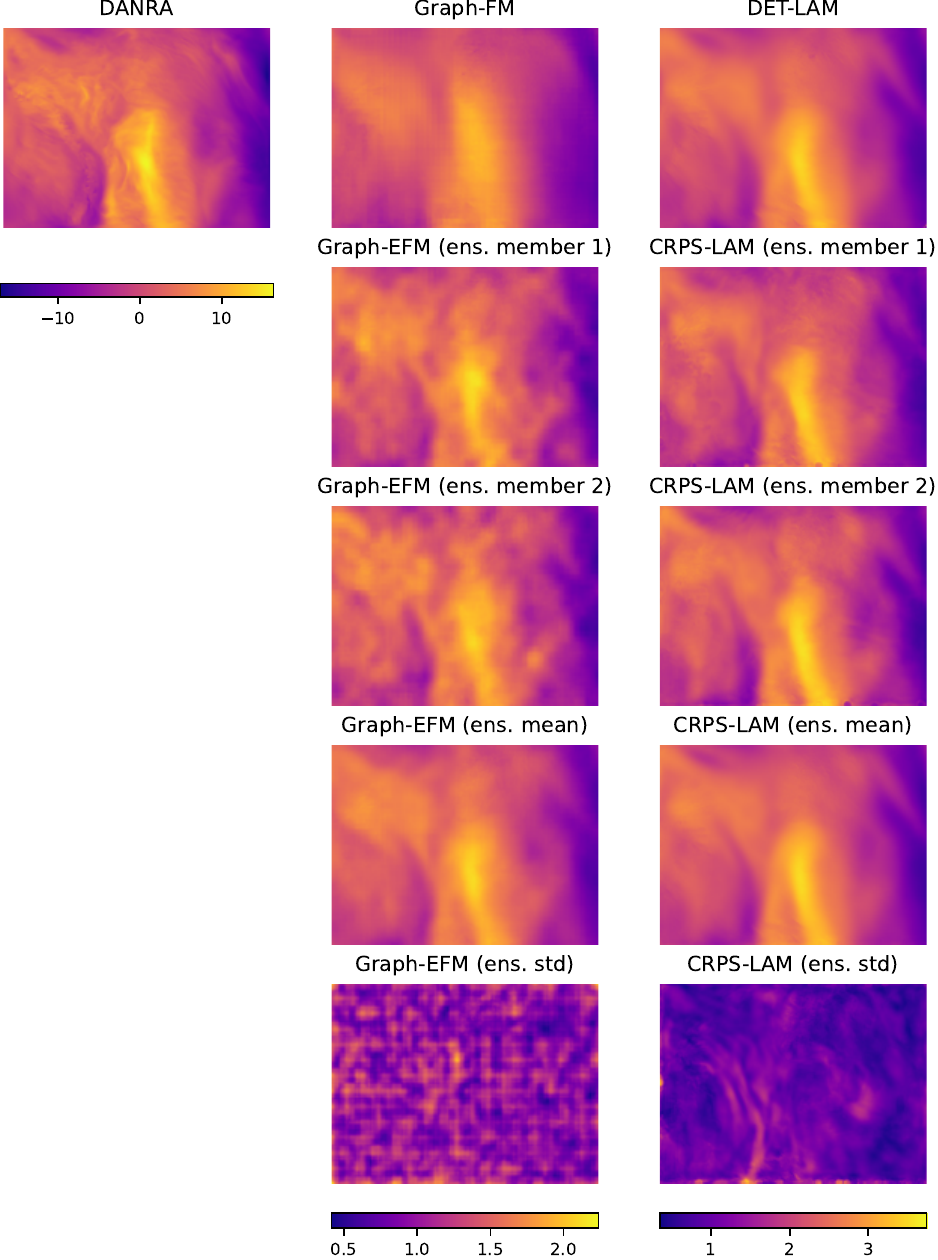}
    \caption{Forecasts of v wind component at \SI{600}{\hecto\pascal} at \SI{72}{\hour} lead time.}
    \label{fig:DANRA_forecast_apx_v600_24}
\end{figure}
\begin{figure}[H]
    \centering
    \includegraphics[width=1.0\linewidth]{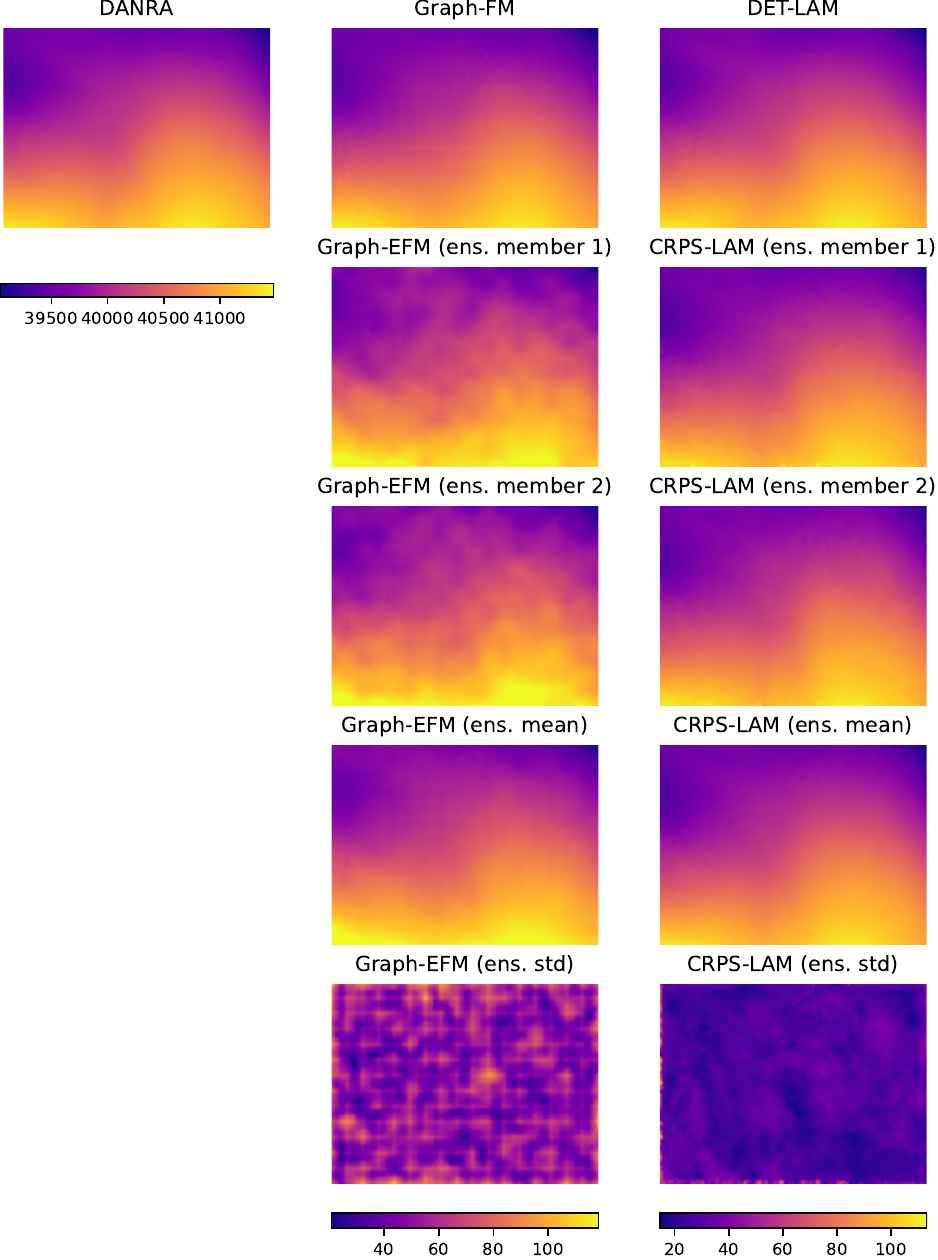}
    \caption{Forecasts of geopotential at \SI{600}{\hecto\pascal} at \SI{72}{\hour} lead time.}
    \label{fig:DANRA_forecast_apx_z600_24}
\end{figure}

\section{Additional Experiments on the MEPS dataset}\label{apx:MEPS}
Here we present details of the experiments\footnote{The code and implementation details will be made publicly available upon acceptance of this paper.
} in a simplified setting using the MEPS dataset\footnote{The MEPS dataset is openly available at \url{https://nextcloud.liu.se/s/meps} under a CC BY 4.0 license.}, including a comparison with a diffusion-based baseline \citep{larsson2025diffusionlam} and Graph-EFM. The problem definition is slightly different compared to to the DANRA problem definition presented in \cref{sec:experiments} as the boundary comes from the ground truth on the same grid and is not overlapping with the interior. An overview of the limited-area MEPS forecasting setup is illustrated in \cref{fig:model_process}.

Since the boundary and interior are on a regular grid, the grid-to-mesh encoders reduce to an MLP (with different weights for the interior and boundary), since there is no need for a GNN to map an irregular grid to the mesh.

\subsection{The MEPS Dataset}\label{apx:dataset_details}
Since the training objective is based on forecasts rather than actual observations, the objective is to develop an emulator model for MEPS. The \SI{6069} forecasts in the dataset are from the time period April 2021 to March 2023. For simplicity and consistency with \citet{oskarsson2024probabilistic} we use the same training, validation and test split. We use forecasts from April 2021 to June 2022 for training (\SI{2713} samples) and validation (\SI{678} samples), and forecasts from July 2022 to March 2023 for testing (\SI{2678} samples).

\begin{table}[h]
\caption{Variables in the MEPS dataset. \textsuperscript{*}Level 65 in the MEPS system is approximately \SI{12.5}{\meter} over the ground \citep{arome_metcoop}.}
\label{tab:dataset_variables}
\begin{center}
\begin{tabular}{lccc}
\multicolumn{1}{l}{\bf Description}  & \multicolumn{1}{c}{\bf Abbreviation} & \multicolumn{1}{c}{\bf Unit} & \multicolumn{1}{c}{\bf Residual standard deviation}
\\ \hline \\
Net longwave solar radiation flux at the surface & nlwrs & \si{\watt\per\metre\squared} & $0.0583$ \\
Net shortwave solar radiation flux at the surface & nswrs & \si{\watt\per\metre\squared} & $0.0583$ \\
Atmospheric pressure at ground level & pres\_0g & \si{\pascal} & $0.6399$ \\
Atmospheric pressure at sea level & pres\_0s & \si{\pascal} & $0.7608$ \\
Relative humidity at \SI{2}{\meter} & r\_2 & [0, 1] & $0.5534$ \\
Relative humidity at level 65\textsuperscript{*} & r\_65 & [0, 1] & $0.5371$ \\
Temperature at \SI{2}{\meter} & t\_2 & \si{\kelvin} & $0.2197$ \\
Temperature at level 65\textsuperscript{*} & t\_65 & \si{\kelvin} & $0.1950$ \\
Temperature at \SI{500}{\hecto\pascal} & t\_500 & \si{\kelvin} & $0.1319$ \\
Temperature at \SI{850}{\hecto\pascal} & t\_850 & \si{\kelvin} & $0.1294$ \\
$u$-component of wind at level 65\textsuperscript{*} & u\_65 & \si{\meter\per\second} & $0.3885$ \\
$u$-component of wind at \SI{850}{\hecto\pascal} & u\_850 & \si{\meter\per\second} & $0.3530$ \\
$v$-component of wind at level 65\textsuperscript{*} & v\_65 & \si{\meter\per\second} & $0.3815$ \\
$v$-component of wind at \SI{850}{\hecto\pascal} & v\_850 & \si{\meter\per\second} & $0.3861$ \\
Water vapor for the full integrated column & wvint\_0 & \si{\kilo\gram\per\meter\squared} & $0.2473$ \\
Geopotential at \SI{1000}{\hecto\pascal} & z\_1000 & \si{\meter\squared\per\second\squared} & $0.1202$ \\
Geopotential at \SI{500}{\hecto\pascal} & z\_500 & \si{\meter\squared\per\second\squared} & $0.0720$ \\
\end{tabular}
\end{center}
\end{table}

\begin{table}[h]
\caption{Forcing features in the MEPS dataset.}
\label{tab:dataset_forcing}
\begin{center}
\begin{tabular}{lcc}
\multicolumn{1}{l}{\bf Description}  & \multicolumn{1}{c}{\bf Abbreviation} & \multicolumn{1}{c}{\bf Unit}
\\ \hline \\
Solar radiation flux at the top of the atmosphere & toa & \si{\watt\per\meter\squared} \\
Fraction of open water at the surface & water & [0, 1] \\
Sine-encoded time of day & sin\_tod & [0, 1] \\
Cosine-encoded time of day & cos\_tod & [0, 1] \\
Sine-encoded time of year & sin\_toy & [0, 1] \\
Cosine-encoded time of year & cos\_toy & [0, 1] \\
\end{tabular}
\end{center}
\end{table}

\begin{table}[h]
\caption{Static features for each grid position in the MEPS dataset.}
\label{tab:dataset_static}
\begin{center}
\begin{tabular}{lcc}
\multicolumn{1}{l}{\bf Description}  & \multicolumn{1}{c}{\bf Abbreviation} & \multicolumn{1}{c}{\bf Unit}
\\ \hline \\
Topology (geopotential at the surface) & topology & \si{\meter\squared\per\second\squared} \\
x-coordinate in the MEPS projection & x\_coord & [0, 1] \\
y-coordinate in the MEPS projection & y\_coord & [0, 1] \\
Boundary mask (indicating which pixels belong to the border) & border\_mask & 0/1 \\
Interior mask (indicating which pixels belong to the interior) & interior\_mask & 0/1 \\
\end{tabular}
\end{center}
\end{table}

\subsection{Experiment Details}
All forecast trajectories are initialized from the ground-truth states and generated autoregressively up to a lead time of \SI{57}{\hour}, with 25 ensemble members produced for each model. The model is trained to perform \SI{3}{\hour} forecasts (see \cref{fig:model_process}) and is rolled out autoregressively to produce longer forecasts. CRPS-LAM can generate \SI{57}{\hour} forecasts in approximately \SI{0.5}{\second} per ensemble member on a single A100 GPU. Moreover, an arbitrary number of ensemble members can be sampled in parallel through batched inference on one or multiple GPUs. This corresponds to a sampling speed comparable to that of Graph-EFM or a deterministic model (on a per-member basis) and approximately $39\times$ faster than Diffusion-LAM, depending on the number of solver steps that are used.

\begin{figure}[h]
\begin{center}
\includegraphics[width=\textwidth]{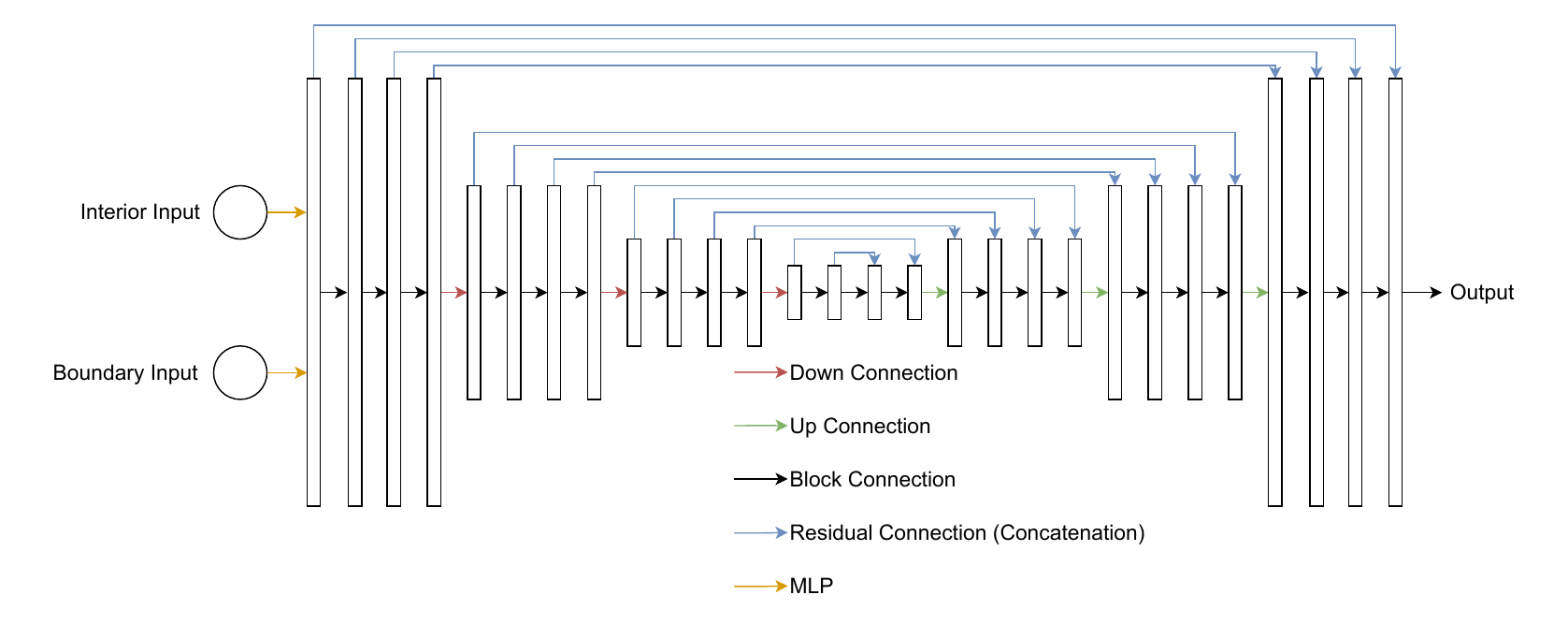}
\end{center}
\caption{An overview of the backbone model.}
\label{fig:unet}
\end{figure}

\begin{figure}[h]
\begin{center}
\includegraphics[width=\textwidth]{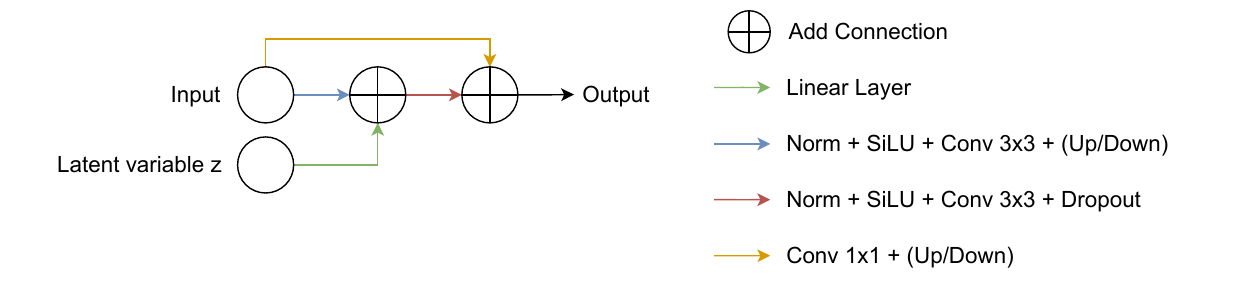}
\end{center}
\caption{A description of each U-Net block.}
\label{fig:unet_block}
\vspace{-1em} 
\end{figure}

An overview schematic of our architecture of our backbone model is shown in \cref{fig:unet} with the details of each U-Net shown in \cref{fig:unet_block}. Note that the MLP blocks are different from the U-Net blocks and consist of a two-layer MLP followed by a conditional layer normalization, where the latent variable $z$ is included as an additional conditioning input.

The training procedure for CRPS-LAM is summarized in \cref{tab:training_steps}. We begin by training the model on single-step forecasts until convergence, after which we extend the training to two-step autoregressive rollouts. In this setup, we did not observe any additional benefits from training with more autoregressive steps.

\begin{table}[h!]
\centering
\caption{Training schedule.}
\label{tab:training_steps}
\begin{tabular}{lcc}
\toprule
\textbf{Epochs} & \textbf{Learning Rate} & \textbf{Autoregressive Steps} \\
\midrule
600  & 0.001   & 1 \\
400  & 0.0001  & 1 \\
200  & 0.00001 & 2 \\
\bottomrule
\end{tabular}
\end{table}

During the initial training phase, we sometimes observe instability when training with the unbiased \gls{crps} estimator. Similar behavior has been reported by \citet{lang2024aifscrpsensembleforecastingusing}, who proposed a modified training objective to address it, and by \citet{fourcastnet3}, who mitigated the issue by using a biased \acrshort{crps} estimator with a large ensemble early in training, later transitioning to a smaller ensemble and the unbiased fair \acrshort{crps} estimator. 

Prior to introducing autoregressive training with two-step trajectories, we observed artifacts in a small subset of ensemble members at longer lead times. However, these artifacts disappeared once the model was trained with autoregressive forecasting steps, suggesting that training with multiple autoregressive steps helps stabilize the usage of latent variables at longer lead times. 

In contrast to Graph-EFM and Diffusion-LAM, we do not apply any weighting of the loss function across atmospheric levels or variables, as we found it unnecessary to achieve competitive results. However, if it is desired to emphasize specific atmospheric levels or variables, such weighting could easily be incorporated into the loss function within this framework.

When preparing the dataset, we subsampled every fourth grid point, reducing the spatial resolution from the native \SI{2.5}{\kilo\meter} of MEPS to \SI{10}{\kilo\meter} (yielding $238 \times 268$ pixels). Consequently, spectral components at wavenumbers above $10^2$ are not physically meaningful, since the models cannot represent scales finer than those resolved by the input data. In an operational setting, such unresolved scales should be suppressed by applying a suitable low-pass filter.

\subsection{Detailed Evaluation}\label{apx:MEPS_detailed_results}
\begin{figure}[tbp]
    \centering
    \includegraphics[width=0.7\textwidth]{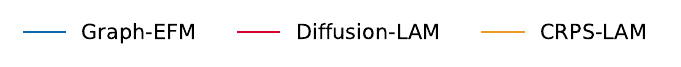}
    \begin{subfigure}[b]{0.3\textwidth}
        \centering
        \includegraphics[width=\textwidth]{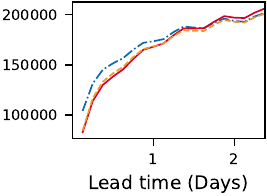}
        \caption{\texttt{nlwrs\_0}}
    \end{subfigure}
    \hfill
    \begin{subfigure}[b]{0.3\textwidth}
        \centering
        \includegraphics[width=\textwidth]{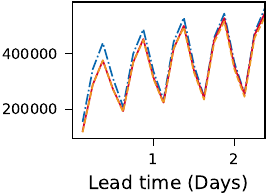}
        \caption{\texttt{nswrs\_0}}
    \end{subfigure}
    \hfill
    \begin{subfigure}[b]{0.3\textwidth}
        \centering
        \includegraphics[width=\textwidth]{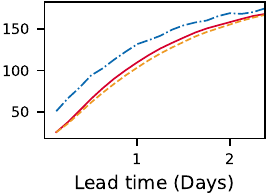}
        \caption{\texttt{pres\_0g}}
    \end{subfigure}
    \hfill
    \begin{subfigure}[b]{0.3\textwidth}
        \centering
        \includegraphics[width=\textwidth]{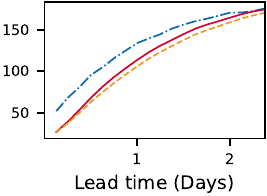}
        \caption{\texttt{pres\_0s}}
    \end{subfigure}
    \hfill
    \begin{subfigure}[b]{0.3\textwidth}
        \centering
        \includegraphics[width=\textwidth]{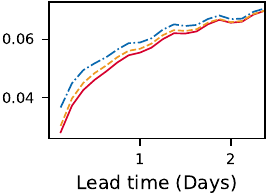}
        \caption{\texttt{r\_2}}
    \end{subfigure}
    \hfill
    \begin{subfigure}[b]{0.3\textwidth}
        \centering
        \includegraphics[width=\textwidth]{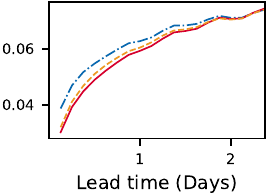}
        \caption{\texttt{r\_65}}
    \end{subfigure}
    \hfill
    \begin{subfigure}[b]{0.3\textwidth}
        \centering
        \includegraphics[width=\textwidth]{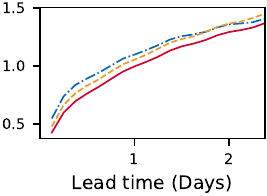}
        \caption{\texttt{t\_2}}
    \end{subfigure}
    \hfill
    \begin{subfigure}[b]{0.3\textwidth}
        \centering
        \includegraphics[width=\textwidth]{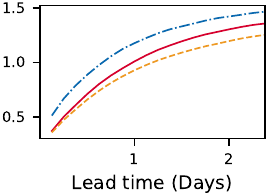}
        \caption{\texttt{t\_500}}
    \end{subfigure}
    \hfill
    \begin{subfigure}[b]{0.3\textwidth}
        \centering
        \includegraphics[width=\textwidth]{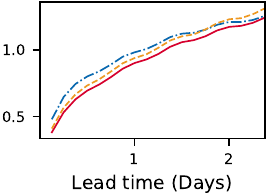}
        \caption{\texttt{t\_65}}
    \end{subfigure}
    \hfill
    \begin{subfigure}[b]{0.3\textwidth}
        \centering
        \includegraphics[width=\textwidth]{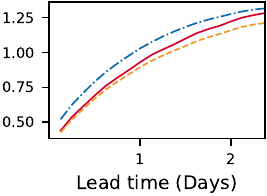}
        \caption{\texttt{t\_850}}
    \end{subfigure}
    \hfill
    \begin{subfigure}[b]{0.3\textwidth}
        \centering
        \includegraphics[width=\textwidth]{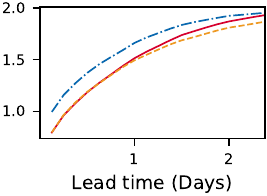}
        \caption{\texttt{u\_65}}
    \end{subfigure}
    \hfill
    \begin{subfigure}[b]{0.3\textwidth}
        \centering
        \includegraphics[width=\textwidth]{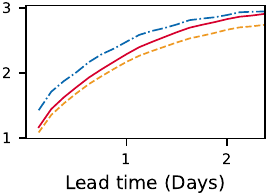}
        \caption{\texttt{u\_850}}
    \end{subfigure}
    \hfill
    \begin{subfigure}[b]{0.3\textwidth}
        \centering
        \includegraphics[width=\textwidth]{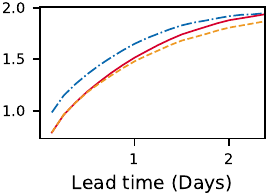}
        \caption{\texttt{v\_65}}
    \end{subfigure}
    \hfill
    \begin{subfigure}[b]{0.3\textwidth}
        \centering
        \includegraphics[width=\textwidth]{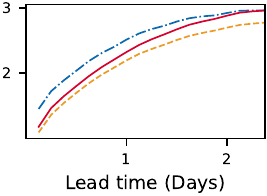}
        \caption{\texttt{v\_850}}
    \end{subfigure}
    \hfill
    \begin{subfigure}[b]{0.3\textwidth}
        \centering
        \includegraphics[width=\textwidth]{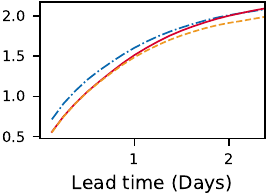}
        \caption{\texttt{wvint\_0}}
    \end{subfigure}
    \hfill
    \begin{subfigure}[b]{0.3\textwidth}
        \centering
        \includegraphics[width=\textwidth]{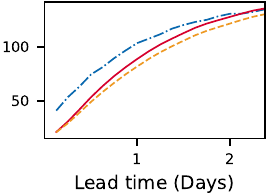}
        \caption{\texttt{z\_1000}}
    \end{subfigure}
    \begin{subfigure}[b]{0.3\textwidth}
        \centering
        \includegraphics[width=\textwidth]{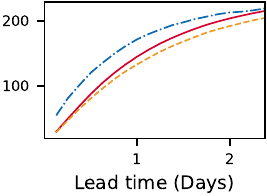}
        \caption{\texttt{z\_500}}
    \end{subfigure}
    \hfill
    \caption{The \acrshort{rmse} results for each variable.}
    \label{fig:rmse_all}
\end{figure}

\begin{figure}[tbp]
    \centering
    \includegraphics[width=0.7\textwidth]{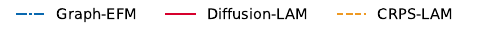}
    \begin{subfigure}[b]{0.3\textwidth}
        \centering
        \includegraphics[width=\textwidth]{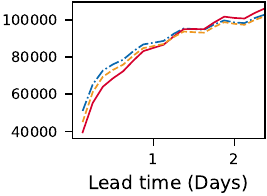}
        \caption{\texttt{nlwrs\_0}}
    \end{subfigure}
    \hfill
    \begin{subfigure}[b]{0.3\textwidth}
        \centering
        \includegraphics[width=\textwidth]{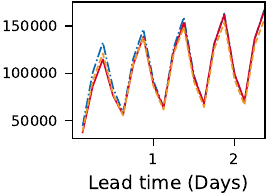}
        \caption{\texttt{nswrs\_0}}
    \end{subfigure}
    \hfill
    \begin{subfigure}[b]{0.3\textwidth}
        \centering
        \includegraphics[width=\textwidth]{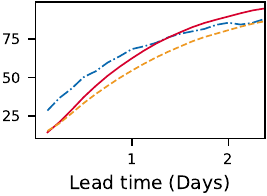}
        \caption{\texttt{pres\_0g}}
    \end{subfigure}
    \hfill
    \begin{subfigure}[b]{0.3\textwidth}
        \centering
        \includegraphics[width=\textwidth]{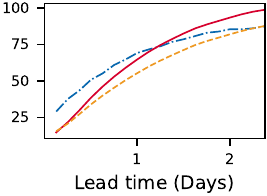}
        \caption{\texttt{pres\_0s}}
    \end{subfigure}
    \hfill
    \begin{subfigure}[b]{0.3\textwidth}
        \centering
        \includegraphics[width=\textwidth]{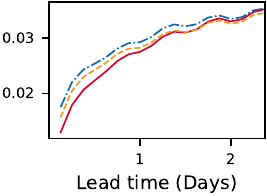}
        \caption{\texttt{r\_2}}
    \end{subfigure}
    \hfill
    \begin{subfigure}[b]{0.3\textwidth}
        \centering
        \includegraphics[width=\textwidth]{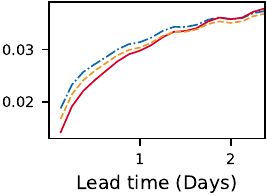}
        \caption{\texttt{r\_65}}
    \end{subfigure}
    \hfill
    \begin{subfigure}[b]{0.3\textwidth}
        \centering
        \includegraphics[width=\textwidth]{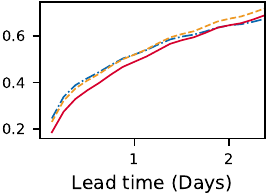}
        \caption{\texttt{t\_2}}
    \end{subfigure}
    \hfill
    \begin{subfigure}[b]{0.3\textwidth}
        \centering
        \includegraphics[width=\textwidth]{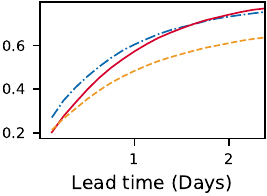}
        \caption{\texttt{t\_500}}
    \end{subfigure}
    \hfill
    \begin{subfigure}[b]{0.3\textwidth}
        \centering
        \includegraphics[width=\textwidth]{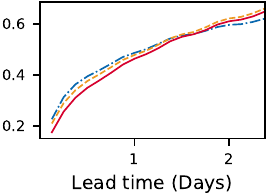}
        \caption{\texttt{t\_65}}
    \end{subfigure}
    \hfill
    \begin{subfigure}[b]{0.3\textwidth}
        \centering
        \includegraphics[width=\textwidth]{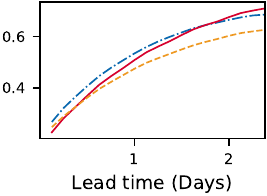}
        \caption{\texttt{t\_850}}
    \end{subfigure}
    \hfill
    \begin{subfigure}[b]{0.3\textwidth}
        \centering
        \includegraphics[width=\textwidth]{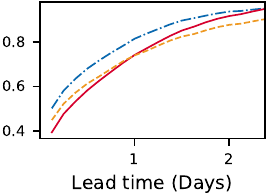}
        \caption{\texttt{u\_65}}
    \end{subfigure}
    \hfill
    \begin{subfigure}[b]{0.3\textwidth}
        \centering
        \includegraphics[width=\textwidth]{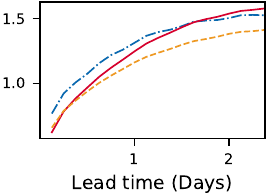}
        \caption{\texttt{u\_850}}
    \end{subfigure}
    \hfill
    \begin{subfigure}[b]{0.3\textwidth}
        \centering
        \includegraphics[width=\textwidth]{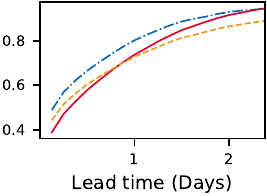}
        \caption{\texttt{v\_65}}
    \end{subfigure}
    \hfill
    \begin{subfigure}[b]{0.3\textwidth}
        \centering
        \includegraphics[width=\textwidth]{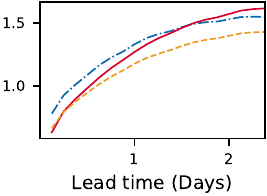}
        \caption{\texttt{v\_850}}
    \end{subfigure}
    \hfill
    \begin{subfigure}[b]{0.3\textwidth}
        \centering
        \includegraphics[width=\textwidth]{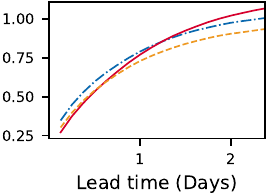}
        \caption{\texttt{wvint\_0}}
    \end{subfigure}
    \hfill
    \begin{subfigure}[b]{0.3\textwidth}
        \centering
        \includegraphics[width=\textwidth]{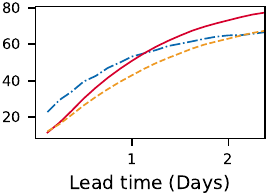}
        \caption{\texttt{z\_1000}}
    \end{subfigure}
    \begin{subfigure}[b]{0.3\textwidth}
        \centering
        \includegraphics[width=\textwidth]{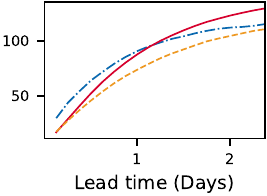}
        \caption{\texttt{z\_500}}
    \end{subfigure}
    \hfill
    \caption{The \acrshort{crps} results for each variable.}
    \label{fig:crps_all}
\end{figure}

\begin{figure}[tbp]
    \centering
    \includegraphics[width=\textwidth]{figures/spskr/spskr_legend.pdf}
    \begin{subfigure}[b]{0.3\textwidth}
        \centering
        \includegraphics[width=\textwidth]{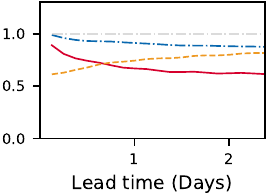}
        \caption{\texttt{nlwrs\_0}}
    \end{subfigure}
    \hfill
    \begin{subfigure}[b]{0.3\textwidth}
        \centering
        \includegraphics[width=\textwidth]{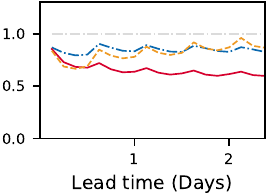}
        \caption{\texttt{nswrs\_0}}
    \end{subfigure}
    \hfill
    \begin{subfigure}[b]{0.3\textwidth}
        \centering
        \includegraphics[width=\textwidth]{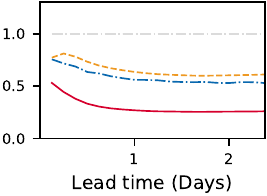}
        \caption{\texttt{pres\_0g}}
    \end{subfigure}
    \hfill
    \begin{subfigure}[b]{0.3\textwidth}
        \centering
        \includegraphics[width=\textwidth]{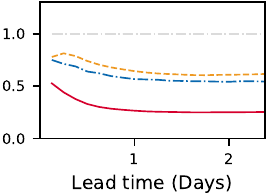}
        \caption{\texttt{pres\_0s}}
    \end{subfigure}
    \hfill
    \begin{subfigure}[b]{0.3\textwidth}
        \centering
        \includegraphics[width=\textwidth]{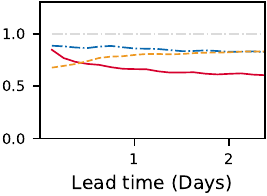}
        \caption{\texttt{r\_2}}
    \end{subfigure}
    \hfill
    \begin{subfigure}[b]{0.3\textwidth}
        \centering
        \includegraphics[width=\textwidth]{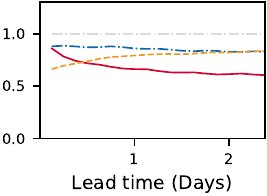}
        \caption{\texttt{r\_65}}
    \end{subfigure}
    \hfill
    \begin{subfigure}[b]{0.3\textwidth}
        \centering
        \includegraphics[width=\textwidth]{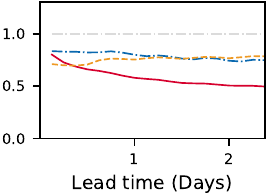}
        \caption{\texttt{t\_2}}
    \end{subfigure}
    \hfill
    \begin{subfigure}[b]{0.3\textwidth}
        \centering
        \includegraphics[width=\textwidth]{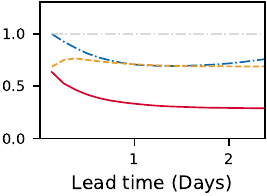}
        \caption{\texttt{t\_500}}
    \end{subfigure}
    \hfill
    \begin{subfigure}[b]{0.3\textwidth}
        \centering
        \includegraphics[width=\textwidth]{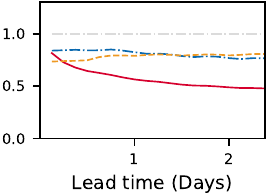}
        \caption{\texttt{t\_65}}
    \end{subfigure}
    \hfill
    \begin{subfigure}[b]{0.3\textwidth}
        \centering
        \includegraphics[width=\textwidth]{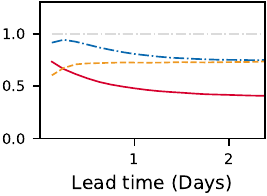}
        \caption{\texttt{t\_850}}
    \end{subfigure}
    \hfill
    \begin{subfigure}[b]{0.3\textwidth}
        \centering
        \includegraphics[width=\textwidth]{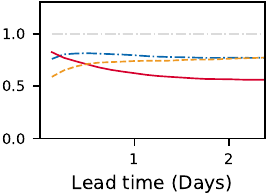}
        \caption{\texttt{u\_65}}
    \end{subfigure}
    \hfill
    \begin{subfigure}[b]{0.3\textwidth}
        \centering
        \includegraphics[width=\textwidth]{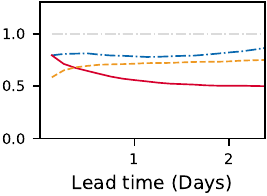}
        \caption{\texttt{u\_850}}
    \end{subfigure}
    \hfill
    \begin{subfigure}[b]{0.3\textwidth}
        \centering
        \includegraphics[width=\textwidth]{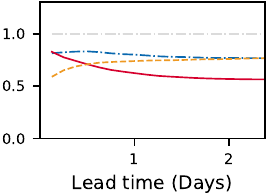}
        \caption{\texttt{v\_65}}
    \end{subfigure}
    \hfill
    \begin{subfigure}[b]{0.3\textwidth}
        \centering
        \includegraphics[width=\textwidth]{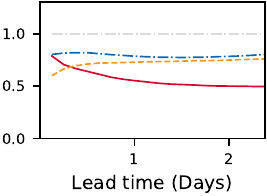}
        \caption{\texttt{v\_850}}
    \end{subfigure}
    \hfill
    \begin{subfigure}[b]{0.3\textwidth}
        \centering
        \includegraphics[width=\textwidth]{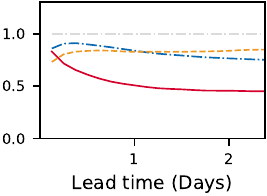}
        \caption{\texttt{wvint\_0}}
    \end{subfigure}
    \hfill
    \begin{subfigure}[b]{0.3\textwidth}
        \centering
        \includegraphics[width=\textwidth]{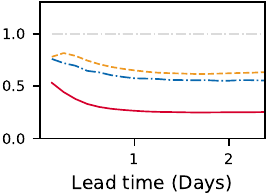}
        \caption{\texttt{z\_1000}}
    \end{subfigure}
    \begin{subfigure}[b]{0.3\textwidth}
        \centering
        \includegraphics[width=\textwidth]{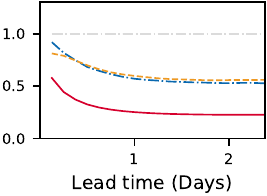}
        \caption{\texttt{z\_500}}
    \end{subfigure}
    \hfill
    \caption{The \acrshort{ssr} results for each variable.}
    \label{fig:spskr_all}
\end{figure}

\begin{figure}[h!]
    \centering
    \includegraphics[width=\textwidth]{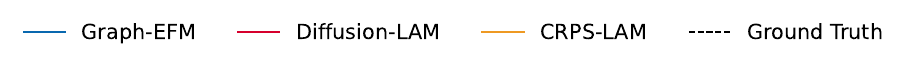}
    \begin{subfigure}[b]{0.32\textwidth}
        \centering
        \includegraphics[width=\textwidth]{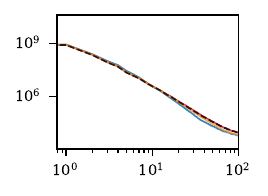}
        \caption{\texttt{nlwrs\_0} at \SI{3}{\hour}}
    \end{subfigure}
    \hfill
    \begin{subfigure}[b]{0.32\textwidth}
        \centering
        \includegraphics[width=\textwidth]{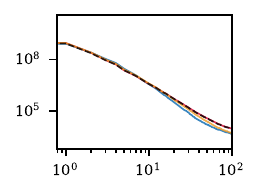}
        \caption{\texttt{nlwrs\_0} at \SI{30}{\hour}}
    \end{subfigure}
    \hfill
    \begin{subfigure}[b]{0.32\textwidth}
        \centering
        \includegraphics[width=\textwidth]{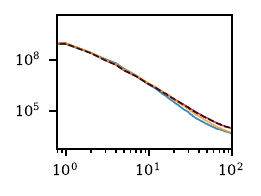}
        \caption{\texttt{nlwrs\_0} at \SI{57}{\hour}}
    \end{subfigure}
    \hfill
    \begin{subfigure}[b]{0.32\textwidth}
        \centering
        \includegraphics[width=\textwidth]{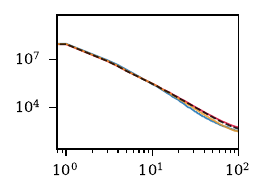}
        \caption{\texttt{nswrs\_0} at \SI{3}{\hour}}
    \end{subfigure}
    \hfill
    \begin{subfigure}[b]{0.32\textwidth}
        \centering
        \includegraphics[width=\textwidth]{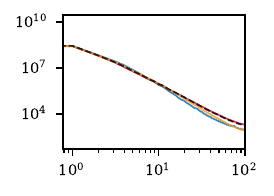}
        \caption{\texttt{nswrs\_0} at \SI{30}{\hour}}
    \end{subfigure}
    \hfill
    \begin{subfigure}[b]{0.32\textwidth}
        \centering
        \includegraphics[width=\textwidth]{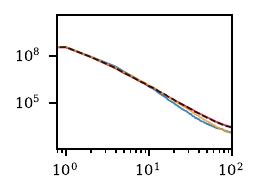}
        \caption{\texttt{nswrs\_0} at \SI{57}{\hour}}
    \end{subfigure}
    \hfill
    \begin{subfigure}[b]{0.32\textwidth}
        \centering
        \includegraphics[width=\textwidth]{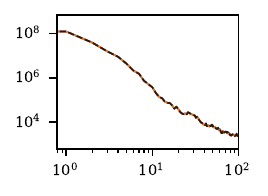}
        \caption{\texttt{pres\_0g} at \SI{3}{\hour}}
    \end{subfigure}
    \hfill
    \begin{subfigure}[b]{0.32\textwidth}
        \centering
        \includegraphics[width=\textwidth]{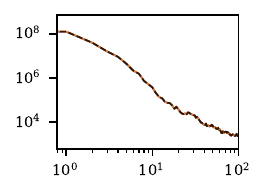}
        \caption{\texttt{pres\_0g} at \SI{30}{\hour}}
    \end{subfigure}
    \hfill
    \begin{subfigure}[b]{0.32\textwidth}
        \centering
        \includegraphics[width=\textwidth]{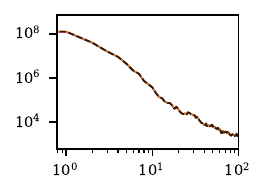}
        \caption{\texttt{pres\_0g} at \SI{57}{\hour}}
    \end{subfigure}
    \hfill
    \begin{subfigure}[b]{0.32\textwidth}
        \centering
        \includegraphics[width=\textwidth]{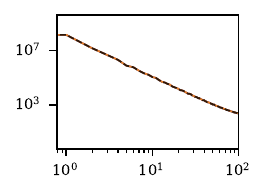}
        \caption{\texttt{pres\_0s} at \SI{3}{\hour}}
    \end{subfigure}
    \hfill
    \begin{subfigure}[b]{0.32\textwidth}
        \centering
        \includegraphics[width=\textwidth]{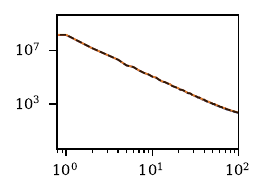}
        \caption{\texttt{pres\_0s} at \SI{30}{\hour}}
    \end{subfigure}
    \hfill
    \begin{subfigure}[b]{0.32\textwidth}
        \centering
        \includegraphics[width=\textwidth]{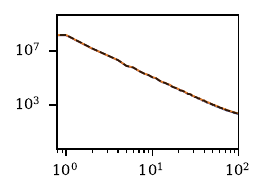}
        \caption{\texttt{pres\_0s} at \SI{57}{\hour}}
    \end{subfigure}
    \hfill
    \begin{subfigure}[b]{0.32\textwidth}
        \centering
        \includegraphics[width=\textwidth]{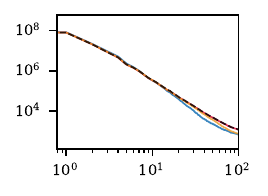}
        \caption{\texttt{r\_2} at \SI{3}{\hour}}
    \end{subfigure}
    \hfill
    \begin{subfigure}[b]{0.32\textwidth}
        \centering
        \includegraphics[width=\textwidth]{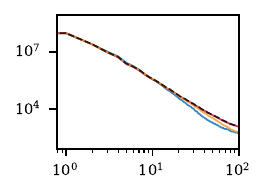}
        \caption{\texttt{r\_2} at \SI{30}{\hour}}
    \end{subfigure}
    \hfill
    \begin{subfigure}[b]{0.32\textwidth}
        \centering
        \includegraphics[width=\textwidth]{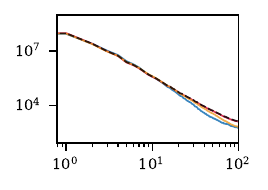}
        \caption{\texttt{r\_2} at \SI{57}{\hour}}
    \end{subfigure}
    \hfill
    \caption{The energy spectra for each variable at the lead times \SI{3}{\hour}, \SI{30}{\hour}, and \SI{57}{\hour}.}
    \label{fig:spectra_all_0}
\end{figure}

\begin{figure}[tbp]
    \centering
    \begin{subfigure}[b]{0.32\textwidth}
        \centering
        \includegraphics[width=\textwidth]{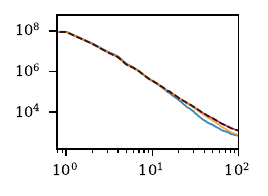}
        \caption{\texttt{r\_65} at \SI{3}{\hour}}
    \end{subfigure}
    \hfill
    \begin{subfigure}[b]{0.32\textwidth}
        \centering
        \includegraphics[width=\textwidth]{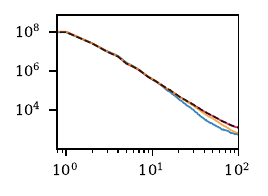}
        \caption{\texttt{r\_65} at \SI{30}{\hour}}
    \end{subfigure}
    \hfill
    \begin{subfigure}[b]{0.32\textwidth}
        \centering
        \includegraphics[width=\textwidth]{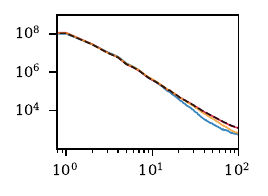}
        \caption{\texttt{r\_65} at \SI{57}{\hour}}
    \end{subfigure}
    \hfill
    \begin{subfigure}[b]{0.32\textwidth}
        \centering
        \includegraphics[width=\textwidth]{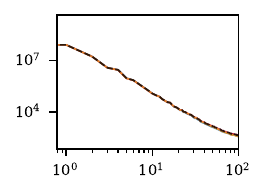}
        \caption{\texttt{t\_2} at \SI{3}{\hour}}
    \end{subfigure}
    \hfill
    \begin{subfigure}[b]{0.32\textwidth}
        \centering
        \includegraphics[width=\textwidth]{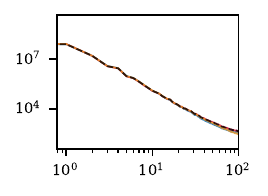}
        \caption{\texttt{t\_2} at \SI{30}{\hour}}
    \end{subfigure}
    \hfill
    \begin{subfigure}[b]{0.32\textwidth}
        \centering
        \includegraphics[width=\textwidth]{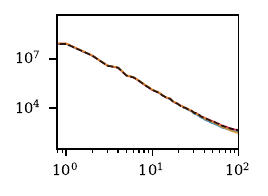}
        \caption{\texttt{t\_2} at \SI{57}{\hour}}
    \end{subfigure}
    \hfill
    \begin{subfigure}[b]{0.32\textwidth}
        \centering
        \includegraphics[width=\textwidth]{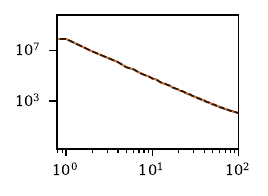}
        \caption{\texttt{t\_500} at \SI{3}{\hour}}
    \end{subfigure}
    \hfill
    \begin{subfigure}[b]{0.32\textwidth}
        \centering
        \includegraphics[width=\textwidth]{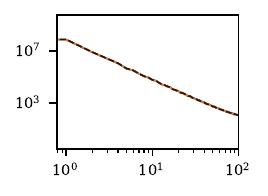}
        \caption{\texttt{t\_500} at \SI{30}{\hour}}
    \end{subfigure}
    \hfill
    \begin{subfigure}[b]{0.32\textwidth}
        \centering
        \includegraphics[width=\textwidth]{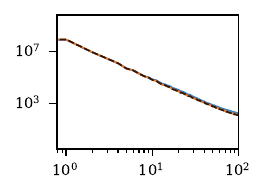}
        \caption{\texttt{t\_500} at \SI{57}{\hour}}
    \end{subfigure}
    \hfill
    \begin{subfigure}[b]{0.32\textwidth}
        \centering
        \includegraphics[width=\textwidth]{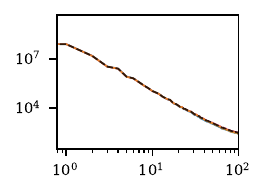}
        \caption{\texttt{t\_65} at \SI{3}{\hour}}
    \end{subfigure}
    \hfill
    \begin{subfigure}[b]{0.32\textwidth}
        \centering
        \includegraphics[width=\textwidth]{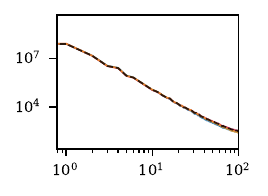}
        \caption{\texttt{t\_65} at \SI{30}{\hour}}
    \end{subfigure}
    \hfill
    \begin{subfigure}[b]{0.32\textwidth}
        \centering
        \includegraphics[width=\textwidth]{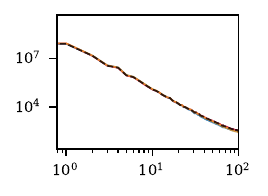}
        \caption{\texttt{t\_65} at \SI{57}{\hour}}
    \end{subfigure}
    \hfill
    \begin{subfigure}[b]{0.32\textwidth}
        \centering
        \includegraphics[width=\textwidth]{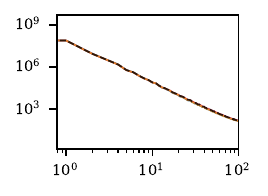}
        \caption{\texttt{t\_850} at \SI{3}{\hour}}
    \end{subfigure}
    \hfill
    \begin{subfigure}[b]{0.32\textwidth}
        \centering
        \includegraphics[width=\textwidth]{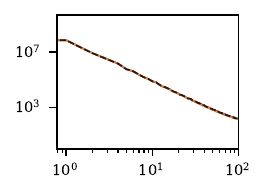}
        \caption{\texttt{t\_850} at \SI{30}{\hour}}
    \end{subfigure}
    \hfill
    \begin{subfigure}[b]{0.32\textwidth}
        \centering
        \includegraphics[width=\textwidth]{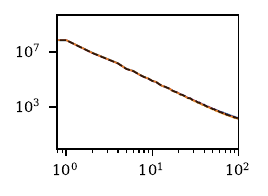}
        \caption{\texttt{t\_850} at \SI{57}{\hour}}
    \end{subfigure}
    \hfill

    \caption{The energy spectra for each variable at the lead times \SI{3}{\hour}, \SI{30}{\hour}, and \SI{57}{\hour}.}
    \label{fig:spectra_all_1}
\end{figure}

\begin{figure}[tbp]
    \centering
    \begin{subfigure}[b]{0.32\textwidth}
        \centering
        \includegraphics[width=\textwidth]{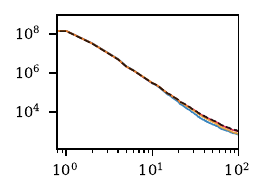}
        \caption{\texttt{u\_65} at \SI{3}{\hour}}
    \end{subfigure}
    \hfill
    \begin{subfigure}[b]{0.32\textwidth}
        \centering
        \includegraphics[width=\textwidth]{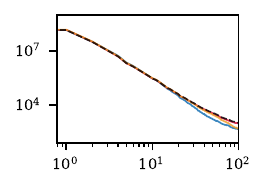}
        \caption{\texttt{u\_65} at \SI{30}{\hour}}
    \end{subfigure}
    \hfill
    \begin{subfigure}[b]{0.32\textwidth}
        \centering
        \includegraphics[width=\textwidth]{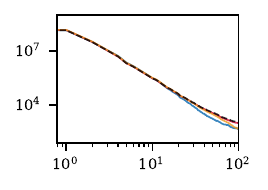}
        \caption{\texttt{u\_65} at \SI{57}{\hour}}
    \end{subfigure}
    \hfill
    \begin{subfigure}[b]{0.32\textwidth}
        \centering
        \includegraphics[width=\textwidth]{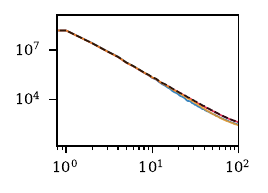}
        \caption{\texttt{u\_850} at \SI{3}{\hour}}
    \end{subfigure}
    \hfill
    \begin{subfigure}[b]{0.32\textwidth}
        \centering
        \includegraphics[width=\textwidth]{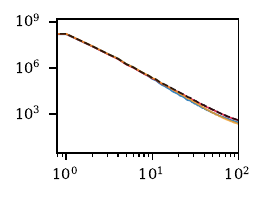}
        \caption{\texttt{u\_850} at \SI{30}{\hour}}
    \end{subfigure}
    \hfill
    \begin{subfigure}[b]{0.32\textwidth}
        \centering
        \includegraphics[width=\textwidth]{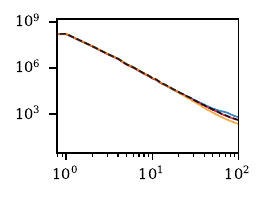}
        \caption{\texttt{u\_850} at \SI{57}{\hour}}
    \end{subfigure}
    \hfill
    \begin{subfigure}[b]{0.32\textwidth}
        \centering
        \includegraphics[width=\textwidth]{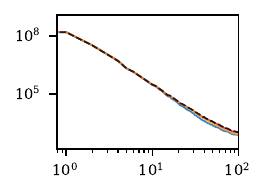}
        \caption{\texttt{v\_65} at \SI{3}{\hour}}
    \end{subfigure}
    \hfill
    \begin{subfigure}[b]{0.32\textwidth}
        \centering
        \includegraphics[width=\textwidth]{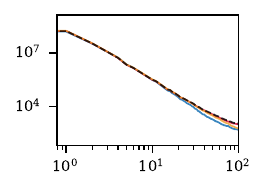}
        \caption{\texttt{v\_65} at \SI{30}{\hour}}
    \end{subfigure}
    \hfill
    \begin{subfigure}[b]{0.32\textwidth}
        \centering
        \includegraphics[width=\textwidth]{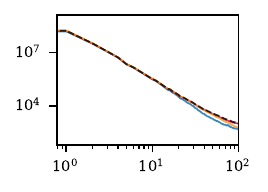}
        \caption{\texttt{v\_65} at \SI{57}{\hour}}
    \end{subfigure}
    \hfill
    \begin{subfigure}[b]{0.32\textwidth}
        \centering
        \includegraphics[width=\textwidth]{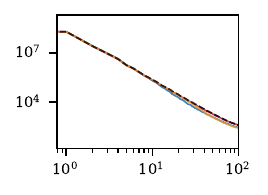}
        \caption{\texttt{v\_850} at \SI{3}{\hour}}
    \end{subfigure}
    \hfill
    \begin{subfigure}[b]{0.32\textwidth}
        \centering
        \includegraphics[width=\textwidth]{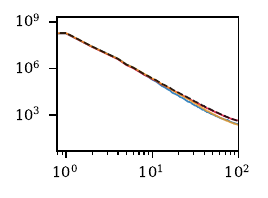}
        \caption{\texttt{v\_850} at \SI{30}{\hour}}
    \end{subfigure}
    \hfill
    \begin{subfigure}[b]{0.32\textwidth}
        \centering
        \includegraphics[width=\textwidth]{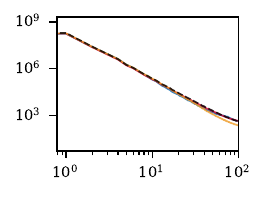}
        \caption{\texttt{v\_850} at \SI{57}{\hour}}
    \end{subfigure}
    \hfill
    \begin{subfigure}[b]{0.32\textwidth}
        \centering
        \includegraphics[width=\textwidth]{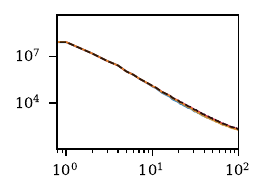}
        \caption{\texttt{wvint\_0} at \SI{3}{\hour}}
    \end{subfigure}
    \hfill
    \begin{subfigure}[b]{0.32\textwidth}
        \centering
        \includegraphics[width=\textwidth]{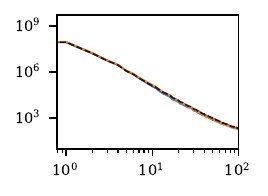}
        \caption{\texttt{wvint\_0} at \SI{30}{\hour}}
    \end{subfigure}
    \hfill
    \begin{subfigure}[b]{0.32\textwidth}
        \centering
        \includegraphics[width=\textwidth]{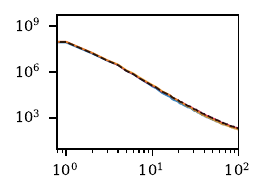}
        \caption{\texttt{wvint\_0} at \SI{57}{\hour}}
    \end{subfigure}
    \hfill
    \caption{The energy spectra for each variable at the lead times \SI{3}{\hour}, \SI{30}{\hour}, and \SI{57}{\hour}.}
    \label{fig:spectra_all_2}
\end{figure}

\begin{figure}[tbp]
    \centering
    \begin{subfigure}[b]{0.32\textwidth}
        \centering
        \includegraphics[width=\textwidth]{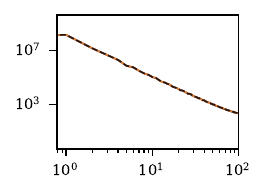}
        \caption{\texttt{z\_1000} at \SI{3}{\hour}}
    \end{subfigure}
    \hfill
    \begin{subfigure}[b]{0.32\textwidth}
        \centering
        \includegraphics[width=\textwidth]{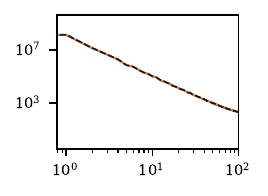}
        \caption{\texttt{z\_1000} at \SI{30}{\hour}}
    \end{subfigure}
    \hfill
    \begin{subfigure}[b]{0.32\textwidth}
        \centering
        \includegraphics[width=\textwidth]{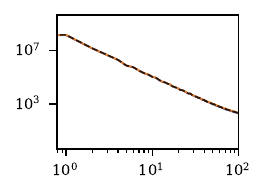}
        \caption{\texttt{z\_1000} at \SI{57}{\hour}}
    \end{subfigure}
    \hfill
    \begin{subfigure}[b]{0.32\textwidth}
        \centering
        \includegraphics[width=\textwidth]{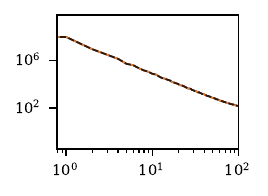}
        \caption{\texttt{z\_500} at \SI{3}{\hour}}
    \end{subfigure}
    \hfill
    \begin{subfigure}[b]{0.32\textwidth}
        \centering
        \includegraphics[width=\textwidth]{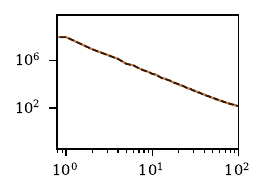}
        \caption{\texttt{z\_500} at \SI{30}{\hour}}
    \end{subfigure}
    \hfill
    \begin{subfigure}[b]{0.32\textwidth}
        \centering
        \includegraphics[width=\textwidth]{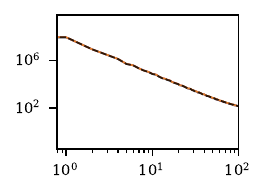}
        \caption{\texttt{z\_500} at \SI{57}{\hour}}
    \end{subfigure}
    \hfill
    \caption{The energy spectra for each variable at the lead times \SI{3}{\hour}, \SI{30}{\hour}, and \SI{57}{\hour}.}
    \label{fig:spectra_all_3}
\end{figure}

We provide detailed per-variable evaluation results in \cref{fig:rmse_all}, \cref{fig:crps_all}, and \cref{fig:spskr_all}, together with a \SI{57}{\hour} forecast example for a randomly selected test case for a subset of the variables in \cref{fig:samples_all}. The corresponding energy spectra for all variables are shown in \cref{fig:spectra_all_0,fig:spectra_all_1,fig:spectra_all_2,fig:spectra_all_3}. For an evaluation of deterministic models and less competitive probabilistic baselines on the MEPS dataset, we refer the reader to \citet{oskarsson2023graph-lam,oskarsson2024probabilistic}.
\begin{figure}[tbp]
    \centering
    \begin{subfigure}[b]{\textwidth}
        \centering
        \includegraphics[width=\textwidth,height=0.98\textheight,keepaspectratio]{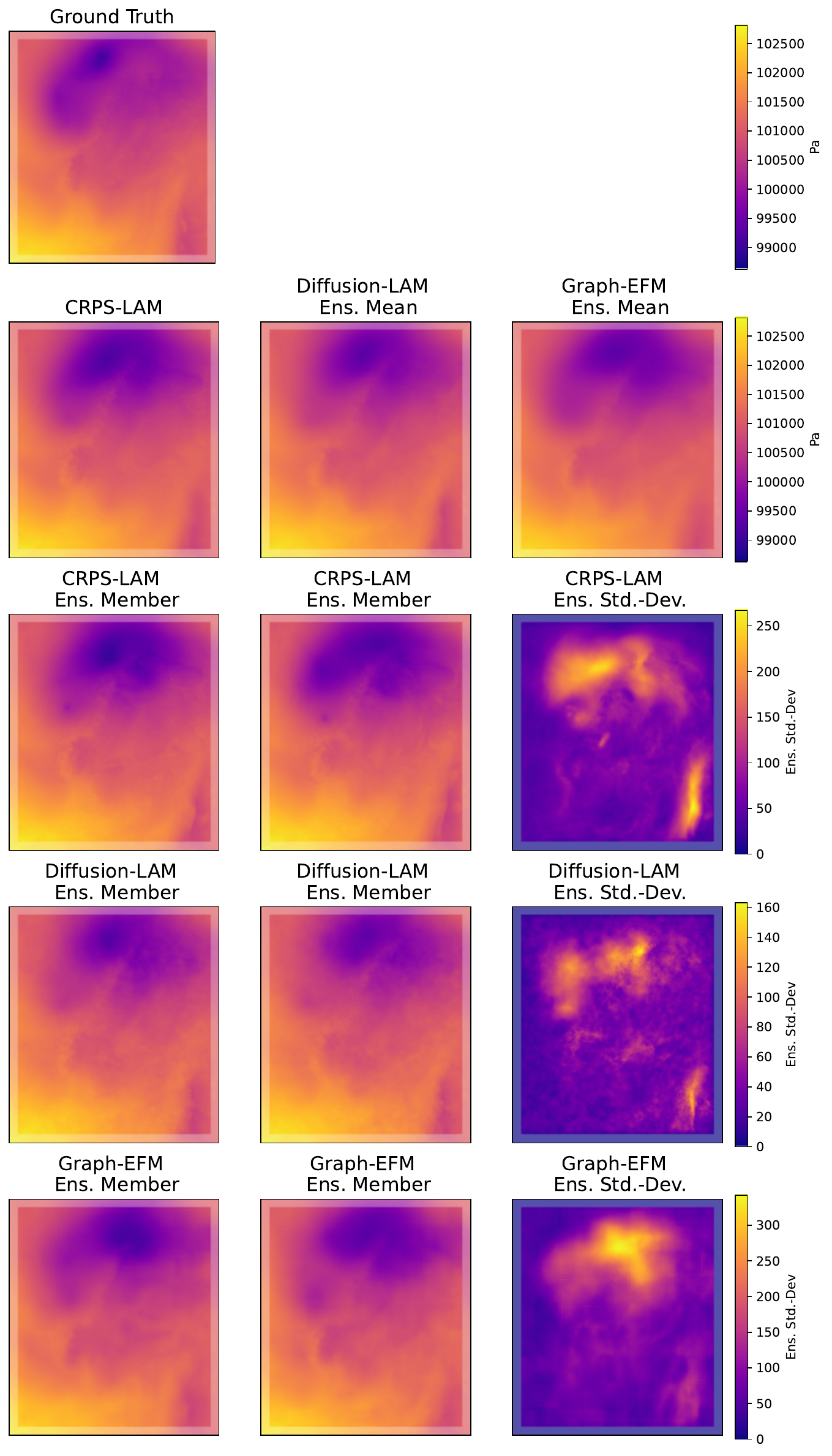}
        \caption{\texttt{pres\_0s}}
    \end{subfigure}
\end{figure}
\begin{figure}[tbp]\ContinuedFloat
    \centering
    \begin{subfigure}[b]{\textwidth}
        \centering
        \includegraphics[width=\textwidth,height=0.98\textheight,keepaspectratio]{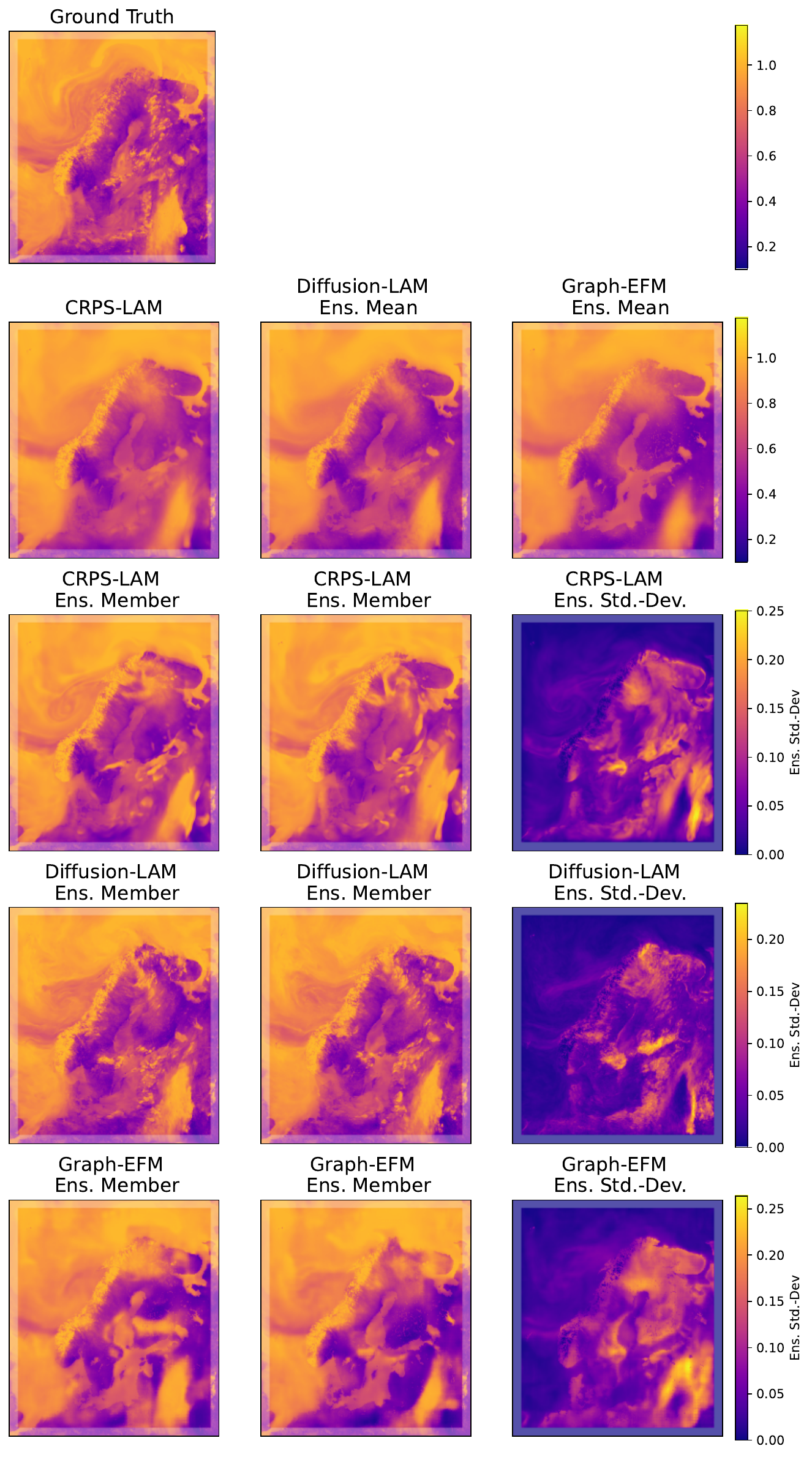}
        \caption{\texttt{r\_2}}
    \end{subfigure}
\end{figure}
\begin{figure}[tbp]\ContinuedFloat
    \centering
    \begin{subfigure}[b]{\textwidth}
        \centering
        \includegraphics[width=\textwidth,height=0.98\textheight,keepaspectratio]{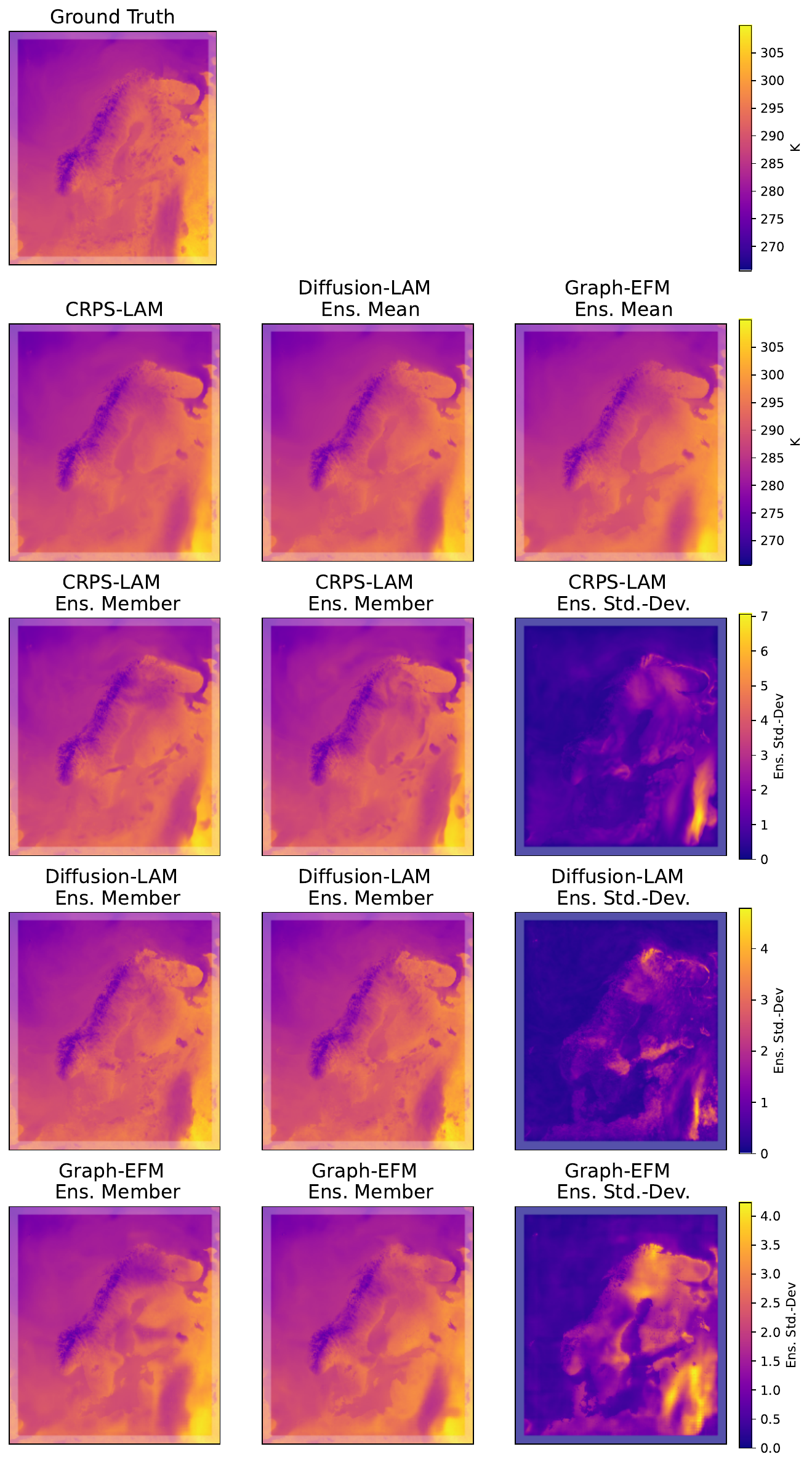}
        \caption{\texttt{t\_2}}
    \end{subfigure}
\end{figure}

\begin{figure}[tbp]\ContinuedFloat
    \centering
    \begin{subfigure}[b]{\textwidth}
        \centering
        \includegraphics[width=\textwidth,height=0.98\textheight,keepaspectratio]{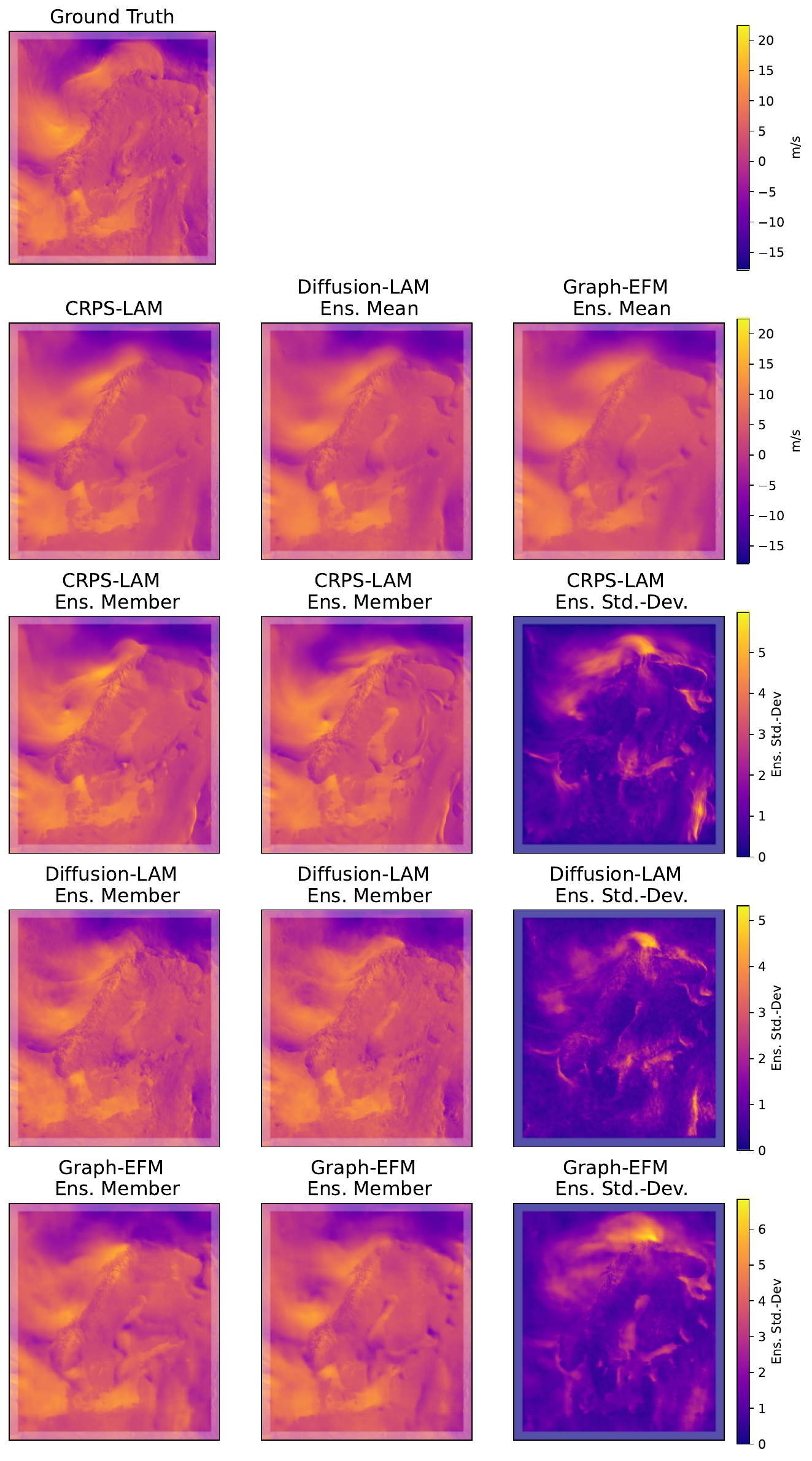}
        \caption{\texttt{u\_65}}
    \end{subfigure}
\end{figure}
\begin{figure}[tbp]\ContinuedFloat
    \centering
    \begin{subfigure}[b]{\textwidth}
        \centering
        \includegraphics[width=\textwidth,height=0.98\textheight,keepaspectratio]{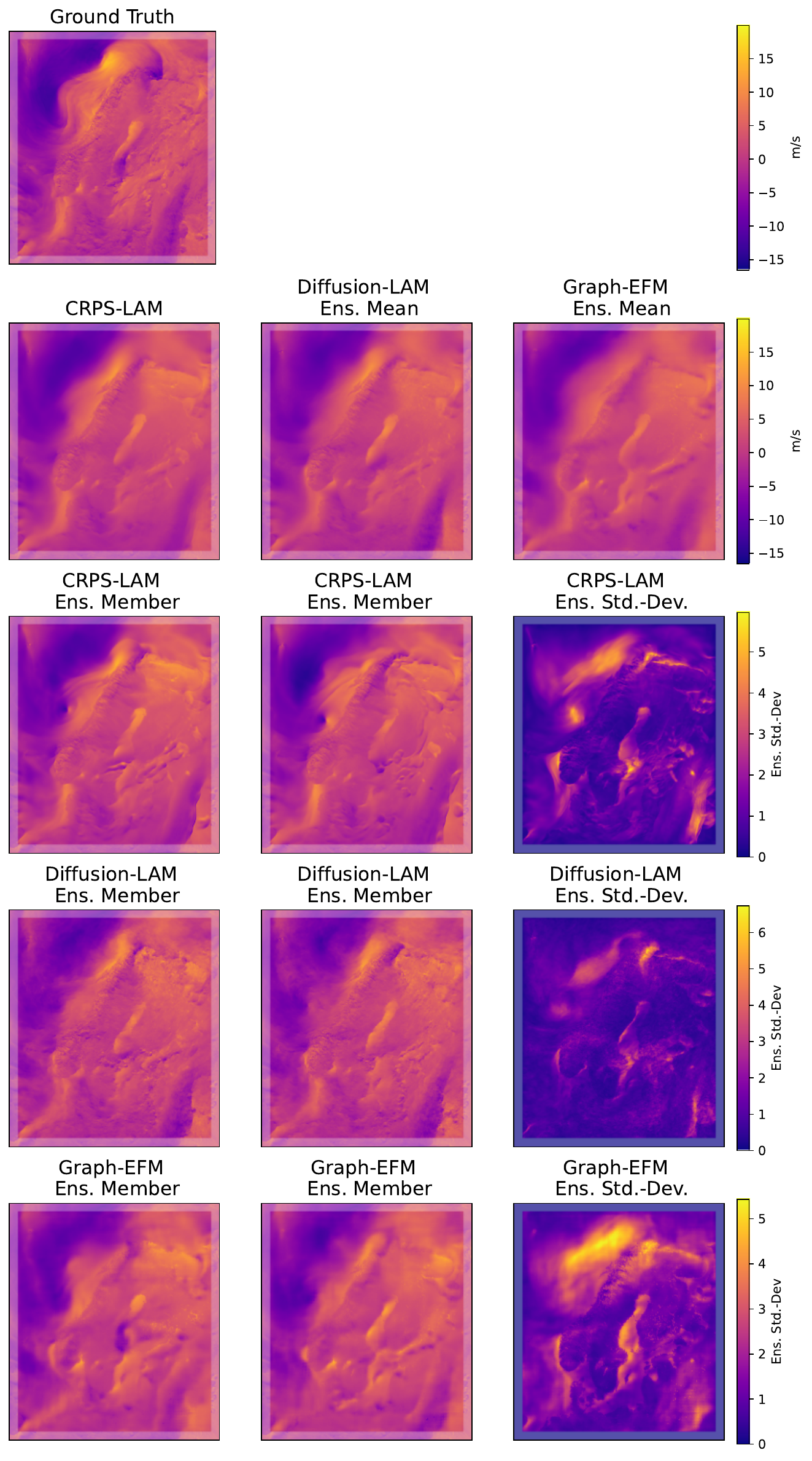}
        \caption{\texttt{v\_65}}
    \end{subfigure}
\end{figure}
\begin{figure}[tbp]\ContinuedFloat
    \centering
    \begin{subfigure}[b]{\textwidth}
        \centering
        \includegraphics[width=\textwidth,height=0.98\textheight,keepaspectratio]{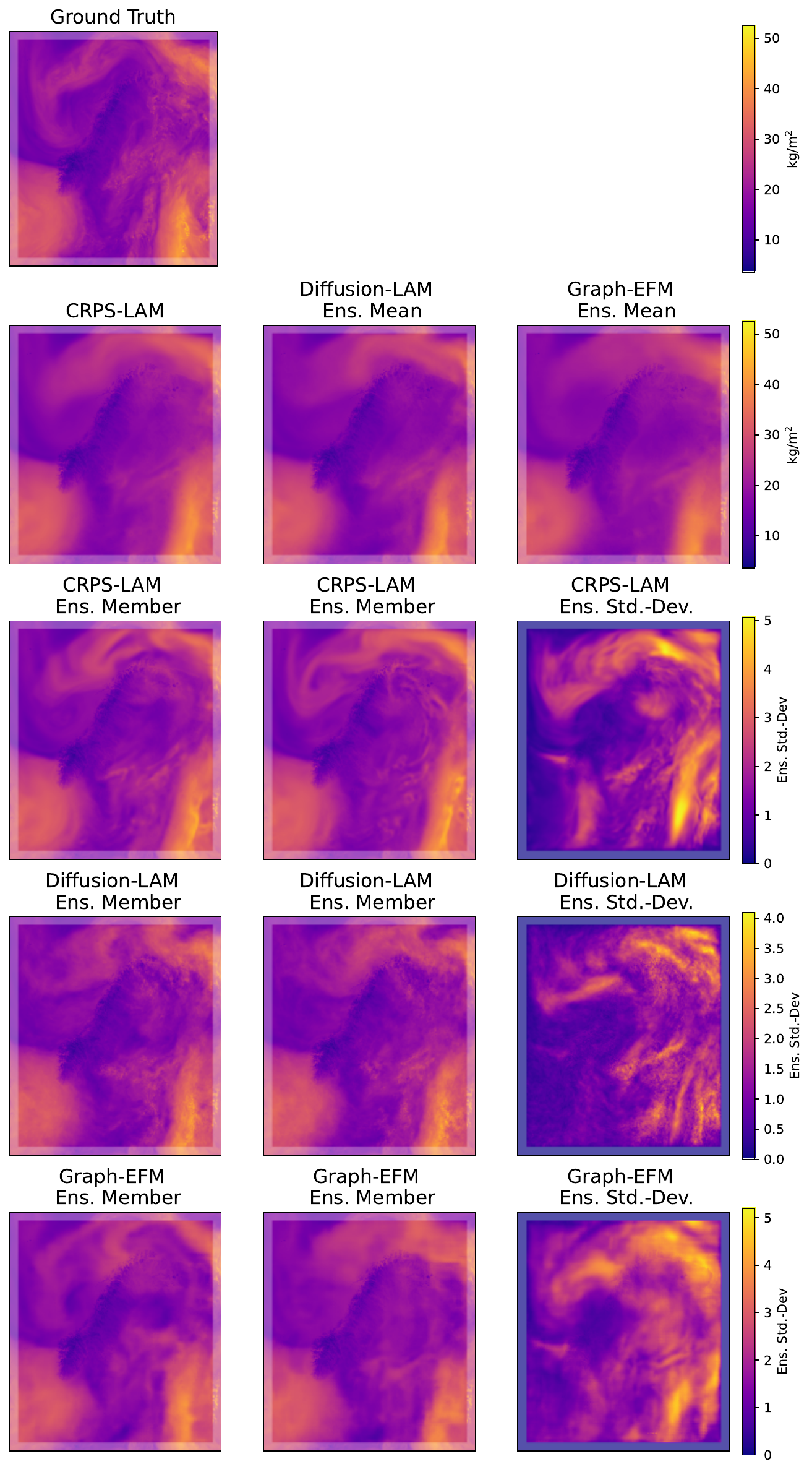}
        \caption{\texttt{wvint\_0}}
    \end{subfigure}
\end{figure}
\begin{figure}[tbp]\ContinuedFloat
    \centering
    \begin{subfigure}[b]{\textwidth}
        \centering
        \includegraphics[width=\textwidth,height=0.98\textheight,keepaspectratio]{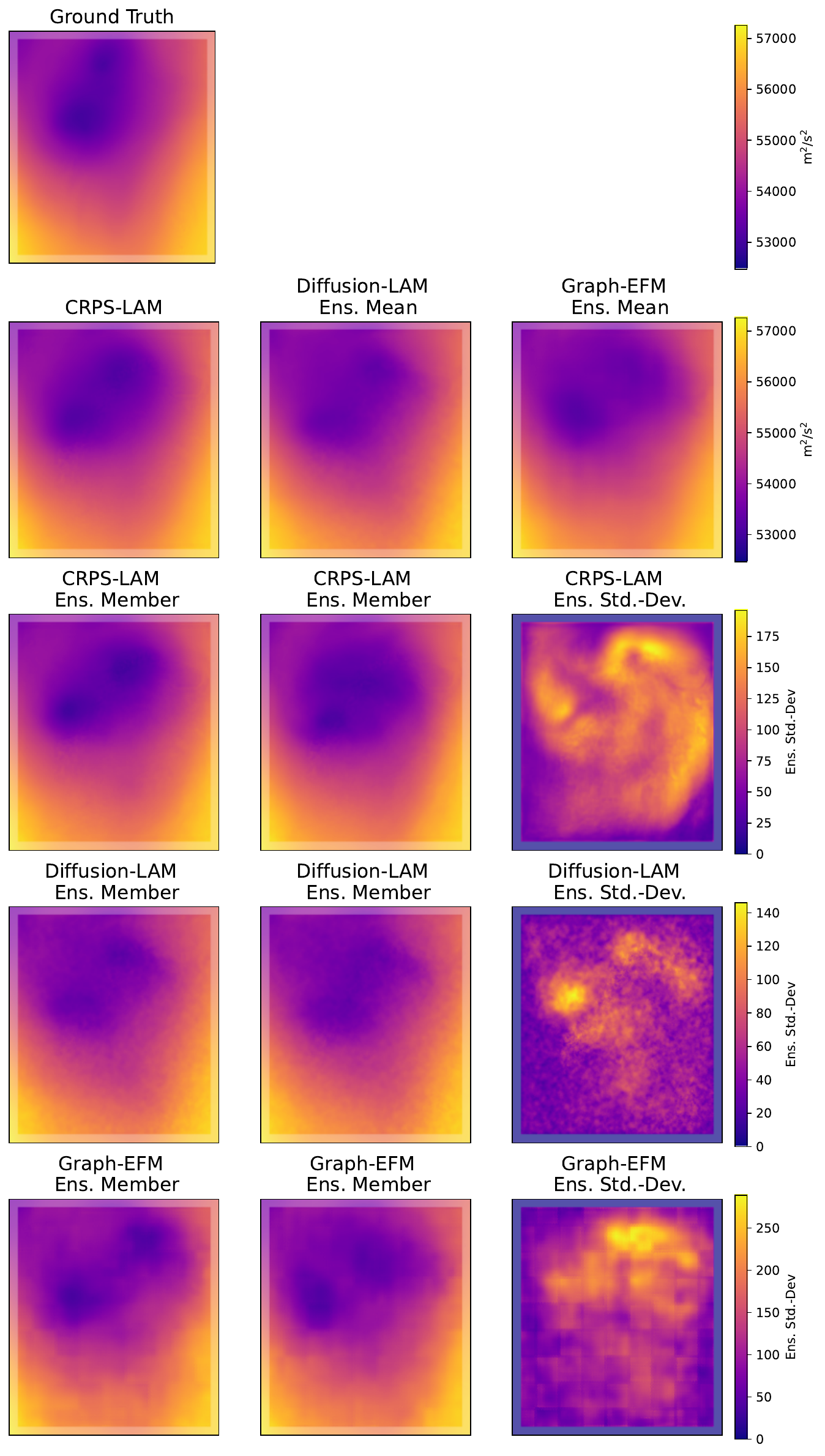}
        \caption{\texttt{z\_500}}
    \end{subfigure}
    \caption{An ensemble forecasts for a few variables at \SI{57}{\hour}.}
    \label{fig:samples_all}
\end{figure}

\end{document}